\documentclass[a4paper,11pt]{article}
\PassOptionsToPackage{hyperfootnotes=false,pdfpagemode=UseNone,bookmarksopen=false,colorlinks=true,urlcolor=blue,citecolor=myorange,citebordercolor=blue,linkcolor=blue}{hyperref}
\PassOptionsToPackage{hyphens}{url}
\usepackage[utf8]{inputenc} 
\usepackage[T1]{fontenc}
\usepackage{microtype}

\usepackage{stmaryrd}
\SetSymbolFont{stmry}{bold}{U}{stmry}{m}{n}

\usepackage{xcolor}

\usepackage{graphicx}
\usepackage{hyperref}
\usepackage{amsmath,tikz,stmaryrd,float,amsthm,amssymb,mathrsfs}
\usetikzlibrary{arrows.meta,positioning,calc,fit}
\usepackage{mathtools}
\usepackage[nameinlink, capitalize, noabbrev]{cleveref}
\usepackage{mathdots}
\usepackage{etoolbox}
\usepackage{thmtools}
\usepackage{array}
\usepackage{bm}

\usepackage{layout-arxiv}
\declaretheoremstyle[
  spaceabove=\parskip,
  spacebelow=\parskip,
  headfont=\normalfont\bfseries,
  bodyfont=\normalfont,
  headpunct={.},
  notefont=\normalfont\bfseries,
  headformat=\NAME~\NUMBER\NOTE
]{parskipplain}

\declaretheoremstyle[
  spaceabove=\parskip,
  spacebelow=\parskip,
  headfont=\normalfont\bfseries,
  bodyfont=\normalfont\itshape,
  headpunct={.},
  notefont=\normalfont\bfseries
]{parskipremark}

\declaretheorem[style=parskipremark, name=Proposition]{proposition}
\declaretheorem[style=parskipremark, name=Theorem, sibling=proposition]{theorem}
\declaretheorem[style=parskipremark, name=Corollary, sibling=proposition]{corollary}
\declaretheorem[style=parskipremark, name=Claim, sibling=proposition]{claim}
\declaretheorem[style=parskipremark, name=Informal Theorem, numbered=no]{informaltheorem}
\declaretheorem[style=parskipplain, name=Lemma, within=section]{lemma}

\allowdisplaybreaks

\DeclareUnicodeCharacter{00A0}{~}

\newcommand{\R}{\mathbb{R}}
\renewcommand{\d}{\mathrm{d}}
\newcommand{\N}{\mathcal{N}}
\newcommand{\1}{\bm{1}}
\renewcommand{\L}{\mathcal{L}}
\newcommand{\tr}{\mathrm{Tr}}

\newcommand{\bU}{\mathbf{U}}
\newcommand{\bV}{\mathbf{V}}
\newcommand{\bC}{\mathbf{C}}
\newcommand{\bSigma}{\mathbf{\Sigma}}
\newcommand{\bW}{\mathbf{W}}
\newcommand{\bY}{\mathbf{Y}}
\newcommand{\bZ}{\mathbf{Z}}
\newcommand{\bA}{\mathbf{A}}
\newcommand{\bM}{\mathbf{M}}
\newcommand{\bN}{\mathbf{N}}
\newcommand{\bR}{\mathbf{R}}
\newcommand{\bLambda}{\mathbf{\Lambda}}
\newcommand{\bQ}{\mathbf{Q}}
\newcommand{\bO}{\mathbf{O}}
\newcommand{\bB}{\mathbf{B}}
\newcommand{\bK}{\mathbf{K}}
\newcommand{\bH}{\mathbf{H}}
\newcommand{\bP}{\mathbf{P}}
\newcommand{\bGamma}{\mathbf{\Gamma}}
\newcommand{\bD}{\mathbf{D}}
\newcommand{\bX}{\mathbf{X}}
\newcommand{\bS}{\mathbf{S}}
\newcommand{\bG}{\mathbf{G}}
\newcommand{\bF}{\mathbf{F}}
\newcommand{\bPi}{\mathbf{\Pi}}
\newcommand{\bI}{\mathbf{I}}
\newcommand{\bImb}{\mathbf{Imb}}
\newcommand{\bPhi}{\mathbf{\Phi}}
\newcommand{\bE}{\mathbf{E}}
\newcommand{\bzero}{\mathbf{0}}
\newcommand{\bx}{\mathbf{x}}
\newcommand{\by}{\mathbf{y}}
\newcommand{\bu}{\mathbf{u}}
\newcommand{\bv}{\mathbf{v}}
\newcommand{\bq}{\mathbf{q}}
\newcommand{\bz}{\mathbf{z}}
\newcommand{\be}{\mathbf{e}}
\newcommand{\bo}{\mathbf{o}}
\newcommand{\bsigma}{\boldsymbol{\sigma}}

\newtheoremstyle{assumpstyle}{\parskip}{3pt}{\normalfont}{}{\bfseries}{}{0.5em}
    {\thmname{#1}~\thmnumber{#2}.\newline \vspace*{-8pt}\\}

\theoremstyle{assumpstyle}
\newtheorem{assumption}{Assumption}

\newcounter{subassumption}[assumption]
\renewcommand{\thesubassumption}{\theassumption.\arabic{subassumption}}



\crefname{subassumption}{Assumption}{Assumptions}

\makeatletter
\newenvironment{assumpitem}[1][]{
  \refstepcounter{subassumption}
  \edef\@currentlabel{\thesubassumption}
  \noindent\textbf{\thesubassumption\mbox{.\hspace{0.4em}#1}}
}{\par}
\makeatother

\crefname{assumption}{Assumption}{Assumptions}


\hypersetup{pdftitle={Silent Alignment in Two-Layer Linear Networks: A High-Dimensional Perspective}, pdfauthor={Simon Martin, Francis Bach, Giulio Biroli}}
\newcommand{\alttext}[1]{}

\newenvironment{mapfigure}{\begin{figure}[tbp]}{\end{figure}}

\usepackage{natbib}
\graphicspath{{Code/figures/pdf/}}

\makeatletter

\newcommand{\tocsection}[1]{\arxivsection*{#1}\phantomsection\addcontentsline{toc}{section}{#1}}
\newcommand{\sectionstarintoc}{\let\arxivsection\section\renewcommand{\section}{\@ifstar{\tocsection}{\arxivsection}}}
\makeatother

\begin{document}
\title{\textbf{Silent Alignment in Two-Layer Linear Networks: \\ A High-Dimensional Perspective}}

\date{}

\author[1,2]{Simon Martin\thanks{To whom correspondence should be sent: \texttt{smartin@di.ens.fr}.}}
\author[1]{Francis Bach}
\author[2]{Giulio Biroli}

\affil[1]{INRIA - \'Ecole Normale Sup\'erieure, PSL Research University, Paris, France.}
\affil[2]{Laboratoire de Physique de l’\'Ecole Normale Sup\'erieure, ENS, Universit\'e PSL, CNRS, Sorbonne Universit\'e, Universit\'e de Paris, F-75005 Paris, France.}

\maketitle

\begin{abstract}
We study gradient flow in a two-layer linear network in the high-dimensional regime where the input, inner and output dimensions diverge proportionally. Under whitened inputs and isotropic imbalance, we combine a closed-form solution of the dynamics with prior random matrix theory results to derive a high-dimensional limit for the dynamics. Our method yields exact asymptotic spectral distributions for the weight matrices at all times and characterizes the evolution of the learned subspace. In the small-initialization limit, we identify a silent alignment phenomenon: the singular directions of the network align on an order-one timescale, while the singular values and empirical loss remain at their initial values. Visible learning occurs later, on a logarithmic timescale, as singular values progressively escape zero and the loss decreases. This phase is characterized by a saddle-to-saddle dynamics, which becomes a continuous process of spectral learning in the high-dimensional limit.
\end{abstract}

\tableofcontents

\newpage

\section{Introduction} \label{Sec:Intro}
Neural networks learn representations from data rather than relying only on fixed features \citep{bengio2013representation}. Understanding this process requires describing how training selects useful directions, how those directions evolve, and when their evolution begins to improve prediction. The nature of this feature evolution, and the tools available to analyze it, depend on the scaling regime under consideration.

\paragraph{Regimes of feature learning.} A line of work that has successfully characterized feature learning in neural networks is the study of single and multi-index models, associated with target functions depending on a low-dimensional feature space. Several works study how gradient-based methods recover the relevant directions in such nonlinear models \citep{damian2022representations,bietti2022single,bietti2023multi}. When the number of samples grows proportionally with the input dimension, dynamical studies of finite-width shallow networks rely on low-dimensional order parameters to obtain phase diagrams and study signal recovery \citep{saad1995online,goldt2019dynamics,refinetti2021classifying,arnaboldi2023high,montanari2026phase}.

Going beyond these low-dimensional representations is still a challenge. Indeed, in large-width limits or random feature models, features often remain frozen during training \citep{rahimi2007random,jacot2018neural,chizat2019lazy,mei2022generalization}. In the same regime, mean-field and other large-width parameterizations allow for non-trivial feature evolution \citep{chizat2018global,yang2021tensor,bordelon2022self}, but their nonlinear dynamics are more challenging to describe. 

A complementary line of work studies joint limits in which the network width grows with the input dimension, allowing for feature learning with a macroscopic number of features. For quadratic neural networks, these limits have led to precise dynamical descriptions \citep{martin2024impact,ben2026learning,martin2026high}, as for the closely related problem of matrix denoising \citep{bodin2023gradient}. Beyond quadratic networks, related scalings have been studied for deep linear networks \citep{bordelon2025deep,lauditi2026spectral} and for two-layer networks after a few gradient steps \citep{ba2022high,moniri2026feature}. We focus throughout the paper on this proportional setting, often referred to as the extensive-width regime.

\paragraph{Linear networks and factorized models.} Factorized linear models have motivated a long line of work, as they produce nonlinear but tractable dynamics that exhibit complex phenomena also observed in modern neural networks. These models include two-layer and deep linear networks, diagonal networks and low-rank matrix factorization. They have been used to study convergence and alignment of the successive layers \citep{bah2022learning,ji2018gradient}, conservation laws \citep{marcotte2023abide}, and the implicit bias toward low-rank solutions \citep{gissin2019implicit,gidel2019implicit,gunasekar2017implicit,arora2019implicit}. Other works reveal saddle-to-saddle dynamics and successive low-rank or sparse learning \citep{jacot2021saddle,pesme2023saddle,li2021greedy,chou2024gradient,kunin2025alternating,min2025understanding}. At finite stepsize, these models can also exhibit edge-of-stability and oscillatory regimes \citep{even2023s,ghosh2025learning}.

In two-layer linear networks, training is known to be exactly tractable under structured initialization \citep{fukumizu1998effect,braun2022exact}, leading to an Oja flow \citep{oja1982simplified,yan1994global} that decouples along the singular values for a spectral initialization \citep{saxe2013exact,tarmoun2021understanding,domine2025lazy}. In the overparameterized setting, \citet{varre2023spectral} studied the implicit bias of the gradient flow dynamics and showed that, for whitened inputs and a specific initialization, the singular vectors remain frozen during training, while singular values escape zero following a discrete jump process in the small-initialization regime \citep[see also][]{gidel2019implicit,bach2026closed}. 

Small initialization can also produce an early phase in which the learned directions align while the weights remain small, in linear networks and matrix sensing \citep{stoger2021small,soltanolkotabi2023implicit,tu2024mixed}, as well as in ReLU networks \citep{min2024early,boursier2025early}. Close to our setting, \citet{atanasov2021neural} called this phase \emph{silent alignment} when the loss also stays at its initial value, and analyzed it in linear networks through the evolution of the neural tangent kernel. \Citet{braun2022exact} derived it in a simple two-dimensional setting from an exact solution, \citet{karkada2025closed} in symmetric matrix factorization, and \citet{zhang2026saddle} described a similar early phase in linear networks with vector outputs, through the dynamics linearized near initialization. In high dimension, \citet{lampinen2019analytic} showed that the singular values are learned as a wave sweeping the spectrum, but only from an initialization already aligned with the data. A challenge we address here is to characterize the whole dynamics from a random initialization in the extensive-width regime, where a macroscopic number of directions take part in both the alignment and the learning stages.

\paragraph{Contributions.} We consider a two-layer linear network trained by gradient flow on the square loss, with whitened inputs. Our contributions are twofold:
\begin{itemize}
    \item \textbf{High-dimensional dynamics.} By combining a known closed-form solution of the dynamics with prior random matrix theory results, we obtain a high-dimensional description of the training trajectory in the extensive-width regime, which equivalently describes the factorization of a matrix of extensive rank. At all times, we characterize the spectral distribution of the prediction matrix and quantify how the learned subspace moves, through explicit limits for distances between range projectors. This allows us to follow feature learning throughout training.
    
    \item \textbf{Silent alignment.} We make the silent alignment mechanism explicit from a random initialization, both in finite dimension and in the high-dimensional regime where the network learns an extensive number of features. In the small-initialization limit, the singular directions of the prediction align on an order-one timescale, while its singular values and loss remain at their initial values. The singular values then escape progressively on a later timescale and the loss starts to decrease. In finite dimension, this leads to a saddle-to-saddle dynamics that becomes a continuous spectral learning process in the high-dimensional limit.
\end{itemize}

\section{Setting} \label{Sec:Setting}

In this section, we introduce the model we are interested in, as well as our main assumptions. We then state an informal version of our main result.

\subsection{Notation and Conventions} \label{Subsec:NotationConventions}

For a matrix $\bA \in \R^{d \times m}$, we denote by $\bA^\top$ its transpose, $\tr(\bA)$ its trace, $\|\bA\|_F = \big[\tr(\bA\bA^\top) \big]^{1/2}$ its Frobenius norm, $\mathrm{Im}(\bA)$ its range and $\mathrm{Ker}(\bA)$ its kernel. We denote by $\mathcal{S}_d(\R), \mathcal{S}_d^+(\R)$ and $\mathcal{S}_d^{++}(\R)$ the sets of $d \times d$ symmetric, positive semidefinite (PSD) and positive definite matrices. For $\bA, \bB \in \mathcal{S}_d(\R)$, we write $\bA \succeq \bB$ when $\bA - \bB$ is PSD, and equivalently write $\bB \preceq \bA$. Whenever $m \leq d$, we define the Stiefel manifold:
\begin{equation*}
    \mathrm{St}(d, m) = \Big\{ \bR \in \R^{d \times m}, \, \bR^\top \bR = \bI_m \Big\}. 
\end{equation*}
Given $E$ a Euclidean space and $L \colon E \to \R$ continuously differentiable, we let $\nabla L(\bx) \in E$ be its gradient at $\bx \in E$. When $L$ is a function of several variables, we denote by $\nabla_{\bx} L$ the gradient with respect to the variable $\bx$.

If $\bA \in \mathcal{S}_d(\R)$ has spectral decomposition $\bA=\sum_i \lambda_i \bq_i\bq_i^\top$, then for a scalar function $f$ we write $f(\bA) = \sum_i f(\lambda_i) \bq_i\bq_i^\top$. For a rectangular matrix $\bM$ with singular value decomposition (SVD) $\bM = \sum_i \sigma_i \bu_i \bv_i^\top$, we similarly write $f(\bM)=\sum_i f(\sigma_i)\bu_i\bv_i^\top$. 

\subsection{Model} \label{Subsec:Model}

In the following we consider a linear network of the form $\bx \in \R^d \mapsto \bU \bV^\top \bx \in \R^p$, where $\bU \in \R^{p \times m}$ and $\bV \in \R^{d \times m}$. From a dataset $\mathcal{D}_n = \big\{ (\bx_k, \by_k) \big\}_{1 \leq k \leq n}$ where $\bx_1, \dots, \bx_n \in \R^d$ and $\by_1, \dots, \by_n \in \R^p$, we aim at minimizing the loss function:
\begin{equation} \label{eq:DefLoss}
    \L(\bU,\bV) = \frac{1}{2n} \sum_{k=1}^n \big\| \bU \bV^\top \bx_k - \by_k \big\|_2^2.
\end{equation}
Up to an additive constant, we have:
\begin{equation*}
    \L(\bU, \bV) = \frac{1}{2} \tr \big( \bU^\top \bU \bV^\top \bSigma \bV \big) - \tr \big( \bU \bV^\top \bC^\top \big),
\end{equation*}
with
\begin{equation*}
    \bSigma = \frac{1}{n} \sum_{k=1}^n \bx_k\bx_k^\top \in \mathcal{S}_d^+(\R), \hspace{1.5cm} \bC = \frac{1}{n} \sum_{k=1}^n \by_k \bx_k^\top \in \R^{p \times d}. 
\end{equation*}
$\bSigma$ is the second-moment matrix of the inputs $(\bx_1, \dots, \bx_n)$ and $\bC$ is the input-output cross-moment of the dataset. We optimize the loss function $\L$ using gradient flow on the matrices $\bU, \bV$:
\begin{equation*}
    \dot \bU(t) = - \nabla_{\bU} \L \big( \bU(t), \bV(t) \big), \hspace{1.5cm} \dot \bV(t) = - \nabla_{\bV} \L \big( \bU(t), \bV(t) \big).
\end{equation*}
The dynamics then reads:
\begin{equation} \label{eq:GFDynamics}
\begin{aligned}
    \dot \bU(t) &= - \big( \bU(t) \bV(t)^\top \bSigma - \bC \big) \bV(t), \\
    \dot \bV(t) &= - \big( \bSigma \bV(t) \bU(t)^\top - \bC^\top \big) \bU(t). 
\end{aligned}
\end{equation}
This leads to a nonlinear system of coupled matrix-differential equations. It is known \citep{arora2018optimization,du2018algorithmic} that the matrix $\bImb(t) = \bV(t)^\top \bV(t) - \bU(t)^\top \bU(t)$ remains constant along the flow and is therefore determined by the initialization $\bU_0 := \bU(t= 0)$ and $\bV_0 := \bV(t = 0)$. Several works \citep{braun2022exact,varre2023spectral,domine2025lazy} have studied the dynamics under balanced or structured imbalance assumptions. In what follows, we call $\bImb(t)$ the \textit{imbalance matrix}. 

\subsection{Main Assumptions} \label{Subsec:Assumptions}

In this paper, we work under the following assumptions:
\begin{assumption} \label{Assumption1}
\begin{assumpitem}[Whitened data.] \label{Assumption1:Covariance} The second-moment matrix of the inputs satisfies $\bSigma = \bI_d$. In particular, this implies the large-data regime $n \geq d$. 
\end{assumpitem}
\begin{assumpitem}[High-dimensional scaling.]\label{Assumption1:Scaling} The inner and output dimensions scale proportionally to the input dimension $d$ as it diverges to infinity: 
\begin{equation}
    m \underset{d \to \infty}{\sim} \kappa d, \hspace{1.5cm} p \underset{d \to \infty}{\sim} \tau d,
\end{equation}
for some fixed constants $\kappa \in (0,1)$ and $\tau > 0$. 
\end{assumpitem}
\begin{assumpitem}[Cross-moment matrix.] \label{Assumption1:Teacher} The empirical measure of the singular values of $\bC \in \R^{p \times d}$
\begin{equation*}
    \mu_C^d = \frac{1}{\min(p,d)} \sum_{i=1}^{\min(p,d)} \delta_{\sigma_i},
\end{equation*}
weakly converges as $d \to \infty$ to a compactly supported probability measure $\mu_C$. Moreover, $\mu_C$ admits the decomposition:
\begin{equation*}
    \mu_C(\d x) = (1 - \kappa^*) \delta_0(\d x) + \kappa^* \phi_C(x) \d x,
\end{equation*}
for $\kappa^* \in (0,1]$. We assume that $\phi_C$ is a bounded density with compact support bounded away from zero. We denote $x_0 = \inf \, \mathrm{Supp}(\phi_C) > 0$. Additionally, there exist constants $0<c<L$, independent of $d$, such that every non-zero singular value of $\bC$ belongs to $[c,L]$ for all $d \geq 1$.
\end{assumpitem}
\begin{assumpitem}[Initialization.] \label{Assumption1:Initialization} The dynamics~\eqref{eq:GFDynamics} is initialized with $\bU_0 = \bzero$ and $\bV_0 = \sqrt{2\rho} \bR$, for some $\rho > 0$ and $\bR \in \R^{d \times m}$ uniformly drawn on the Stiefel manifold $\mathrm{St}(d, m)$.
\end{assumpitem}
\end{assumption}

We now briefly comment on these assumptions and give several consequences.
\begin{itemize}
    \item \emph{Loss and generalization.} Under \cref{Assumption1:Covariance}, the empirical loss can be written, up to an additive constant, as
    \begin{equation} \label{eq:LossDistance}
        \L(\bU,\bV) = \frac{1}{2}\|\bU \bV^\top - \bC\|_F^2 + \mathrm{constant}.
    \end{equation}
    The constant is the part of the empirical loss that cannot be fitted by a linear map of the inputs. Thus, the squared Frobenius norm is twice the excess empirical loss. Additionally, our analysis studies how the network learns the dataset through $\bC$, rather than its generalization performance. Studying generalization would require an additional model for the data beyond the empirical cross-moment matrix.
    \item \emph{Extensive-width regime.} Together, \cref{Assumption1:Scaling,Assumption1:Teacher} define an extensive-width regime \citep{maillard2024bayes,martin2026high}: the input, inner, and output dimensions grow proportionally, and the cross-moment matrix $\bC$ possesses a macroscopic number of non-zero singular directions to be learned. 
    \item \emph{Overparameterized and underparameterized regimes.} \cref{Assumption1:Teacher} implies that the rank of $\bC$ (that we denote $m^*$ in \cref{Sec:FiniteDim}) is asymptotically $\beta \kappa^* d$, where $\beta = \min(1, \tau)$. Throughout the paper, we define the terms \emph{underparameterized} and \emph{overparameterized} relative to this rank: in finite dimension they correspond to $m < m^*$ and $m > m^*$, while in the high-dimensional limit they correspond to $\kappa < \beta \kappa^*$ and $\kappa > \beta \kappa^*$ respectively. We do not treat the critical high-dimensional case $\kappa=\beta\kappa^*$, which would require a separate analysis.
    \item \emph{Structured initialization.} \cref{Assumption1:Initialization} is similar to the one used by \citet{varre2023spectral}, except that they work in the regime where $m \geq d$. This assumption implies that the imbalance matrix is given by $\bImb(t) = 2 \rho \bI_m$ at all times. This setting of isotropic imbalance has also been studied by \citet{tarmoun2021understanding,domine2025lazy}. 
    \item \emph{Matrix factorization.} Under \cref{Assumption1:Covariance}, the expression of the loss in equation \eqref{eq:LossDistance} shows that the dynamics \eqref{eq:GFDynamics} is also the gradient flow of the asymmetric factorization of $\bC$, so that our results also apply to matrix factorization with a target of extensive rank.
 \end{itemize}

\subsection{Overview of the Results} \label{Subsec:InformalTheorem}

The following informal theorem summarizes the main result of the paper: in the small-initialization limit, learning takes place on two distinct timescales.

\begin{informaltheorem}
    Consider the gradient flow \eqref{eq:GFDynamics} under \cref{Assumption1}. In the high-dimensional limit and for small initialization $\rho$, the dynamics separates into two phases:
    \begin{description}
        \item[$\bullet$ Alignment phase.] At $t = \Theta(1)$ as $\rho \to 0$, the loss remains at its initial value and the prediction $\bU(t) \bV(t)^\top$ remains near zero, but its singular subspaces become correlated with those of $\bC$.
        \item[$\bullet$ Learning phase.] At $t = \Theta \big( \log(\rho^{-1}) \big)$, the singular values of $\bU(t) \bV(t)^\top$ escape zero to match those of $\bC$. The loss progressively decreases, and after a time $t = s \log(\rho^{-1})$, the prediction has recovered the singular values of $\bC$ above the threshold 
        \begin{equation*}
            x_s = \frac{1}{2s},
        \end{equation*}
        as long as the rank of the prediction allows it.
\end{description}
\end{informaltheorem}

\cref{fig:Map} summarizes in more detail the contributions of the paper. 

\begin{mapfigure}
    \centering
    \newcommand{\mapsplitblock}[5]{
        \begin{scope}
            \clip[rounded corners=3pt] ({#2-0.5*#4},{#3-0.5*#5}) rectangle ({#2+0.5*#4},{#3+0.5*#5});
            \fill[blue!7] ({#2-0.5*#4},{#3}) rectangle ({#2+0.5*#4},{#3+0.5*#5});
            \fill[orange!14] ({#2-0.5*#4},{#3-0.5*#5}) rectangle ({#2+0.5*#4},{#3});
        \end{scope}
        \draw[black!40, line width=0.4pt] ({#2-0.5*#4},{#3}) -- ({#2+0.5*#4},{#3});
        \node[draw=black!40, rounded corners=3pt, minimum width=#4cm, minimum height=#5cm, inner sep=0pt] (#1) at (#2,#3) {};
    }
    \newcommand{\mapgroup}[3]{%
        \draw[black!45, line width=0.6pt] (1.0,#2) -- (0.85,#2) -- (0.85,#3) -- (1.0,#3);
        \node[font=\sffamily\small, text=black!75, rotate=90] at (0.5,{0.5*(#2+#3)}) {#1};
    }
    \resizebox{\linewidth}{!}{%
    \begin{tikzpicture}[
        font=\sffamily,
        >={Latex[length=2.2mm]},
        flow/.style={->, line width=0.8pt, draw=black!60, shorten <=2pt, shorten >=2pt},
        stagecontent/.style={align=center, text width=3.8cm, inner sep=0pt,
            font=\sffamily\footnotesize},
        widecontent/.style={align=flush center, text width=8.4cm, inner sep=0pt,
            font=\sffamily\footnotesize},
        stageref/.style={align=center, text width=3.8cm, inner sep=0pt,
            font=\sffamily\scriptsize},
        header/.style={font=\sffamily\bfseries\small, align=center}
    ]
        \fill[blue!7, draw=black!40] (4.6,0.45) rectangle (4.9,0.2);
        \node[font=\sffamily\footnotesize, anchor=west] at (5.0,0.325) {finite dimension $d$};
        \fill[orange!14, draw=black!40] (8.0,0.45) rectangle (8.3,0.2);
        \node[font=\sffamily\footnotesize, anchor=west] at (8.4,0.325) {high-dimensional limit $d\to\infty$};

        \mapgroup{any $\rho$}{-0.75}{-3.35}
        \node[header] at (5.7,-0.4) {All times};

        \mapsplitblock{allt}{5.7}{-2.05}{8.9}{2.6}
        \node[widecontent, anchor=north] at ([yshift=-2.5mm]allt.north)
        {Closed-form solution of the dynamics};
        \node[stageref, text width=8.4cm, font=\sffamily\footnotesize, anchor=south] at ([yshift=1.5mm]allt.center)
        {\cref{Prop:SolutionOja,Prop:SVDPrediction}};
        \node[widecontent, anchor=north] at ([yshift=-2.5mm]allt.center)
        {Prediction spectrum and subspace alignment};
        \node[stageref, text width=8.4cm, font=\sffamily\footnotesize, anchor=south] at ([yshift=1.5mm]allt.south)
        {\cref{Thm:Stieltjes,Thm:Projector1}};

        \mapgroup{small initialization ($\rho \to 0$)}{-4.3}{-8.2}
        \node[header] at (3.25,-3.95) {Alignment};
        \node[header] at (8.15,-3.95) {Learning};

        \mapsplitblock{align}{3.25}{-6.25}{4.0}{3.9}
        \node[stagecontent, anchor=north] at ([yshift=-3mm]align.north)
        {Top-$k$ singular spaces \\
        align at rate \\
        $e^{-2t(\sigma_k-\sigma_{k+1})}$};
        \node[stageref, anchor=south] at ([yshift=2mm]align.center)
        {\cref{Thm:TopEigenvectors}};
        \node[stagecontent, anchor=north] at ([yshift=-3mm]align.center)
        {Top-$\alpha d$ singular spaces \\
        align at power-law rate};
        \node[stageref, anchor=south] at ([yshift=2mm]align.south)
        {\cref{Thm:ProjectorLimit,Thm:ProjectorLimit2,Thm:TopEigenvectors}};

        \mapsplitblock{learn}{8.15}{-6.25}{4.0}{3.9}
        \node[stagecontent, anchor=north] at ([yshift=-3mm]learn.north)
        {\emph{Saddle-to-saddle dynamics} \\
        Amplitude $\sigma_i$ learned\\
        at $s=(2\sigma_i)^{-1}$};
        \node[stageref, anchor=south] at ([yshift=2mm]learn.center)
        {\cref{Thm:FiniteDimensionLogTimes}};
        \node[stagecontent, anchor=north] at ([yshift=-3mm]learn.center)
        {\emph{Continuous spectral front} \\
        Singular values above\\
        $x_s=(2s)^{-1}$ are learned};
        \node[stageref, anchor=south] at ([yshift=2mm]learn.south)
        {\cref{Thm:StieltjesLimit,Thm:LossTimescales}};

        \node[header] at (13.05,-1.6) {Endpoint {\normalfont\footnotesize (any $\rho$)}};
        \mapsplitblock{end}{13.05}{-4.15}{4.0}{4.4}
        \node[stagecontent, anchor=north] at ([yshift=-3mm]end.north)
        {Prediction converges \\
        to the $\min(m,m^*)$ largest \\
        singular modes of $\bC$};
        \node[stageref, anchor=south] at ([yshift=2mm]end.center)
        {\cref{Prop:LimitBlockMatrices}};
        \node[stagecontent, anchor=north] at ([yshift=-3mm]end.center)
        {Target density learned \\
        above the spectral \\
        threshold $x_*$};
        \node[stageref, anchor=south] at ([yshift=2mm]end.south)
        {\cref{Cor:StieltjesInf}};

        \draw[flow] (align) -- (learn);
        \draw[flow] ([yshift=3mm]allt.south east) -- ([yshift=1.5mm]end.west);
        \draw[flow] ([yshift=-9.5mm]learn.north east) -- ([yshift=-1.5mm]end.west);

        \draw[->, line width=0.9pt, draw=black!75] (1.15,-8.75) -- (15.15,-8.75);
        \foreach \x/\lab in {3.25/{$t=\Theta(1)$}, 8.15/{$t=s\log(\rho^{-1})$}, 13.05/{$t\to\infty$}} {
            \draw[line width=0.7pt, draw=black!75] (\x,-8.68) -- (\x,-8.82);
            \node[font=\sffamily\footnotesize, anchor=north] at (\x,-8.88) {\lab};
        }
        \node[font=\sffamily\footnotesize, anchor=west] at (15.1,-8.75) {$t$};
    \end{tikzpicture}%
    }
    \caption{Map of the results, with time increasing from left to right. The upper block gathers the results that hold for any $\rho > 0$ at all times. The two lower blocks consider the small-initialization limit $\rho \to 0$, where the dynamics separates into an alignment stage and a learning stage. Both lead to the same long-time limit, which holds for any $\rho$ (right). In each block, the top part gives the finite-dimensional result and the bottom part the high-dimensional limit $d \to \infty$, taken before $\rho \to 0$.}
    \alttext{Diagram summarizing the results along a time axis. For any initialization scale, a closed-form solution and its high-dimensional limit describe all times. For small initialization, the subspaces align first, then the singular values are learned one by one, or as a continuous front in high dimension. Both lead to the same long-time limit, where only the largest singular values are learned.}
    \label{fig:Map}
\end{mapfigure}

\begin{figure}[htbp]
    \centering
    \includegraphics[width=\linewidth]{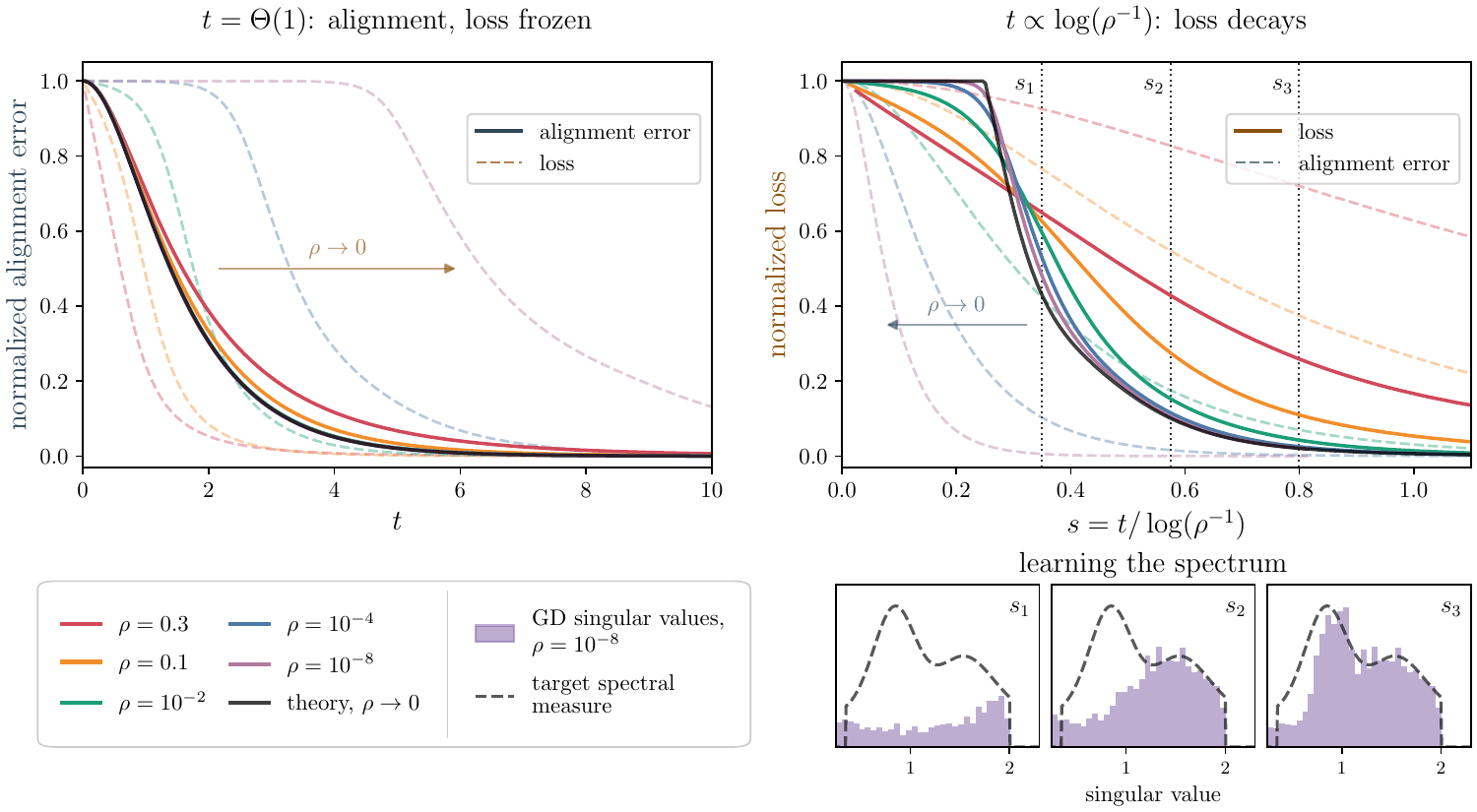}
    \caption{Colored curves are finite-dimensional gradient descent simulations of the dynamics \eqref{eq:GFDynamics} at positive $\rho$, averaged over 10 repetitions, with $d=500$, stepsize $\eta=10^{-3}$, $\tau=1$, $\kappa=0.75$, and $\kappa^*=0.55$. Both quantities are divided by their initial values. \emph{Top-left panel:} alignment error (solid) on the timescale $t=\Theta(1)$ for decreasing values of $\rho$, with the $\rho\to0$ prediction of \cref{Thm:ProjectorLimit} in black. Dashed curves show the loss on the same timescale, which starts to decrease later as $\rho$ decreases and stays at its initial value as $\rho \to 0$. \emph{Top-right panel:} loss (solid) on the scale $s=t/\log(\rho^{-1})$, with its $\rho \to 0$ limit from \cref{Thm:LossTimescales} in black. Dashed curves show the alignment error on this scale, whose $\rho \to 0$ limit vanishes for any $s>0$. \emph{Bottom-right panels:} histograms of the singular values of $\bU(t)\bV(t)^\top$ for $\rho=10^{-8}$ at the times $s_1,s_2,s_3$, normalized by the total number of singular values, compared with the target $\kappa^*\phi_C$: the singular values of $\bC$ are learned from the largest ones downwards.}
    \alttext{Alignment error, loss and learned singular values along training, for several initialization scales. Top left: the alignment error decreases on a time of order one in nearly the same way for all scales, while the loss stays at its initial value longer for smaller scales. Top right: against rescaled time, the loss stays at its initial value and then decreases, close to the black limit curve for the smallest scales, while the alignment error has already dropped. Bottom right: histograms of the learned singular values at three increasing times, filling the target density from its largest values downwards.}
    \label{fig:Fig1}
\end{figure}

The two phases are visualized in \cref{fig:Fig1}. The alignment error is the normalized right range-projector distance introduced in \cref{Subsec:RangeProjectors}, while the loss is the normalized squared distance between $\bU(t)\bV(t)^\top$ and $\bC$, equivalently the excess empirical loss by equation \eqref{eq:LossDistance}. On the timescale $t=\Theta(1)$, the finite-dimensional alignment curves collapse onto the $\rho\to0$ prediction of \cref{Thm:ProjectorLimit}, whereas the loss remains essentially constant. On the scale $t=s\log(\rho^{-1})$, the loss curves collapse instead as $\rho \to 0$, and their decay as a function of $s$ is directly linked with the progressive learning of the singular values of $\bC$ starting from the largest ones. 

The following sections are dedicated to our results. They are organized as follows:
\begin{itemize}
    \item In \cref{Sec:FiniteDim}, we analyze the gradient flow dynamics in finite dimension. We recall an explicit closed-form solution for the dynamics \citep[originally due to][]{fukumizu1998effect} and study its convergence at long times.
    \item In \cref{Sec:HighDim}, we analyze the dynamics in the high-dimensional limit. We derive deterministic spectral distributions for the matrices associated with the dynamics and quantify the evolution of the learned subspace through projector distances. 
    \item In \cref{Sec:SilentAlignment}, we consider the small-initialization limit $\rho \to 0$ and derive the two timescales of the dynamics, which correspond to a silent alignment phenomenon \citep{atanasov2021neural}.
\end{itemize}

A few of our results already exist in the literature or are closely related to prior works. We recall them as they are essential for subsequent new results and lead to a more complete picture of the phenomena we describe. Whenever this is the case, we make it explicit and provide the relevant references.

The code for all simulations, figures and numerical checks, together with animations of the learning dynamics, is available in a \href{https://github.com/simonmartin15/silent-alignment-linear-networks}{GitHub repository}. More detail on the numerical simulations can be found in \cref{App:Simulations}.

\section{Finite-Dimensional Dynamics} \label{Sec:FiniteDim}

In this section, we study the gradient flow dynamics \eqref{eq:GFDynamics} in finite dimension.
\begin{itemize}
    \item In \cref{Subsec:ClosedForm}, we recall a closed-form solution for the gradient flow dynamics.
    \item In \cref{Subsec:PredictionMatrix}, we show that in our setting, the prediction matrix $\bU(t) \bV(t)^\top$ can be understood from the Gram matrices $\bU(t) \bU(t)^\top$, $\bV(t) \bV(t)^\top$. 
    \item In \cref{Subsec:LimitDynamics}, we compute the long-time limit of the Gram and prediction matrices.
\end{itemize}

\subsection{Closed-Form Solution} \label{Subsec:ClosedForm}

The first result we present is a closed-form solution for the dynamics, which shows that the gradient flow reduces to an Oja flow \citep{oja1982simplified}.

\begin{proposition}[Oja flow representation, \citealp{fukumizu1998effect,tarmoun2021understanding}] \label{Prop:SolutionOja}
    Consider the solution $\bU(t), \bV(t)$ of the gradient flow dynamics \eqref{eq:GFDynamics}. Then, under \cref{Assumption1:Covariance,Assumption1:Initialization}, the matrix
    \begin{equation*}
        \bW(t) = \begin{pmatrix} \bU(t) \\ \bV(t)
        \end{pmatrix} \in \R^{(p+d) \times m},
    \end{equation*}
    is a solution of the matrix differential equation:
    \begin{equation} \label{eq:ODEW}
        \dot \bW(t) = - \frac{1}{2} \bW(t) \bW(t)^\top \bW(t) + \bA_{\rho} \bW(t),
    \end{equation}
    where
    \begin{equation*}
        \bA_\rho = \begin{pmatrix} -\rho \bI_p & \bC \\ \bC^\top & \rho \bI_d
        \end{pmatrix} \in \mathcal{S}_{p+d}(\R). 
    \end{equation*}
    As a consequence, we have the closed-form solution for all $t \geq 0$:
    \begin{equation} \label{eq:SolutionOja}
        \bW(t) \bW(t)^\top = e^{t \bA_\rho} \bW_0 \left( \bI_m + \bW_0^\top \int_0^t e^{2s\bA_\rho} \d s \, \bW_0 \right)^{-1} \bW_0^\top e^{t \bA_\rho}. 
    \end{equation}
\end{proposition}

\citet{fukumizu1998effect} obtained this result in the case of a balanced initialization, and \citet{tarmoun2021understanding,domine2025lazy} extended it to an imbalance proportional to the identity.
For completeness we prove this result in \cref{App:Subsec:ProofPropSolutionOja}. Additionally, in \cref{App:Subsec:BlockExpressions}, we derive explicit expressions for the Gram matrices $\bU(t) \bU(t)^\top$, $\bV(t) \bV(t)^\top$ and the prediction $\bU(t) \bV(t)^\top$ from the above expression of $\bW(t) \bW(t)^\top$. 

The closed-form expression in equation \eqref{eq:SolutionOja} forms the basis of our analysis and allows us to study this dynamics in the high-dimensional limit, under the scaling of \cref{Assumption1:Scaling}. This result turns the high-dimensional dynamics into a random matrix problem involving $\bW_0$ and $\bA_\rho$. The matrix $\bA_\rho$ can be interpreted as a $\rho$-deformation of the Hermitian dilation of $\bC$ \citep[see][Definition 2.1.5]{tropp2015introduction}, therefore its spectral structure follows directly from that of $\bC$.

Equation \eqref{eq:ODEW} is an Oja flow. This nonlinear matrix dynamics is classical in finite dimension \citep{yan1994global,tarmoun2021understanding,tsuzuki2025global}, and closely related flows appear in matrix denoising and quadratic neural networks, including in the extensive-width regime \citep{bodin2023gradient,martin2024impact,martin2026high}. We derive new results for the present setting, and our techniques can also be applied to obtain a more precise understanding of the Oja flow in high dimension. 

\subsection{The Prediction Matrix} \label{Subsec:PredictionMatrix}

We are now interested in the prediction matrix $\bZ(t) := \bU(t) \bV(t)^\top$, as the loss function in equation \eqref{eq:DefLoss} only depends on $\bU, \bV$ through $\bZ$. In the following, our analysis separates two aspects of learning:
\begin{itemize}
    \item The recovery of feature directions, characterized by the singular vectors of $\bZ(t)$ and their overlap with those of $\bC$.
    \item The amplitudes associated with these directions, characterized by the singular values of $\bZ(t)$. 
\end{itemize}
In order to study these objects, we relate them to the symmetric matrices $\bU(t) \bU(t)^\top$ and $\bV(t) \bV(t)^\top$ by exploiting the imbalance conservation. 

\begin{proposition}[Singular structure of the prediction] \label{Prop:SVDPrediction}
    Consider the solution $\bU(t), \bV(t)$ of the gradient flow dynamics \eqref{eq:GFDynamics}, and suppose \cref{Assumption1:Covariance,Assumption1:Initialization} hold. Let $m^* = \mathrm{rank}(\bC) \leq \min(p, d)$. Then, with probability one (under the random initialization $\bR$), for all $t > 0$:
    \begin{equation} \label{eq:RanksAlltimes}
        \mathrm{rank} \, \bU(t) = \min(m, m^*), \hspace{1.2cm} \mathrm{rank} \, \bV(t) = m, \hspace{1.2cm} \mathrm{rank} \, \bZ(t) = \min(m, m^*).
    \end{equation}
    Set $r_Z = \min(m, m^*)$. There exist eigendecompositions
    \begin{equation*}
        \bU(t) \bU(t)^\top = \sum_{i=1}^{r_Z} \lambda_i^U(t) \widehat{\bu}_i(t) \widehat{\bu}_i(t)^\top, \hspace{1.5 cm} \bV(t) \bV(t)^\top = \sum_{i=1}^m \lambda_i^V(t) \widehat{\bv}_i(t) \widehat{\bv}_i(t)^\top,
    \end{equation*}
    where $\lambda_1^U(t) \geq \dots \geq \lambda_{r_Z}^U(t) > 0$ and $\lambda_1^V(t) \geq \dots \geq \lambda_m^V(t) > 0$, such that $\lambda_i^V(t) = \lambda_i^U(t) + 2 \rho$ for $i \in \{1, \dots, r_Z\}$ and $\lambda_i^V(t) = 2 \rho$ for $i \in \{r_Z+1, \dots, m\}$. Finally, we have the singular value decomposition:
    \begin{equation} \label{eq:SVDPrediction}
        \bZ(t)=\sum_{i=1}^{r_Z} \sqrt{\lambda_i^U(t)\big(\lambda_i^U(t)+2\rho\big)} \, \widehat{\bu}_i(t) \widehat{\bv}_i(t)^\top .
    \end{equation}
\end{proposition}

We prove this proposition in \cref{App:Subsec:SVDPrediction}. Note that a similar version of this result exists in the closely related work of \citet[Theorem C.11]{varre2023spectral}, in the case $m \geq d$, where singular vectors of the prediction remain frozen. In our setting (where $m < d$), the singular vectors evolve, leading to a genuine alignment phase. 

This result is essential for our analysis as it shows that the singular values and singular vectors of $\bZ(t)$ can be analyzed through the eigenvalues and eigenvectors of the Gram matrices $\bU(t) \bU(t)^\top$ and $\bV(t) \bV(t)^\top$. As symmetric matrices are more easily analyzed in the high-dimensional limit (see \cref{Sec:HighDim}), we focus on these two matrices to study the singular structure of $\bZ(t)$. 

\subsection{Limit of the Dynamics} \label{Subsec:LimitDynamics}

We now move on to the limit of the dynamics \eqref{eq:GFDynamics} as $t \to \infty$. Thanks to \cref{Prop:SolutionOja}, this limit can be characterized by studying the long-time behavior of the Oja flow.

\begin{proposition}[Long-time limit] \label{Prop:LimitBlockMatrices}
Let $\bC = \sum_{i=1}^{m^*} \sigma_i \bu_i \bv_i^\top$ be a singular value decomposition for $\bC$, with $m^* \leq \min(p, d)$ and $\sigma_1 \geq \dots \geq \sigma_{m^*} > 0$. Whenever $m < m^*$, assume in addition that $\sigma_m > \sigma_{m+1}$. Then, under \cref{Assumption1:Covariance,Assumption1:Initialization}, with probability one (with respect to the randomness of~$\bR$), we have the convergence: 
\begin{subequations} \label{eq:LimitBlockMatrices}
\begin{align}
    \bU(t) \bU(t)^\top &\xrightarrow[t \to \infty]{} \sum_{i=1}^{\min(m,m^*)} (\gamma_i - \rho) \bu_i \bu_i^\top, \label{eq:LimitBlockMatricesU} \\
    \bV(t) \bV(t)^\top &\xrightarrow[t \to \infty]{} \sum_{i=1}^{\min(m,m^*)} (\gamma_i + \rho) \bv_i \bv_i^\top + 2 \rho \1_{m > m^*} \overline \bPi, \label{eq:LimitBlockMatricesV} \\
    \bU(t) \bV(t)^\top &\xrightarrow[t \to \infty]{} \sum_{i=1}^{\min(m,m^*)} \sigma_i \bu_i \bv_i^\top, \label{eq:LimitBlockMatricesZ}
\end{align}
\end{subequations}
where $\gamma_i = \sqrt{\sigma_i^2 + \rho^2}$ and $\overline \bPi$ is the orthogonal projection onto $\mathrm{Im}(\bR) \cap \mathrm{Ker}(\bC) \subset \R^d$. In the case $m>m^*$, with probability one, $\overline \bPi$ has rank $m-m^*$. 
\end{proposition}

Selection of the dominant eigenspace by the Oja flow is classical \citep{yan1994global,tsuzuki2025global}. Our proof first identifies the leading eigenspace selected by the Oja flow dynamics \eqref{eq:ODEW}. In the overparameterized regime, we isolate the invariant subspace inherited from the initialization, which leads to the projection $\overline \bPi$. The three limits then follow by identifying the blocks of $\bW(t) \bW(t)^\top$.

Interestingly, the presence of the projection $\overline \bPi$ (in the overparameterized case $m > m^*$) reveals an initialization-dependent implicit bias. A related phenomenon was studied by \citet{varre2023spectral} in the regime $m \geq d$. In our case, any projection of rank $m-m^*$ with range contained in $\mathrm{Ker}(\bC)$ would leave the loss and the prediction unchanged, but the dynamics selects a specific one from the initialization. The prediction $\bZ = \bU \bV^\top$ is insensitive to this choice and selects the largest singular values of the cross-moment matrix $\bC$. The projection term in equation \eqref{eq:LimitBlockMatricesV} is associated with an \textit{uninformative} part since its range is contained in $\mathrm{Ker}(\bC)$. 

Apart from the projection $\overline \bPi$, the limiting Gram matrices are spectral functions of $\bC \bC^\top$ and $\bC^\top \bC$. Comparing $\bU(t) \bU(t)^\top$ and $\bV(t) \bV(t)^\top$ to these reference matrices will then indicate how the dynamics correlates with the dataset, and more precisely the cross-moment matrix $\bC$. 

\begin{figure}[htbp]
    \centering
    \includegraphics[width=\linewidth]{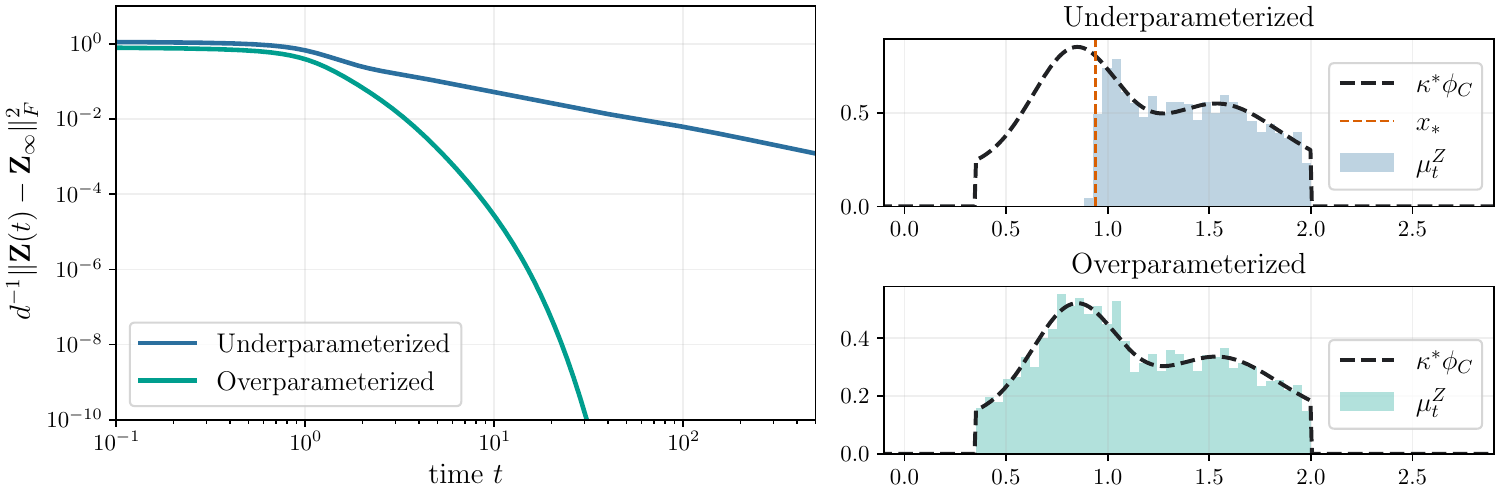}
    \caption{Colored curves and histograms are gradient descent simulations of the dynamics \eqref{eq:GFDynamics}, with $d = 500$, stepsize $\eta=10^{-3}$, $\tau=1$, and $\rho=0.1$, over 10 repetitions. \emph{Left panel:} normalized distance to convergence $d^{-1}\|\bZ(t)-\bZ_\infty\|_F^2$ as a function of $t$, averaged over the repetitions, where $\bZ_\infty$ is the limit in \eqref{eq:LimitBlockMatricesZ}. \emph{Right panels:} histogram of the non-zero singular values of $\bZ(t)$ at $t=500$ for $\kappa = 0.55, \kappa^* = 0.9$ (underparameterized regime) and $ \kappa = 0.75, \kappa^* = 0.55$ (overparameterized regime). The dashed curve is the density $\kappa^* \phi_C$, and the vertical line marks the rank-selection threshold $x_*$ defined in \cref{Cor:StieltjesInf}.}
    \alttext{Left: log-log plot of the distance between the prediction and its limit against time, decreasing slowly in the underparameterized regime and much faster in the overparameterized regime. Right: histograms of the learned singular values at long times. In the underparameterized regime they match the target density only above the threshold. In the overparameterized regime they match the whole density.}
    \label{fig:Fig2}
\end{figure}

\Cref{fig:Fig2} illustrates the spectral selection mechanism of the dynamics from the point of view of the prediction. The left panel measures the distance to the limit $\bZ_\infty$ given by \cref{Prop:LimitBlockMatrices}, and confirms convergence to zero. The right panels show which singular values survive at convergence: in the underparameterized regime only the top part of the spectrum of $\bC$ is selected, while in the overparameterized regime the whole non-zero spectrum is recovered by the dynamics. 

\section{High-Dimensional Limit} \label{Sec:HighDim}

This section describes the dynamics in the high-dimensional limit. The starting point of our analysis is the explicit solution in equation \eqref{eq:SolutionOja}, from which we derive exact expressions for the Gram matrices $\bU(t) \bU(t)^\top \in \mathcal{S}_p^+(\R)$ and $\bV(t) \bV(t)^\top \in \mathcal{S}_d^+(\R)$ (see \cref{App:Subsec:BlockExpressions}). Since these matrices are high-dimensional, we study them through their Stieltjes transforms, a central object in random matrix theory:
\begin{equation*}
    \mathfrak m^A(w) = \frac{1}{d} \tr \Big( \big( \bA - w \bI_d \big)^{-1} \Big), 
\end{equation*}
for a matrix $\bA \in \mathcal{S}_d(\R)$ and $w \in \mathbb{C}\setminus \R$. This object is particularly suited to analyze random matrices in high dimension: the Stieltjes transform is scalar-valued and uniquely determines the empirical spectral distribution of $\bA$. More details can be found in \cref{App:Subsec:RMT}.

Stieltjes transforms encode spectral distributions but contain no information about the eigenvectors of the underlying matrices. Since our work also studies the evolution of learned subspaces, we directly analyze range-projector distances in \cref{Subsec:RangeProjectors}. A complementary approach is to introduce a generalized Stieltjes transform comparing the learned Gram matrices with the cross-moment Gram matrices $\bC\bC^\top$ and $\bC^\top\bC$. This quantity encodes eigenvector overlaps and can therefore track how the prediction learns the relevant subspace. We discuss this extension and derive its high-dimensional limit in \cref{App:Subsec:DoubleStieltjes}.

\subsection{Stieltjes Transforms} \label{Subsec:StieltjesTransforms}

In the following, we are interested in the Stieltjes transforms of the matrices $\bU(t) \bU(t)^\top$, $\bV(t) \bV(t)^\top$:
\begin{equation} \label{eq:DefStieltjes}
    \mathfrak m_t^U(w) = \frac{1}{p} \tr \Big( \big( \bU(t) \bU(t)^\top - w \bI_p \big)^{-1} \Big), \hspace{1.2cm} \mathfrak m_t^V(w) = \frac{1}{d} \tr \Big( \big( \bV(t) \bV(t)^\top - w \bI_d \big)^{-1} \Big). 
\end{equation}
It is also possible to define a Stieltjes transform for the matrix $\bZ(t) = \bU(t) \bV(t)^\top$. Since it is rectangular, we instead consider the empirical measure associated with its singular values and the corresponding Stieltjes transform:
\begin{equation} \label{eq:DefStieltjesZ}
    \mathfrak m_t^Z(w) = \frac{1}{\min(p,d)} \sum_{i=1}^{\min(p,d)} \frac{1}{\sigma_i(t) - w},
\end{equation}
for $w \in \mathbb{C} \setminus \R^+$. Thus $\mathfrak m_t^U$ and $\mathfrak m_t^V$ describe the empirical eigenvalue distributions of the symmetric Gram matrices, whereas $\mathfrak m_t^Z$ describes the empirical singular-value distribution of the prediction. Throughout the paper, we denote by $\mu_t^U$, $\mu_t^V$ and $\mu_t^Z$ the empirical measures respectively associated with the $p$ eigenvalues of $\bU(t) \bU(t)^\top$, the $d$ eigenvalues of $\bV(t) \bV(t)^\top$ and the $\min(p, d)$ singular values of $\bZ(t)$. The connection between these three measures is precisely the content of \cref{Prop:SVDPrediction}.

We now give deterministic limits for these transforms at any fixed training time.

\begin{theorem}[Spectra at fixed time] \label{Thm:Stieltjes}
    Fix $t \geq 0$. In the high-dimensional limit, under \cref{Assumption1}, with probability one, we have the convergence for all $w \in \mathbb{C}\setminus\R^+$:
    \begin{align}
        \mathfrak m_t^U(w) &\xrightarrow[d \to \infty]{} - \frac{1}{w} \left(1 - \frac{\kappa}{\tau} \right) + \frac{\kappa}{\tau} g_t(w) \int \frac{q_t(\gamma_x)}{1 - \kappa + \kappa g_t(w) \big( r_t^\mathsf{L}(\gamma_x) - w q_t(\gamma_x) \big)} \tilde \mu_C(\d x), \label{eq:StieltjesU} \\
        \mathfrak m_t^V(w) &\xrightarrow[d \to \infty]{} - \frac{1 - \kappa}{w} + \kappa g_t(w - 2 \rho) \int \frac{q_t(\gamma_x)}{1 - \kappa + \kappa g_t(w - 2\rho) \big( r_t^\mathsf{R}(\gamma_x) - w q_t(\gamma_x) \big)} \tilde \mu_C(\d x), \label{eq:StieltjesV}
    \end{align}
    where $\gamma_x = \sqrt{x^2 + \rho^2}$, $\tilde \mu_C = (1 - \beta) \delta_0 + \beta \mu_C$ with $\beta = \min(1, \tau)$, and, for $\gamma \geq \rho$,
    \begin{equation*}
    \begin{aligned}
        q_t(\gamma) &= 1 + \frac{\rho}{\gamma} \sinh(2t\gamma) + \frac{\rho^2}{\gamma^2} \big(\cosh(2t\gamma) - 1 \big), \\
        r_t^\mathsf{L}(\gamma) &= 2\rho  \left(1 - \frac{\rho^2}{\gamma^2} \right) \sinh(t\gamma)^2, \\
        r_t^\mathsf{R}(\gamma) &= 2\rho \left(\cosh(t\gamma)+\frac{\rho}{\gamma}\sinh(t\gamma)\right)^2
        =r_t^\mathsf{L}(\gamma)+2\rho q_t(\gamma).
    \end{aligned}
    \end{equation*}
    Moreover, $g_t$ denotes the unique holomorphic solution on $\mathbb{C}\setminus \R^+$ of the equation:
    \begin{equation} \label{eq:SCgt}
        1 = \int \frac{\tilde \mu_C(\d x)}{1 - \kappa + \kappa g_t(w) \big( r_t^\mathsf{L}(\gamma_x) - w q_t(\gamma_x)\big)},
    \end{equation}
    for which $g_t(w) > 0$ whenever $w < 0$.
    
    The limiting measures $\mu_t^U, \mu_t^V$ can be recovered from the Stieltjes--Perron inversion formula (see \cref{Thm:StieltjesPerron}). Additionally, $\mu_t^Z$ is related to $\mu_t^U$ by
    \begin{equation} \label{eq:StieltjesUtoZ}
        \mu_t^Z = \left(1 - \frac{\tau}{\beta} \right) \delta_0 + \frac{\tau}{\beta} (\theta_\rho)_\# \mu_t^U,
    \end{equation}
    where $\theta_\rho(x) = \sqrt{x(x+2 \rho)}$ and $(\theta_\rho)_\#$ denotes the pushforward by $\theta_\rho$ (see \cref{App:Subsec:RMT}).
\end{theorem}

By \cref{Thm:StieltjesWeakConvergence}, the convergence of these Stieltjes transforms is equivalent to the weak convergence of the corresponding spectral measures. The function $g_t$ in \cref{Thm:Stieltjes} is computed self-consistently from an average over the distribution $\tilde \mu_C$ through equation \eqref{eq:SCgt}, as is usual in random matrix theory. Here $\tilde \mu_C$ is the asymptotic spectral measure of $\big(\bC^\top \bC \big)^{1/2} \in \mathcal{S}_d^+(\R)$.

We prove this result in \cref{App:Subsubsec:StieltjesProof}. The main idea of the proof is to combine the explicit solution of \cref{Prop:SolutionOja} with a random matrix result (see \cref{Thm:RMTConvergence}) derived from a deterministic equivalent obtained by \citet{lejeune2024asymptotics}.

Finally, we emphasize the very similar form between the expressions of $\mathfrak m_t^U, \mathfrak m_t^V$. The main differences originate from the difference in normalizations ($\bU \in \R^{p \times m}$ while $\bV \in \R^{d \times m}$), and the asymmetry of the initializations ($\bU_0 = \bzero$ while $\bV_0 = \sqrt{2 \rho} \bR$). The fact that the two Stieltjes transforms share the same self-consistent function $g_t$ evaluated at $w$ and $w - 2 \rho$ is directly linked to the shift $2 \rho$ between the eigenvalues of $\bU(t) \bU(t)^\top$ and $\bV(t) \bV(t)^\top$ established in \cref{Prop:SVDPrediction}. Additionally, we have the stronger identity:
\begin{equation} \label{eq:LinkStieltjes}
    \mathfrak m_t^V(w) = \tau \mathfrak m_t^U(w - 2 \rho) + \frac{\kappa - \tau}{2 \rho - w} - \frac{1 - \kappa}{w},
\end{equation}
which can be derived directly from \cref{Prop:SVDPrediction}.

\begin{figure}[htbp]
    \centering
    \includegraphics[width=\linewidth]{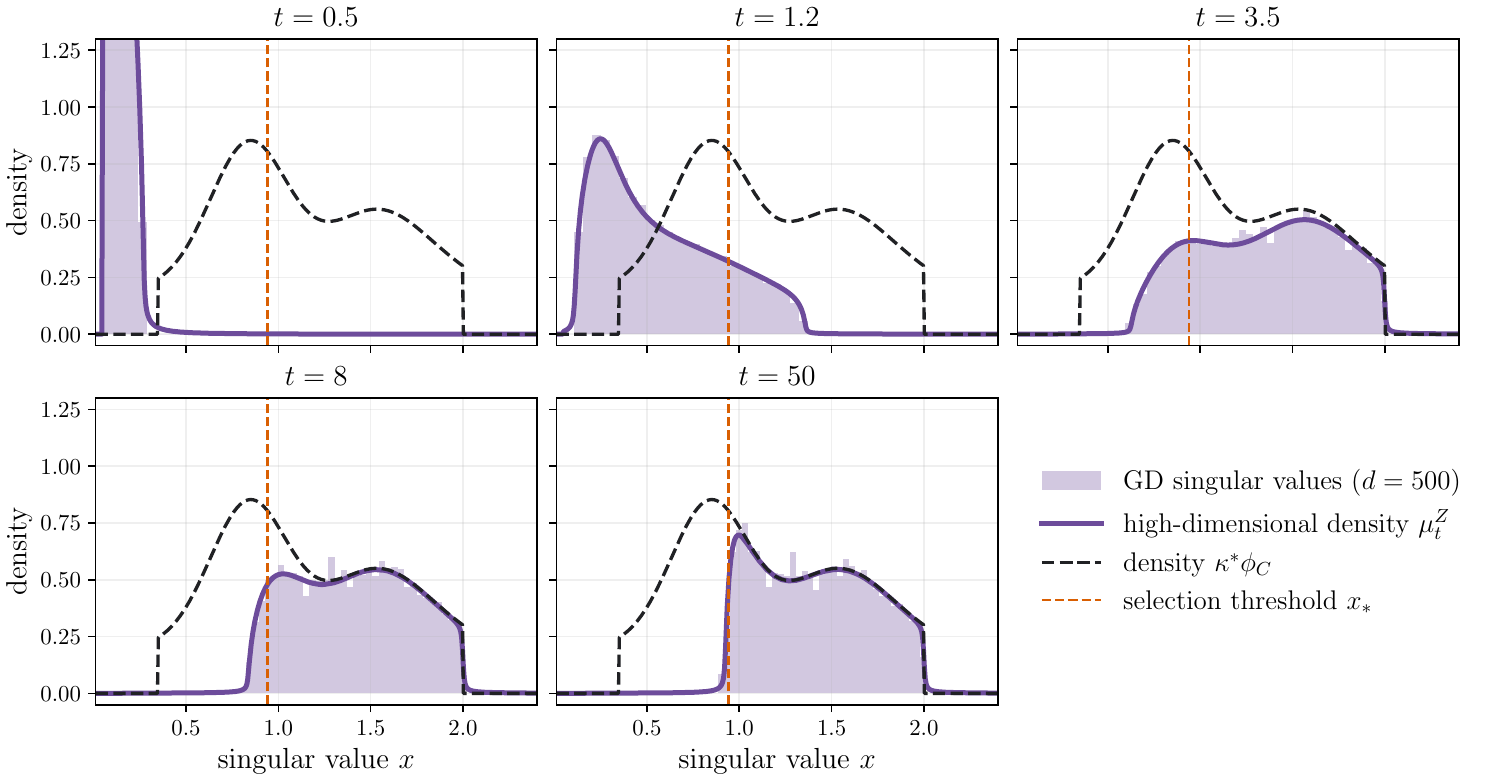}
    \caption{Comparison between the empirical non-zero singular values of $\bZ(t)$ (purple histograms), obtained by gradient descent simulations of equation \eqref{eq:GFDynamics}, and the high-dimensional density derived from equation \eqref{eq:StieltjesUtoZ} (purple curves). Gradient descent simulations are averaged over 20 repetitions, with $d = 500$, stepsize $\eta=10^{-3}$, $\tau = 1$, $\kappa = 0.55$, $\kappa^* = 0.90$, and $\rho=0.1$. Each panel corresponds to a different training time $t$. The dashed black curve is the density $\kappa^* \phi_C$, and the vertical line marks the threshold $x_*$ from \cref{Cor:StieltjesInf}.}
    \alttext{Five panels of the prediction's singular values at increasing training times, where the gradient-descent histograms coincide with the predicted density curves. At early times the mass sits near zero. It then moves to the right and progressively fills the part of the target density above the threshold.}
    \label{fig:Fig3}
\end{figure}

The equations in \cref{Thm:Stieltjes} can be solved numerically, and the limiting measures can be recovered by the Stieltjes--Perron inversion formula (see \cref{Thm:StieltjesPerron}). In \cref{fig:Fig3}, we focus on $\mu_t^Z$ and compare the numerical integration of the above equations with numerical simulations of gradient descent with a small stepsize. At early times, the singular values of $\bZ(t)$ are still close to zero and as time increases, the spectral mass is transported toward the selected part of the spectrum of $\bC$. The density predicted by the above result matches this finite-dimensional evolution throughout the dynamics. Further details on how to simulate equations \eqref{eq:StieltjesU}, \eqref{eq:StieltjesV} and \eqref{eq:SCgt} can be found in \cref{App:Simulations}.

In \cref{App:Subsubsec:StieltjesInit}, we show that taking $t = 0$ in the system of equations of \cref{Thm:Stieltjes} yields the spectral distribution chosen at initialization (\cref{Assumption1:Initialization}). The next result gives the Stieltjes transforms in the limit $t \to \infty$. 

\begin{corollary}[Spectra at long times] \label{Cor:StieltjesInf}
    Consider the Stieltjes transforms in the high-dimensional limit $\mathfrak m_t^U, \mathfrak m_t^V, \mathfrak m_t^Z$ defined in \cref{Thm:Stieltjes}. As $t \to \infty$, we have the convergence, for every $w \in \mathbb{C} \setminus \R^+$:
    \begin{align}
        \mathfrak m_t^U(w) &\xrightarrow[t \to \infty]{} - \frac{1}{w} \left(1 - \frac{\min(\kappa, \beta \kappa^*)}{\tau} \right) + \frac{\beta \kappa^*}{\tau} \int_{x_*}^\infty \frac{\phi_C(x) \d x}{\gamma_x - \rho - w}, \label{eq:StieltjesUinf} \\
        \mathfrak m_t^V(w) &\xrightarrow[t \to \infty]{} - \frac{1-\kappa}{w} + \frac{(\kappa - \beta \kappa^*)^+}{2 \rho - w} + \beta \kappa^* \int_{x_*}^\infty \frac{\phi_C(x) \d x}{\gamma_x + \rho - w}, \label{eq:StieltjesVinf} \\
        \mathfrak m_t^Z(w) &\xrightarrow[t \to \infty]{} - \frac{1}{w} \left( 1 - \frac{\min(\kappa, \beta \kappa^*)}{\beta} \right) + \kappa^* \int_{x_*}^\infty \frac{\phi_C(x) \d x}{x - w}, \label{eq:StieltjesZinf}
    \end{align}
    where, if $\kappa > \beta \kappa^*$, we use the convention $x_* = x_0 = \inf \, \mathrm{Supp} \, \phi_C$, and if $\kappa < \beta \kappa^*$, $x_*$ is defined as
    \begin{equation} \label{eq:Defxstar}
        \kappa = \beta \kappa^* \int_{x_*}^\infty \phi_C(x) \d x. 
    \end{equation}
\end{corollary}

We prove this result in \cref{App:Subsubsec:StieltjesLimit}. We then compare these formulas with the limiting spectral density found in \cref{Prop:LimitBlockMatrices} and show that they match. As a consequence, this shows that the high-dimensional limit $d \to \infty$ commutes with the long-time limit $t \to \infty$. 

In this result, $x_*$ is the rank-selection threshold. In the underparameterized case, it selects a fraction $\kappa$ of the mass of $\tilde \mu_C$, whereas in the overparameterized case, all the positive singular values are selected, and $x_*$ can be chosen as the lower edge $x_0$ of $\phi_C$. This precisely plays the same role as the selection of the $m$ largest singular values of $\bC$ in \cref{Prop:LimitBlockMatrices}. If the underparameterized threshold is not unique, all solutions of equation \eqref{eq:Defxstar} give the same limiting spectral distribution. Throughout the rest of the paper, we choose the largest such solution in the underparameterized case.

\subsection{Range-Projector Distances} \label{Subsec:RangeProjectors}

The previous results describe singular values but not singular vectors. We now study the left and right singular subspaces of $\bZ(t)$, which correspond to the learned subspaces in output and input spaces. By \cref{Prop:SVDPrediction}, these are related to the ranges of $\bU(t)\bU(t)^\top$ and $\bV(t)\bV(t)^\top$.

We define the orthogonal projections onto the left and right ranges of $\bZ(t)$
\begin{equation*}
    \bP^\mathsf{L}(t) = \mathrm{Proj}_{\mathrm{Im} \, \bZ(t)} \in \mathcal{S}_p(\R), \hspace{1.5cm} \bP^\mathsf{R}(t) = \mathrm{Proj}_{\mathrm{Im} (\bZ(t)^\top)} \in \mathcal{S}_d(\R).
\end{equation*}
The main objects we consider are the squared distances
\begin{equation*}
    \Delta^\mathsf{L}(t)^2 = \frac{1}{d} \big\| \bP^\mathsf{L}(t) - \bP^\mathsf{L}_\infty \big\|_F^2, \hspace{1.5cm} \Delta^\mathsf{R}(t)^2 = \frac{1}{d} \big\| \bP^\mathsf{R}(t) - \bP^\mathsf{R}_\infty \big\|_F^2,
\end{equation*}
where $\bP_\infty^\mathsf{L}, \bP_\infty^\mathsf{R}$ are the $t \to \infty$ limits of $\bP^\mathsf{L}(t), \bP^\mathsf{R}(t)$. Whenever $m < d$, the right range of $\bZ(t)$ does not span $\R^d$ (see \cref{Prop:SVDPrediction}), and its orientation can evolve during training. The same holds for the left range in the underparameterized case $m<m^*$. The range projectors converge to those onto the selected singular spaces of $\bC$ (see \cref{Prop:LimitBlockMatrices}), so the distances quantify how the learned features align with the dataset.

\begin{theorem}[High-dimensional range alignment] \label{Thm:Projector1}
    Under \cref{Assumption1}, fix $\rho>0$ and $t>0$, and assume that the singular values of $\bC$ satisfy $\sigma_m > \sigma_{m+1}$ for all $d$ large enough whenever $m < m^*$. For $\mathsf a = \mathsf R$ when $\kappa \in (0, 1)$, and for $\mathsf a = \mathsf L$ whenever $\kappa \in (0,\beta \kappa^*)$, let $g_t^{\mathsf a}$ be the unique positive solution of
    \begin{equation*}
        1 = \int \frac{\tilde \mu_C(\d x)}{1 - \kappa + \kappa g_t^{\mathsf a} r_t^\mathsf{a} (\gamma_x)}.
    \end{equation*}
    Then the following convergence holds almost surely in the high-dimensional limit:
    \begin{equation*}
        \Delta^\mathsf{a}(t)^2 \xrightarrow[d\to\infty]{} \Delta_\rho^\mathsf{a}(t)^2
        := 2(1 - \kappa) \int_{x_*}^\infty \frac{\tilde \mu_C(\d x)}{1 - \kappa + \kappa g_t^{\mathsf a} r_t^\mathsf{a} (\gamma_x)}.
    \end{equation*}
    In the overparameterized case $\kappa > \beta \kappa^*$, one has instead $\Delta_\rho^\mathsf{L}(t)^2 = 0$.
\end{theorem}
We prove this theorem in \cref{App:Subsec:Projectors1}. This result therefore gives a simple expression of the range-projector distance in the high-dimensional limit. In the overparameterized regime, the fact that $\Delta_\rho^\mathsf{L}(t)^2 = 0$ can be intuitively understood from the zero-initialization of $\bU$. Indeed, we have $\bZ(0) = \bzero$ and:
\begin{equation*}
    \dot{\bZ}(0) = \bC \bV_0 \bV_0^\top = 2 \rho \, \bC \bR \bR^\top.
\end{equation*}
Thus, as soon as the dynamics starts, the range of $\bU(t)$ is generated by $\bC \bR \bR^\top$ at leading order. For $m > m^*$, its range is the full non-zero left singular space of $\bC$ with probability one. 

This result is later illustrated in \cref{fig:Fig4} when we consider the small-initialization limit $\rho \to 0$. This limit is the object of the next section.

\section{Vanishing Initialization and Silent Alignment} \label{Sec:SilentAlignment}

We now consider the limit where the initialization scale $\rho$ goes to zero. We show that at fixed $t > 0$, the relevant singular spaces already start to align with those of $\bC$, while the singular values and the loss remain at their initial scale. We refer to this separation between early directional learning and later amplitude growth as \emph{silent alignment}, by analogy with the kernel alignment studied by \citet{atanasov2021neural}. During the second phase, the singular values escape zero only on the timescale $t \asymp \log(\rho^{-1})$. On this scale, learning becomes visible through a spectral threshold moving across the singular values of $\bC$. This section derives these two timescales, and we refer to \cref{fig:Fig1} for a visual summary.

\subsection{Alignment Phase} \label{Subsec:AlignmentPhase}

We first study the alignment phase at fixed time. We begin with a finite-dimensional intuition for the small-initialization limit $\rho \to 0$, before turning to the high-dimensional results.

\subsubsection{Finite-Dimensional Linearized Dynamics} \label{Subsec:FiniteDimIntuition}

We first consider the small-initialization limit in finite dimension and identify the normalization under which the matrices $\bU(t)\bU(t)^\top$, $\bV(t)\bV(t)^\top$, and $\bZ(t)=\bU(t)\bV(t)^\top$ admit non-trivial limits.

\begin{proposition}[Small initialization at fixed time] \label{Prop:FiniteDimensionIntuition}
    Consider $t \geq 0$ fixed. As $\rho \to 0$, under \cref{Assumption1:Covariance,Assumption1:Initialization}, we have:
    \begin{align}
        \frac{1}{\rho} \bU(t) \bU(t)^\top &\xrightarrow[\rho \to 0]{} 2 \sinh(t \bC) \bR \bR^\top \sinh(t \bC)^\top, \\
        \frac{1}{\rho} \bV(t) \bV(t)^\top &\xrightarrow[\rho \to 0]{} 2 \cosh \left( t (\bC^\top \bC)^{1/2} \right) \bR \bR^\top \cosh \left( t (\bC^\top \bC)^{1/2} \right), \\
        \frac{1}{\rho} \bZ(t) &\xrightarrow[\rho \to 0]{} 2 \sinh(t \bC) \bR \bR^\top \cosh \left( t (\bC^\top \bC)^{1/2} \right), \label{eq:FixedTimesLimitZ}
    \end{align}
    where $\sinh$ is applied to each singular value of $\bC$ and $\cosh$ is applied to each eigenvalue of $(\bC^\top \bC)^{1/2}$. 
\end{proposition}

This proposition is a straightforward consequence of the closed-form solution in \cref{Prop:SolutionOja}. We prove this result in \cref{App:Subsec:FiniteDimIntuition}. Additionally, we show that taking the $\rho \to 0$ limit in the Stieltjes transforms obtained in \cref{Thm:Stieltjes} leads to the high-dimensional spectral distribution of the matrices in \cref{Prop:FiniteDimensionIntuition}.

These expressions correspond to a \textit{linearized regime}, where the cubic term is subleading in the Oja flow dynamics \eqref{eq:ODEW}. At $t = \Theta(1)$ the eigenvalues of $\bU(t) \bU(t)^\top$ and $\bV(t) \bV(t)^\top$ are of order $\rho$ but are exponentially amplified by the matrix $\bC$. They become of order one at times of order $\log(\rho^{-1})$. On the other hand, the corresponding eigenspaces are insensitive to this vanishing scale and can therefore correlate with those of $\bC$ at finite time. This is the manifestation of silent alignment: directions are already being selected while the prediction has vanishing singular values and remains close to zero.

We emphasize that when linearizing the dynamics \eqref{eq:GFDynamics}, the terms involving $\bSigma$ are of higher order and vanish in the $\rho \to 0$ limit. This suggests the early alignment phenomenon we exhibit at $t = \Theta(1)$ should hold without the whitening assumption (\cref{Assumption1:Covariance}). However, as our results rely on the closed-form solution that requires this assumption, we keep it as a requirement for the rest of the section. 

\subsubsection{Range Projectors} \label{Subsubsec:RangeProjectorsInit}

We now study range projectors in the small-initialization limit. The distance derived in \cref{Thm:Projector1} has a non-trivial limit as $\rho \to 0$ for $t = \Theta(1)$, revealing the alignment of the range subspaces with those of $\bC$ at this early timescale.

\begin{theorem}[Range alignment at small initialization] \label{Thm:ProjectorLimit}
    Consider the setting of \cref{Thm:Projector1}. Then, in the $\rho \to 0$ limit, with fixed $t > 0$, we have:
    \begin{equation*}
        \Delta_\rho^\mathsf{a}(t)^2 \xrightarrow[\rho \to 0]{} \Delta_0^\mathsf{a}(t)^2
        := 2(1 - \kappa)\int_{x_*}^\infty \frac{\tilde \mu_C(\d x)}{1 - \kappa + \kappa b_t^\mathsf{a} \bar r_t^\mathsf{a}(x)},
    \end{equation*}
    for $\mathsf{a}=\mathsf{R}$ in all regimes, and for $\mathsf{a}=\mathsf{L}$ in the underparameterized regime $\kappa<\beta\kappa^*$. In these regimes the map $t \mapsto \Delta_0^\mathsf{a}(t)^2$ is strictly decreasing on $(0, \infty)$. When $\kappa > \beta \kappa^*$, one has instead $\Delta_0^\mathsf{L}(t)^2=0$. Here
    \begin{equation*}
        \bar r_t^\mathsf{L}(x) = \sinh(tx)^2, \hspace{1.5cm} \bar r_t^\mathsf{R}(x) = \cosh(tx)^2,
    \end{equation*}
    and $b_t^\mathsf{a}$ is the unique positive solution of
    \begin{equation} \label{eq:SCbt}
        1 = \int \frac{\tilde \mu_C(\d x)}{1 - \kappa + \kappa b_t^\mathsf{a} \bar r_t^\mathsf{a}(x)}. 
    \end{equation}
\end{theorem}
We prove this result in \cref{App:Subsubsec:ProjectorLimit1,App:Subsubsec:ProjectorMonotone}. The above formula genuinely depends on $t$ after taking the $\rho \to 0$ limit, and it decreases strictly with $t$: the alignment is already under way at any fixed time. The following theorem shows that this quantity vanishes as $t \to \infty$: the prediction recovers the selected left and right singular spaces of $\bC$. The theorem also provides exact convergence rates.

\begin{theorem}[Convergence rates for range alignment] \label{Thm:ProjectorLimit2}
    Under \cref{Assumption1}, consider the limiting projector distances obtained in \cref{Thm:ProjectorLimit}.
    \begin{itemize}
        \item[$(i)$] \textbf{\emph{Underparameterized case.}} Assume that $\kappa < \beta \kappa^*$, that $\phi_C$ is continuous at $x_*$, and that $\phi_C(x_*) > 0$. Then, for $\mathsf{a} = \mathsf{L}, \mathsf{R}$:
        \begin{equation*}
            \Delta_0^\mathsf{a}(t)^2 \underset{t \to \infty}{\sim} \beta \kappa^* \frac{\phi_C(x_*)}{t} \log(2).
        \end{equation*}
        \item[$(ii)$] \textbf{\emph{Overparameterized case.}} Assume that $\kappa \in (\beta \kappa^*, 1)$, let $x_0 = \inf \, \mathrm{Supp}(\phi_C)>0$, and assume that, for some $\phi_0>0$ and $\ell \in \R^+$:
        \begin{equation*}
            \phi_C(x_0 + h) \underset{h \to 0^+}{\sim} \phi_0 h^\ell.
        \end{equation*}
        Then, denoting by $\Gamma$ the Gamma function:
        \begin{equation*}
            \Delta_0^\mathsf{R}(t)^2 \underset{t \to \infty}{\sim} \frac{8 \beta \kappa^* (1 - \kappa)}{\kappa - \beta \kappa^*} \frac{\phi_0 \Gamma(\ell + 1)}{(2t)^{\ell+1}} e^{-2x_0 t}.
        \end{equation*}
    \end{itemize}
\end{theorem}

We prove these asymptotics in \cref{App:Subsubsec:ProjectorLimit2}. 

\begin{figure}[htbp]
    \centering
    \includegraphics[width=\linewidth]{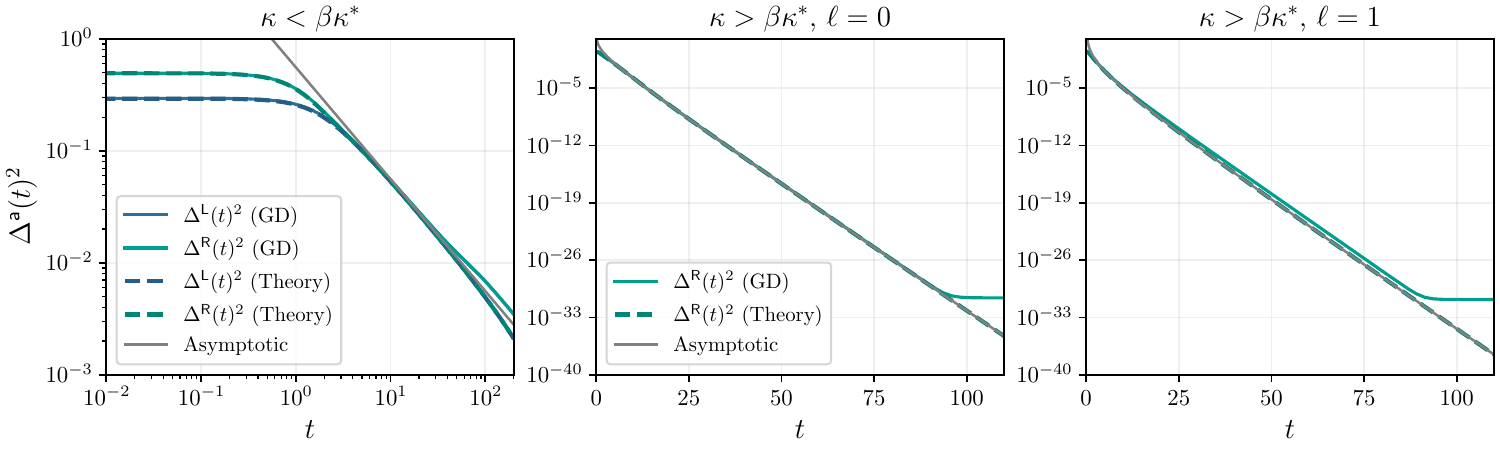}
    \caption{Solid curves are finite-dimensional gradient descent simulations of the dynamics \eqref{eq:GFDynamics}, averaged over 10 repetitions, with $d=500$, stepsize $\eta=10^{-3}$, $\tau=1$, and $\rho=10^{-5}$. Dashed curves are the high-dimensional predictions of \cref{Thm:Projector1}, and gray curves are the long-time asymptotics from \cref{Thm:ProjectorLimit2}. \emph{Left panel:} underparameterized regime, $\kappa = 0.55, \kappa^* = 0.90$ showing the left and right distances. \emph{Middle and right panels:} overparameterized regime, $\kappa = 0.75, \kappa^* = 0.55$, with $\phi_C$ proportional to $(x-0.35)^\ell$ on $[0.35,2]$, for $\ell=0$ in the middle panel and $\ell=1$ in the right panel. The late plateaus in these two panels are numerical precision floors.}
    \alttext{Three plots of subspace distances against time, with logarithmic vertical axes, where the simulations closely follow the predicted curves. In the underparameterized regime (left), the distances decrease slowly, in inverse proportion to time. In the overparameterized regime (middle and right), they decrease exponentially fast for two different shapes of the spectrum's lower edge.}
    \label{fig:Fig4}
\end{figure}

The two regimes have different long-time behaviors. In the overparameterized case, the decay is exponentially fast and is controlled by the left edge $x_0$ of the spectrum of $\bC$, and the exponent $\ell$ only changes the polynomial prefactor. In the underparameterized case, the rank constraint selects the threshold $x_*$, and the singular values around this threshold produce the slower $1/t$ relaxation. This is illustrated in \cref{fig:Fig4}. The finite-dimensional simulations closely follow the high-dimensional prediction from \cref{Thm:Projector1}, while the gray curves show the exact asymptotics from \cref{Thm:ProjectorLimit2}. The small discrepancies come from finite-dimensional effects, and the late plateaus in the middle and right panels are numerical precision floors.

\subsubsection{Top Singular Subspaces} \label{Subsubsec:TopEigenvectors}

Range projectors only measure the learned subspace. We now ask whether the directions inside this subspace are learned in the correct order. To answer this, we compare the projectors onto the top singular spaces of $\bZ(t)$ and $\bC$. By \cref{Prop:SVDPrediction}, those of $\bZ(t)$ are also the top eigenspace projectors of $\bU(t)\bU(t)^\top$ and $\bV(t)\bV(t)^\top$. The next theorem treats a fixed rank in finite dimension and a rank proportional to $d$ in the high-dimensional limit.

\begin{theorem}[Alignment of leading singular subspaces] \label{Thm:TopEigenvectors}
    For $\mathsf{a}=\mathsf{L},\mathsf{R}$ and $k\leq\min(m,m^*)$, let $\bP_{k,\rho}^\mathsf{a}(t)$ and $\bP_k^{C,\mathsf{a}}$ be the top-$k$ left ($\mathsf{a} = \mathsf{L}$) or right ($\mathsf{a} = \mathsf{R}$) singular projectors of $\bZ(t)$ and $\bC$. Suppose \cref{Assumption1:Covariance,Assumption1:Initialization} hold.
    \begin{itemize}
        \item[$(i)$] \textbf{\emph{Finite dimension.}} Assume that $\sigma_k > \sigma_{k+1}$ (with the convention $\sigma_{m^*+1}=0$). Then, almost surely
        \begin{equation} \label{eq:FiniteDimTopProjectorRate}
            \limsup_{\rho \to 0} \big\|\bP_{k, \rho}^\mathsf{a}(t) - \bP_k^{C,\mathsf{a}} \big\|_F^2 \underset{t \to \infty}{=} O \big(e^{-2(\sigma_k - \sigma_{k+1})t}\big).
        \end{equation}
        \item[$(ii)$] \textbf{\emph{High-dimensional limit.}} Assume moreover that \cref{Assumption1} holds, and that $k \sim \alpha d$ for some $\alpha \in \big( 0, \min(\kappa, \beta \kappa^*) \big)$. Let:
        \begin{equation*}
            x_\alpha = \sup \Big\{x > 0, \tilde \mu_C \big((x, \infty) \big) = \alpha \Big\}.
        \end{equation*}
        Consider:
        \begin{equation*}
            \Delta_\alpha^\mathsf{a}(t)^2 = \limsup_{\rho \to 0} \limsup_{d \to \infty} \frac{1}{d} \big\| \bP_{k, \rho}^\mathsf{a}(t) - \bP_k^{C, \mathsf{a}} \big\|_F^2.
        \end{equation*}
        Then, with probability one:
        \begin{itemize}
            \item If there exists $\epsilon_\alpha > 0$ such that $\tilde \mu_C \big((x_\alpha - \epsilon_\alpha, x_\alpha) \big) = 0$, then
            \begin{equation} \label{eq:HighDimTopProjectorRate1}
                \Delta_\alpha^\mathsf{a}(t)^2 \underset{t \to \infty}{=} O(e^{-2 \epsilon_\alpha t}).
            \end{equation}
            \item In all cases,
            \begin{equation} \label{eq:HighDimTopProjectorRate2}
                \Delta_\alpha^\mathsf{a}(t)^2 \underset{t \to \infty}{=} O(t^{-1}).
            \end{equation}
        \end{itemize}
    \end{itemize}
\end{theorem}

We prove this theorem in \cref{App:Subsec:SpectralProjectors}. The finite-dimensional result can be understood from \cref{Prop:FiniteDimensionIntuition}: renormalizing the relevant matrices by $\rho$ allows us to observe how singular values are reordered at early timescales while remaining small. This ordering is exactly what makes the above singular vector convergence possible. The exponentially fast convergence is a direct consequence of the same proposition: for large $t$, a singular direction of $\bC$ associated with a singular value $\sigma$ is amplified by $e^{2\sigma t}$. At the index $k$, its amplification relative to the lower modes is therefore of order $e^{2(\sigma_k - \sigma_{k+1}) t}$. This mechanism is the same as in filtered subspace iteration \citep{stewart1976simultaneous,saibaba2019randomized}, and is closely related in continuous time to the power flow \citep{absil2008continuous}. We also refer to \citet{liu2022convergence} for similar rates in the convergence of the Oja flow. 

The main additional content of \cref{Thm:TopEigenvectors} is its high-dimensional part. If a fixed gap remains below the threshold $x_\alpha$, the exponentially fast convergence is preserved. On the other hand, when $\phi_C(x_\alpha) > 0$, the gap $\sigma_k - \sigma_{k+1}$ typically vanishes in the high-dimensional limit: neighboring singular values accumulate around the threshold, and the convergence becomes a power law. This phenomenon and the associated asymptotics are very similar to \cref{Thm:ProjectorLimit2} for the range projectors.

\begin{figure}[htbp]
    \centering
    \includegraphics[width=\linewidth]{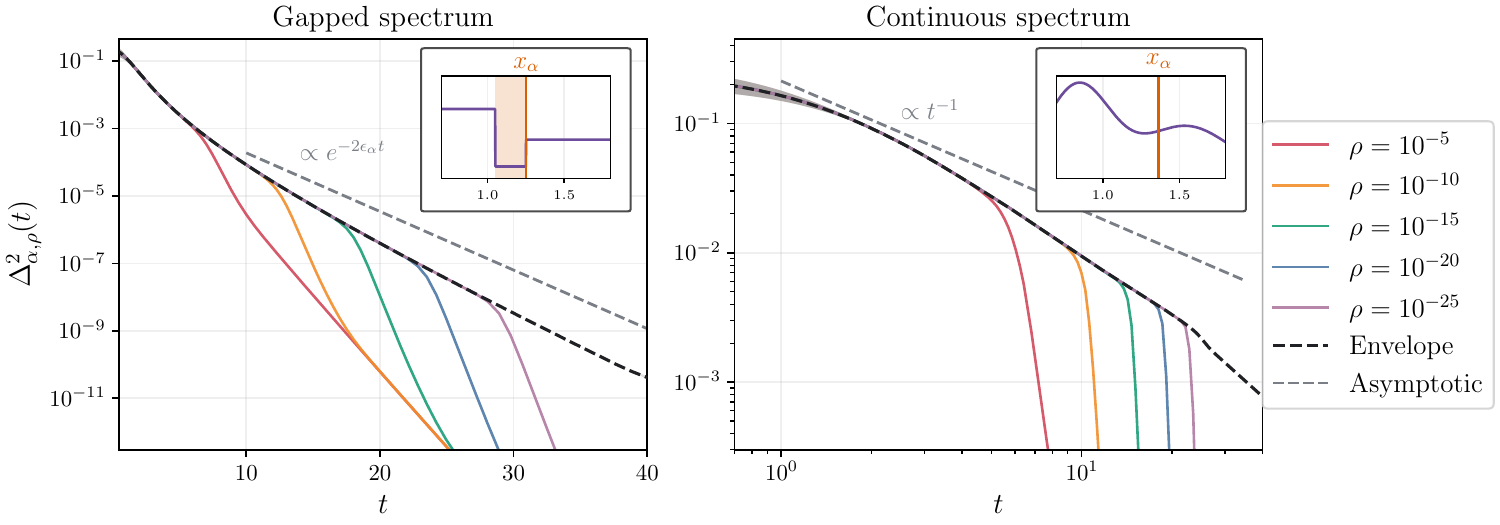}
    \caption{Colored curves are finite-dimensional gradient descent simulations of the dynamics \eqref{eq:GFDynamics}, averaged over 10 repetitions, with $d=500$, stepsize $\eta=10^{-3}$, $\tau=1$, $\kappa=0.55$, $\kappa^*=0.90$, $\alpha=0.30$, and decreasing $\rho$. Each curve averages the left and right distances $d^{-1}\|\bP_{k,\rho}^\mathsf{a}(t)-\bP_k^{C,\mathsf{a}}\|_F^2$ between the top-$k$ singular projectors of $\bZ(t)$ and $\bC$, with $k=\alpha d$. The black dashed curve is the $\rho\to0$ limit of the same distance at $d=500$, computed from the linearized dynamics of \cref{Prop:FiniteDimensionIntuition}. Gray dashed lines show the upper rates from \cref{Thm:TopEigenvectors}, with prefactors chosen for visual comparison. \emph{Left panel:} spectrum with a gap of width $\epsilon_\alpha=0.2$ below $x_\alpha$. \emph{Right panel:} continuous spectrum with positive density at $x_\alpha$. Insets show the local spectral geometry at the cutoff.}
    \alttext{Two plots of the distance between the top singular subspaces of the prediction and of the data, against time, with logarithmic vertical axes, for decreasing initialization scales. Each curve follows a common envelope and then drops to zero, and smaller initializations follow the envelope longer. With a gap in the spectrum, the envelope decreases exponentially, whereas with a continuous spectrum it decreases more slowly.}
    \label{fig:Fig5}
\end{figure}

\Cref{fig:Fig5} illustrates these two regimes. The black envelope is the $\rho \to 0$ projector distance computed from the linearized limit in \cref{Prop:FiniteDimensionIntuition}. At finite $\rho$, the curves first follow this envelope and then leave it as they converge to zero. Decreasing $\rho$ delays this transition and reveals a longer part of the envelope. Gray dashed lines show the bounds obtained in the theorem. Although not tight, these bounds capture the essence of the phenomenon: a gap at the threshold $x_\alpha$ induces exponentially fast convergence (left panel), while a continuous and positive density at the threshold degrades convergence into a power-law asymptotic (right panel). 

\subsection{Learning Phase} \label{Subsec:LearningPhase}

We now consider the learning phase. At the later timescale $t \asymp \log(\rho^{-1})$, the singular vectors have already converged, while the singular values are learned progressively. We start by describing this phenomenon in finite dimension, and then study the high-dimensional limit. 

\subsubsection{Finite-Dimensional Saddle-to-Saddle Dynamics}

We first describe the learning phase in finite dimension and recover a phenomenon of incremental learning, known as saddle-to-saddle dynamics. 

\begin{theorem}[Saddle-to-saddle dynamics] \label{Thm:FiniteDimensionLogTimes}
    Let $s>0$, set $x_s=(2s)^{-1}$, and assume that no singular value of $\bC$ equals $x_s$ and that $\#\{i:\sigma_i>x_s\}\leq m$. Then, with probability one with respect to the randomness of $\bR$, with $t_\rho=s\log(\rho^{-1})$, under \cref{Assumption1:Covariance,Assumption1:Initialization}, the following convergences hold:
    \begin{equation*}
    \begin{aligned}
        \bU(t_\rho)\bU(t_\rho)^\top &\xrightarrow[\rho\to0]{} \mathcal M_s\Big(\big(\bC\bC^\top\big)^{1/2}\Big), \\
        \bV(t_\rho)\bV(t_\rho)^\top &\xrightarrow[\rho\to0]{} \mathcal M_s\Big(\big(\bC^\top\bC\big)^{1/2}\Big), \\
        \bZ(t_\rho) &\xrightarrow[\rho\to0]{} \mathcal M_s(\bC),
    \end{aligned}
    \end{equation*}
    where $\mathcal M_s(x)=x\1_{x>x_s}$ is applied spectrally to $\big(\bC\bC^\top\big)^{1/2}$ and $\big(\bC^\top\bC\big)^{1/2}$, and pointwise to the singular values of $\bC$.
\end{theorem}

We prove this theorem in \cref{App:Subsec:STS}. The proof reads the limits from the closed-form solution, and the only additional step checks that the intermediate limits are strict saddles of the factorized loss. This result is the saddle-to-saddle mechanism of \citet{jacot2021saddle}, which has already been described in two-layer linear networks \citep{gidel2019implicit,varre2023spectral,bach2026closed}. We recall it for completeness as we later show how this discrete process becomes continuous in the high-dimensional limit.

This result makes the spectral nature of the dynamics explicit. The mode associated with $\sigma_i$ is activated when $s>(2\sigma_i)^{-1}$, so between two activation times the prediction is a truncated SVD of $\bC$, which is a saddle point of the loss. Near each saddle, the predictor appears frozen while the next mode grows from a size of order $\rho$. The dynamics thus waits a time of order $\log(\rho^{-1})$ near each saddle, then moves to the next one in a time of order one.

\begin{figure}[htbp]
    \centering
    \includegraphics[width=\linewidth]{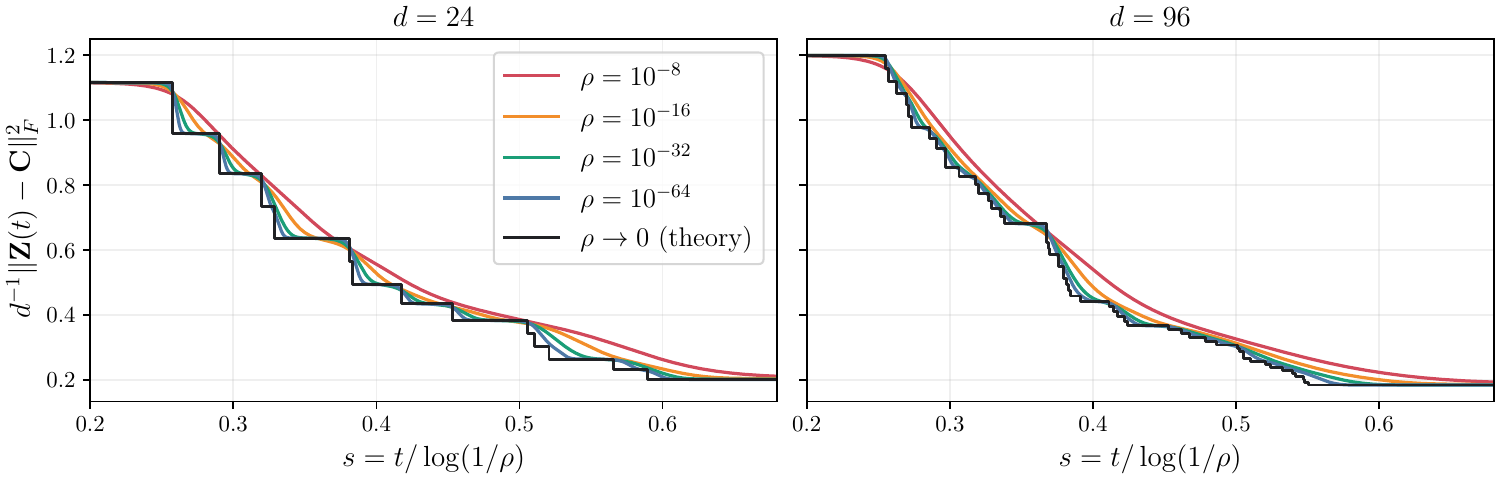}
    \caption{Both panels show $d^{-1}\lVert\bZ(t)-\bC\rVert_F^2$ against $s=t/\log(\rho^{-1})$ for decreasing $\rho$, with $d=24$ on the left and $d=96$ on the right. The black staircases are the corresponding finite-dimensional $\rho\to0$ limits from \cref{Thm:FiniteDimensionLogTimes}. Each plateau is associated with a truncated SVD of $\bC$, and each drop to the activation of another singular mode. Colored curves are computed from gradient descent simulations of the dynamics \eqref{eq:GFDynamics}, with stepsize $\eta = 2 \times 10^{-3}$ and parameters $\tau=1$, $\kappa=0.55$, and $\kappa^*=0.90$.}
    \alttext{Two plots of the loss against rescaled time for decreasing initialization scales. The loss decreases in steps, each step corresponding to one more singular value being learned. As the initialization shrinks, the curves approach a black staircase, with finer steps in the larger dimension.}
    \label{fig:Fig6}
\end{figure}

At each saddle, the loss is carried by the singular values of $\bC$ below $(2s)^{-1}$, as made precise in \cref{Thm:LossTimescales}. \Cref{fig:Fig6} shows this loss as a function of $s$ for decreasing $\rho$ and two values of $d$. In the variable $s$, the moves between saddles become sharper as $\rho$ decreases, which produces the steps of the figure.

\subsubsection{Spectra at Logarithmic Times} \label{Subsubsec:SpectraLogTimes}

We now consider the same logarithmic timescale after taking the high-dimensional limit first. The following theorem shows how the spectral mass progressively escapes zero to match the density $\phi_C$. 

\begin{theorem}[Spectra at logarithmic times] \label{Thm:StieltjesLimit}
    Let $s > 0$ and $t_\rho = s \log(\rho^{-1})$. Denote $x_s = \frac{1}{2s}$. Consider the limiting Stieltjes transforms $\mathfrak m_t^U, \mathfrak m_t^V, \mathfrak m_t^Z$ derived in \cref{Thm:Stieltjes}. For every $w \in \mathbb{C} \setminus \R^+$, we have the limit:
    \begin{align}
        \mathfrak m_{t_\rho}^U(w) &\xrightarrow[\rho \to 0]{} - \frac{1}{w} \left( 1 - \frac{1}{\tau} T_C(x_s) \right) + \frac{\beta \kappa^*}{\tau} \int_{x_s}^\infty \frac{\phi_C(x) \d x}{x-w},  \label{eq:StieltjesLogtimesU} \\ 
        \mathfrak m_{t_\rho}^V(w) &\xrightarrow[\rho \to 0]{} - \frac{1 - T_C(x_s)}{w} + \beta \kappa^* \int_{x_s}^\infty \frac{\phi_C(x) \d x}{x-w},  \label{eq:StieltjesLogtimesV} \\
        \mathfrak m_{t_\rho}^Z(w) &\xrightarrow[\rho \to 0]{} - \frac{1}{w} \left(1 - \frac{T_C(x_s)}{\beta} \right) + \kappa^* \int_{x_s}^\infty \frac{\phi_C(x) \d x}{x - w}, \label{eq:StieltjesLogtimesZ}
    \end{align}
    where for $y > 0$:
    \begin{equation*}
        T_C(y) = \tilde \mu_C \big( (y, \infty) \big).
    \end{equation*}
    This holds as long as $x_s > x_*$, with $x_*$ defined in \cref{Cor:StieltjesInf}.

    For $x_s > x_*$, denote by $\mu_s$ the weak limit of $\mu_{t_\rho(s)}^Z$ as $\rho \to 0$, whose Stieltjes transform is given by equation \eqref{eq:StieltjesLogtimesZ}. Then, the squared 2-Wasserstein distance between $\mu_s$ and $\mu_C$ is given by:
    \begin{equation} \label{eq:WassersteinDistance}
        W_2^2 \big( \mu_s, \mu_C \big) = \kappa^* \int_0^{x_s} x^2 \phi_C(x) \d x. 
    \end{equation}
\end{theorem}

\begin{figure}[htbp]
    \centering
    \includegraphics[width=\linewidth]{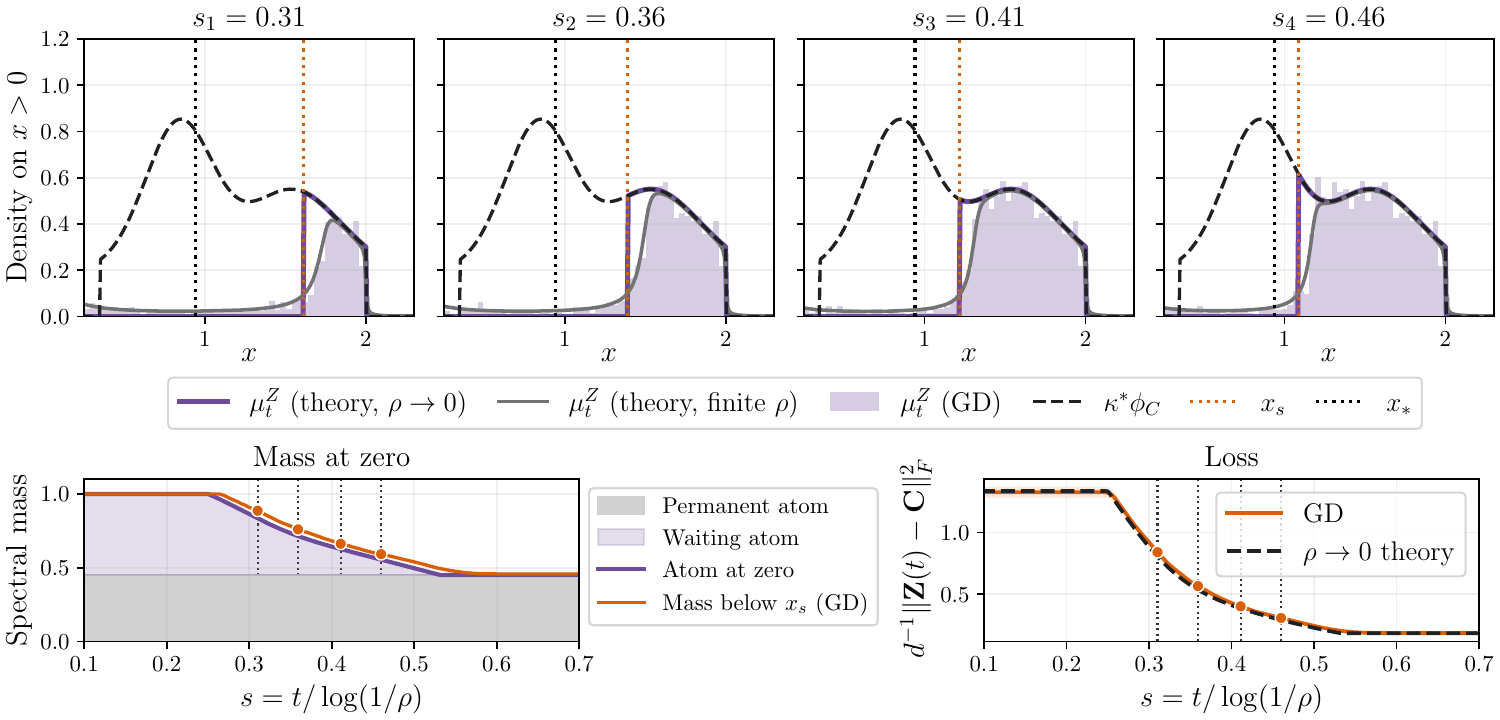}
    \caption{\emph{Top panels:} empirical positive singular-value densities of $\bZ(t)$ at four values of $s=t/\log(\rho^{-1})$, compared with the $\rho\to0$ limit from \cref{Thm:StieltjesLimit}, the finite-$\rho$ high-dimensional density from \cref{Thm:Stieltjes}, and the target density $\kappa^*\phi_C$. The vertical lines mark the moving threshold $x_s$ and the fixed rank-selection threshold $x_*$. \emph{Bottom left panel:} limiting mass at zero, made of a permanent atom due to the rank constraint and a waiting atom of singular values not learned yet, compared with the empirical mass below $x_s$. \emph{Bottom right panel:} normalized prediction loss and its limit from \cref{Thm:LossTimescales}. Orange dots and vertical lines identify the four selected values of $s$. The gradient descent simulations use $\rho=10^{-40}$, $d=500$, stepsize $\eta=10^{-3}$, $\tau=1$, $\kappa=0.55$, and $\kappa^*=0.90$, averaged over 10 repetitions.}
    \alttext{Top: four histograms of the learned singular values at increasing rescaled times, matching the predicted densities. The learned part extends from the largest singular values down to a threshold that decreases over time. Bottom left: the fraction of singular values still at zero, which decreases to a positive plateau. Bottom right: the loss, which decreases and matches its predicted limit.}
    \label{fig:Fig7}
\end{figure}

We prove this in \cref{App:Subsec:StieltjesLimit}. The Stieltjes transforms in \cref{Thm:StieltjesLimit} are already high-dimensional limits: the limit $d\to\infty$ is taken first, followed by $\rho \to 0$ on the logarithmic timescale. The theorem is the high-dimensional counterpart of \cref{Thm:FiniteDimensionLogTimes}. It recovers the same threshold $x_s=(2s)^{-1}$, but the isolated finite-dimensional transitions now merge into a continuous spectral front. Indeed, a direction at spectral location $x > 0$ is amplified with magnitude $\rho e^{2xt}$, which vanishes when $x < x_s$ and diverges when $x > x_s$. As $s$ increases, the front moves continuously from the right edge of the spectrum toward smaller singular values, similarly to the detection wave obtained by \citet{lampinen2019analytic} from an initialization aligned with the data. \Cref{fig:Fig7} follows this evolution through the positive density, the spectral mass at zero, and the corresponding loss.

It is interesting to remark that when considering the renormalized time variable $s$, the spectral front stops at the finite time $s_\text{max} = (2x_*)^{-1}$. We recall that $x_*$ is the spectral cutoff associated with the rank constraint of the prediction: it is either the left edge of the density $\phi_C$ in the overparameterized case, or a larger threshold defined in equation \eqref{eq:Defxstar} in the underparameterized case. In the latter case, for $s > s_\text{max}$, the prediction is constrained by its rank and cannot learn more singular values. 

The Wasserstein identity in equation \eqref{eq:WassersteinDistance} gives a simple interpretation of this spectral evolution. At the logarithmic timescale, the limiting prediction has already learned the singular values above $x_s$, and the remaining error is exactly the cost of transporting the extra mass at zero to the missing singular values below $x_s$.

\subsubsection{Loss Function and Spectral Dynamics} \label{Subsubsec:LossFunction}

To conclude our results in the small-initialization limit, we derive an exact expression of the normalized squared distance between the prediction $\bZ(t)$ and the cross-moment matrix $\bC$. Equation \eqref{eq:LossDistance} shows that the distance $\| \bU \bV^\top - \bC \|_F^2$ is twice the excess empirical loss. Therefore the study of this distance shows how the loss evolves during the dynamics. 

\begin{theorem}[Loss on the two timescales] \label{Thm:LossTimescales}
    Under \cref{Assumption1}, we have the almost-sure limits, for fixed $t \geq 0$ and $s > 0$:
    \begin{align}   
        \lim_{\rho \to 0} \lim_{d \to \infty} \frac{1}{\min(p, d)} \big\| \bZ(t) - \bC \big\|_F^2 &= \kappa^* \int_0^\infty x^2 \phi_C(x) \d x, \label{eq:LossFixedtime} \\
        \lim_{\rho \to 0} \lim_{d \to \infty} \frac{1}{\min(p, d)} \Big\| \bZ \big( s \log(\rho^{-1}) \big) - \bC \Big\|_F^2 &= \kappa^* \int_0^{x_s} x^2 \phi_C(x) \d x, \label{eq:LossLogtime}
    \end{align}
    whenever $x_s = (2s)^{-1} > x_*$.  
\end{theorem}

We prove this theorem in \cref{App:Subsec:Loss}. The proof computes separately the squared norm of the prediction and its inner product with $\bC$. This requires an additional random matrix result, given in \cref{Thm:RMTConvergence2}.

At times $t = \Theta(1)$, the loss is the same as at initialization, although the singular spaces are already aligning, as shown in \cref{Thm:ProjectorLimit}. It decreases only on the scale $t = s\log(\rho^{-1})$, as the singular values of $\bC$ above $x_s$ are learned, and both phases are visible in \cref{fig:Fig1,fig:Fig7}. On this scale, the limit in equation \eqref{eq:LossLogtime} coincides with the squared Wasserstein distance given in equation \eqref{eq:WassersteinDistance}. This shows that the loss contains no alignment error, and is entirely carried by the singular values that have not yet been learned. We refer to \cref{App:Subsubsec:SpectralDynamics} for more details.

\section{Conclusion and Perspectives} \label{Sec:Conclusion}
The main message of this paper is that learning directions and learning amplitudes can take place on different timescales. In two-layer linear networks with small initialization, the singular spaces of the predictor first align with those of $\bC$ while the prediction and the loss remain at their initial values. The singular values grow only later, on a logarithmic timescale, following a threshold that moves down the spectrum of $\bC$. A training phase that looks inactive from the loss can therefore already contain substantial feature learning, and the loss alone is a poor clock for this process.

In the extensive-width regime, combining the closed-form solution with random matrix theory yields explicit limits, at all times, for the spectrum of the prediction and for the distances between the learned and target singular spaces. These limits describe the two successive phases, and show that the saddle-to-saddle transitions that occur in finite dimension merge into a continuous spectral front.

\paragraph{Training time as a rank constraint.} At time $t = s\log(\rho^{-1})$, the predictor keeps the singular values of $\bC$ above $(2s)^{-1}$ and discards the others, until the width becomes the limiting factor. Training time thus controls the effective rank of the predictor, while the width fixes its maximal rank. A similar effect is well known in linear regression, where gradient descent fits the strongest directions of the data first, so that early stopping acts as a spectral regularization \citep{yao2007early,ali2019continuous}. In our setting, the factorized dynamics makes this filter sharp, as each direction is either fully learned or not learned at all.

This ordering helps generalization only if the directions learned are associated with a genuine signal in the data rather than sampling noise. In a spiked model, where a strong enough low-rank signal produces outliers outside a random bulk \citep{baik2005phase,benaychgeorges2011eigenvalues}, the silent phase would first orient the network toward the informative directions, and the spectral front would learn their amplitudes before reaching the bulk, as found by \citet{lampinen2019analytic}. Combined with our dynamical description, such a model could indicate when to stop training, and how much of the noise enters the predictor before that time.

\paragraph{Beyond the exactly solvable setting.} Our analysis relies on whitened inputs, two layers and an isotropic imbalance. Under these assumptions, the whole trajectory depends on a single fixed matrix $\bC$, whose spectrum fixes in advance the order in which directions are learned. For a general input covariance $\bSigma$, the dynamics depends on both $\bSigma$ and $\bC$, whose singular directions differ in general. This spectral structure is then lost, and the learning order may depend on the trajectory itself.

Depth also affects the timescales: for a balanced mode of a depth-$L$ network with $L > 2$, initialized at scale $\sqrt{\rho}$, the dynamics of \citet{saxe2013exact} gives a time of order $\rho^{-(L-2)/2}$ before the amplitude becomes macroscopic, so that the logarithmic timescale is specific to two layers. This analysis, however, starts from directions that are already aligned. Without this assumption, the predictor of a deep linear network and its singular values follow explicit equations for balanced initializations \citep{arora2018optimization,arora2019implicit}. For a single output, \citet{atanasov2021neural} observed numerically that the alignment still precedes the growth of the amplitude, but no longer on a separate timescale. Characterizing rigorously the ordering of alignment and amplitude growth in deeper networks remains open.

\paragraph{The high-dimensional regime without a closed form.} In both of the generalizations discussed above, the closed-form solution on which our high-dimensional analysis relies is no longer available. Whether the two phases identified here persist in the extensive-width regime is therefore an open question. Among high-dimensional tools, dynamical mean-field theory is a natural candidate: it reduces the training dynamics to a few correlation and response functions, and already describes deep linear networks whose width, input dimension and number of samples grow together \citep{bordelon2025deep}, including how outliers leave the bulk of the weight spectra \citep{lauditi2026spectral}. Extending these analyses to small initializations would be a natural way to address this question, with the two-layer linear network as an exactly solvable reference.

\paragraph{Gradient descent and learning schedules.} From a practical viewpoint, early alignment may matter when training is tuned by monitoring the loss. At finite stepsize, the saddle-to-saddle dynamics survives for small steps \citep{gidel2019implicit}, whereas larger steps can change the implicit bias or create oscillations near the edge of stability \citep{even2023s,ghosh2025learning}. A stepsize or schedule chosen from the loss alone may therefore perturb the alignment phase, which the loss does not see. Understanding how a learning schedule acts on each of the two timescales, preserving the alignment while accelerating the later growth of the amplitudes, would make this separation of timescales useful in practice.

{\sectionstarintoc\section*{Funding}
This work was supported by the French government, managed by the Agence Nationale de la Recherche, under the France 2030 program (PR[AI]RIE-PSAI) and the Investissements d'avenir program (PRAIRIE 3IA Institute) [ANR-23-IACL-0008 to F.B. and G.B., ANR-19-P3IA-0001].

\section*{Acknowledgments}
S.M. used large language models (Anthropic's Claude Opus and OpenAI's GPT-5.6) to improve the wording, find references, and help write and check the code for the numerical experiments. These models also helped review the proofs, identify gaps in the arguments, and occasionally find simpler approaches. For instance, they suggested that the high-dimensional part of \cref{Thm:TopEigenvectors} does not require an exact formula for the distance between singular subspaces. They also helped identify and understand the complex analysis tools used to prove convergence of Stieltjes transforms. We checked all results, proofs and code and take full responsibility for the content of this article.

\section*{Data Availability}
The code is available at \url{https://github.com/simonmartin15/silent-alignment-linear-networks}. The data underlying this article are available in Zenodo, at \url{https://doi.org/10.5281/zenodo.23435597}.
}

\bibliography{Biblio}

\newpage
\appendix
\crefalias{section}{appendix}\crefalias{subsection}{subappendix}\crefalias{subsubsection}{subsubappendix}

\section{Prerequisite} \label{App:Prerequisite}
\subsection{Random Matrix Theory} \label{App:Subsec:RMT}

We provide some useful definitions in random matrix theory, and refer to \citet{anderson2010introduction,bai2010spectral} for standard references.

\paragraph{Spectral measures.} If $\bA \in \mathcal{S}_d(\R)$ has eigenvalues $\lambda_1, \dots, \lambda_d$, its empirical spectral measure is given by
\begin{equation*}
    \mu^A = \frac{1}{d} \sum_{i=1}^d \delta_{\lambda_i},
\end{equation*}
where $\delta_\lambda$ is the Dirac mass at $\lambda \in \R$. For a rectangular matrix $\bB \in \R^{p\times d}$, we use the empirical measure of its singular values $\sigma_1, \dots, \sigma_{\min(p,d)}$:
\begin{equation*}
    \mu^B = \frac{1}{\min(p, d)} \sum_{i=1}^{\min(p, d)} \delta_{\sigma_i}.
\end{equation*}
Additionally, we write $\mu_d \Rightarrow \mu$ for weak convergence, meaning convergence against every bounded continuous test function.

\paragraph{Stieltjes transforms.} The Stieltjes transform of a probability measure $\mu$ is:
\begin{equation*}
    \mathfrak m^\mu(w) = \int \frac{\mu(\d x)}{x-w},
\end{equation*}
and is defined for $w \in \mathbb{C} \setminus \mathrm{Supp}(\mu)$. For $\mu = \mu^A$, we write $\mathfrak m^A = \mathfrak m^{\mu^A}$, so that, for a symmetric matrix $\bA \in \mathcal{S}_d(\R)$, the Stieltjes transform of its empirical spectral measure is
\begin{equation*}
    \mathfrak m^A(w) = \frac{1}{d} \mathrm{Tr} \Big( \big( \bA - w \bI_d \big)^{-1} \Big). 
\end{equation*}
The Stieltjes transform characterizes its associated probability measure. Let us denote by $\Im$ the imaginary part of complex numbers.  
\begin{theorem}[Stieltjes--Perron formula, \citet{bai2010spectral}, Theorem B.8] \label{Thm:StieltjesPerron}
    Let $\mu$ be a probability measure and $a < b$ such that $\mu$ does not have an atom at $a$ or $b$. Then:
    \begin{equation*}
        \mu((a,b))=\lim_{\eta\downarrow 0}\frac{1}{\pi}\int_a^b \Im \, \mathfrak m^\mu(x+\mathrm{i}\eta) \d x.
    \end{equation*}
\end{theorem} 
Additionally, pointwise convergence of a sequence of Stieltjes transforms is equivalent to weak convergence of the associated measures. 
\begin{theorem}[\citet{anderson2010introduction}, Theorem 2.4.4] \label{Thm:StieltjesWeakConvergence}
    Given a sequence of probability measures $(\mu_d)_{d \geq 1}$ and a probability measure $\mu$, $\mu_d \Rightarrow \mu$ if and only if $\mathfrak m^{\mu_d}(w) \xrightarrow[d \to \infty]{} \mathfrak m^\mu(w)$ for every $w \in \mathbb{C} \setminus \R$. 
\end{theorem}
Therefore, the convergence results of the Stieltjes transforms in \cref{Thm:Stieltjes,Cor:StieltjesInf,Thm:StieltjesLimit} are to be understood as the convergence of the underlying spectral measures. 

\paragraph{Pushforwards.} If $\theta$ is a measurable map and $\mu$ is a measure, $\theta_\# \mu$ denotes the pushforward of $\mu$ by $\theta$. It is characterized by the following identity, for every bounded continuous function $f$:
\begin{equation*}
    \int f(y) (\theta_\# \mu) (\d y) = \int f(\theta(x)) \mu(\d x).
\end{equation*}

\subsection{Complex Analysis} \label{App:Subsec:ComplexAnalysis}

Stieltjes transforms are functions defined on the complex plane. The convergence results we provide when analyzing them rely on standard complex analysis results that we gather here. 
\begin{lemma} \label{Lemma:StieltjesComplex}
    Let $\mathfrak m$ be the Stieltjes transform of a real probability measure supported on a closed set $S \subset \R$. Then $\mathfrak m$ is holomorphic on $\mathbb{C} \setminus S$. Moreover, given a compact subset $K$ of $\mathbb{C} \setminus S$, we have:
    \begin{equation*}
        \sup_{w \in K} | \mathfrak m(w) | \leq \frac{1}{\mathrm{dist}(S, K)} < \infty. 
    \end{equation*}
\end{lemma}
\begin{proof}
    Since $|x - w| \geq \mathrm{dist}(w, S)$, the function $w \mapsto (x-w)^{-1}$ is holomorphic on $\mathbb{C} \setminus S$. Differentiating under the integral proves holomorphy. Then by the triangle inequality, for every $w \notin S$, we have $|\mathfrak m(w)| \leq \mathrm{dist}(w, S)^{-1} < \infty$. Taking the supremum over $w\in K$ gives the claim.
\end{proof}
We now follow with the Vitali--Porter theorem. 
\begin{theorem}[Vitali--Porter theorem, \citet{schiff1993normal}, Section 2.4] \label{Thm:VitaliPorter}
    Let $(f_n)_{n \geq 1}$ be a sequence of holomorphic functions on a complex domain $\Omega$, such that for all compact subsets $K \subset \Omega$
    \begin{equation*}
        \sup_{w \in K} \sup_{n \geq 1} |f_n(w)| < \infty.
    \end{equation*}
    A sequence satisfying this bound is called locally bounded. Assume that on a set $E \subset \Omega$ with an accumulation point in $\Omega$, $f_n(w)$ converges as $n \to \infty$. Then $f_n$ converges uniformly on compact subsets of $\Omega$ to a holomorphic function. 
\end{theorem}
We now mention the identity theorem, which gives global equality of two holomorphic functions matching on a smaller subset.
\begin{theorem}[Identity theorem, \citet{bak2010complex}, Theorem 6.9] \label{Thm:IdentityTheorem}
    Let $f$ and $g$ be holomorphic on a complex domain $\Omega$. If $f=g$ on a subset of $\Omega$ having an accumulation point in $\Omega$, then $f=g$ on $\Omega$.
\end{theorem}
These two theorems lead to a standard corollary that we use throughout the paper.  
\begin{corollary} \label{Cor:HolomorphicExtension}
    Consider a sequence of holomorphic functions $(f_n)_{n \geq 1}$ satisfying the assumptions of \cref{Thm:VitaliPorter}. If $f$ is a holomorphic function on $\Omega$ and $f_n(w) \xrightarrow[n \to \infty]{} f(w)$ for all $w \in E$, then $f_n$ converges to $f$ uniformly on compact subsets of $\Omega$.
\end{corollary}
\begin{proof}
    By \cref{Thm:VitaliPorter}, the locally bounded sequence converges uniformly on compact subsets of $\Omega$ to some holomorphic function $\tilde f$. On $E$, the pointwise limit gives $\tilde f=f$, hence \cref{Thm:IdentityTheorem} yields $\tilde f=f$ on all of $\Omega$.
\end{proof}
Throughout the paper, we use this corollary to prove convergence of Stieltjes transforms. This result shows that it is enough to establish convergence on a smaller subset. Typically we use $\Omega=\mathbb{C}\setminus\R^+$ and $E=(-\infty,0)$. When the above sequence of functions is indexed by a continuous parameter, the above result is generalized by considering sequences of parameters converging to the desired limit. 

\subsection{Self-Consistent Equations}

Several proofs in the paper involve studying limits of functions computed self-consistently through an integral against a probability measure. Such equations are of the form:
\begin{equation} \label{eq:SelfConsistentGeneral}
    1 = \int \frac{\mu(\d x)}{1 - \kappa + h_\lambda a_\lambda(x)},
\end{equation}
where $\mu$ is a probability measure, $a_\lambda \geq 0$, and $h_\lambda$ is the solution of the previous equation. In the following, the parameter $\lambda$ will be either the initialization scale $\rho \to 0$ or the time $t \to \infty$.

\begin{proposition} \label{Prop:SelfConsistentGeneral}
    Let $\mu$ be a probability measure and $\kappa\in(0,1)$. Consider $\Lambda\subset\R^+$ with a limit point $\lambda_\infty\in[0,+\infty]$ and, for every $\lambda\in\Lambda$, a measurable function $a_\lambda\colon\R\to\R^+$.
    \begin{description}
        \item[1.] Assume that
        \begin{equation*}
            \mu \Big( \big\{ x \in \R, a_\lambda(x) = 0 \big\} \Big) < 1 - \kappa,
        \end{equation*}
        for every $\lambda\in\Lambda$. Then equation \eqref{eq:SelfConsistentGeneral} admits a unique solution $h_\lambda$ in $(0, \infty)$.

        \item[2.] Let $c_\lambda>0$ be such that, for $\mu$-almost every $x$,
        \begin{equation*}
            c_\lambda a_\lambda(x) \xrightarrow[\lambda \to \lambda_\infty]{} a_\infty(x),
        \end{equation*}
        where the limit can be $+\infty$. For $u>0$, define the limit function:
        \begin{equation*}
            \Phi(u)=\int \frac{\mu(\d x)}{1 - \kappa + u a_\infty(x)}.
        \end{equation*}
        Then, consider the cases:
        \begin{description}
            \item[$(i)$] If $\Phi(u) = 1$ admits a unique solution in $(0, \infty)$, let $u_\infty$ be this solution. 
            \item[$(ii)$] If $\Phi(u) < 1$ for every $u > 0$, let $u_\infty = 0$.
            \item[$(iii)$] If $\Phi(u) > 1$ for every $u > 0$, let $u_\infty = \infty$. 
        \end{description}
        In these three cases, we have:
        \begin{equation*}
            \frac{h_\lambda}{c_\lambda} \xrightarrow[\lambda\to\lambda_\infty]{} u_\infty. 
        \end{equation*}

        \item[3.] Moreover, if $u_\infty \in (0, \infty)$ and a family $(f_\lambda)_{\lambda\in\Lambda}$ is uniformly bounded and converges to $f$ $\mu$-almost everywhere as $\lambda\to\lambda_\infty$, then
        \begin{equation} \label{eq:SelfConsistentObservable}
            \int \frac{f_\lambda(x)}{1 - \kappa+h_\lambda a_\lambda(x)} \mu(\d x)
            \xrightarrow[\lambda\to\lambda_\infty]{}
            \int \frac{f(x)}{1 - \kappa+u_\infty a_\infty(x)} \mu(\d x).
        \end{equation}
    \end{description}
\end{proposition}

\begin{proof}
    For \textbf{1.}, let $c_\lambda>0$ be arbitrary, and use the scaling specified in \textbf{2.} for the remaining claims. For $v\geq0$, we work directly with the function:
    \begin{equation*}
        \Phi_\lambda(v) = \int \frac{\mu(\d x)}{1 - \kappa + v c_\lambda a_\lambda(x)}.
    \end{equation*}
    This function is continuous and decreasing, with
    \begin{equation*}
        \Phi_\lambda(0) = \frac{1}{1 - \kappa} > 1,
        \hspace{1.5cm} \lim_{v \to \infty}\Phi_\lambda(v) = \frac{\mu(a_\lambda=0)}{1 - \kappa}< 1.
    \end{equation*}
    It is strictly decreasing because $a_\lambda>0$ on a set of positive measure. Hence the equation $\Phi_\lambda(v)=1$ has a unique solution. By equation \eqref{eq:SelfConsistentGeneral}, this solution is $u_\lambda=h_\lambda/c_\lambda$, which proves \textbf{1.}

    Let $v > 0$ be fixed. Then:
    \begin{equation*}
        \frac{1}{1 - \kappa + v c_\lambda a_\lambda(x)}
        \xrightarrow[\lambda\to\lambda_\infty]{}
        \frac{1}{1 - \kappa + v a_\infty(x)},
    \end{equation*}
    $\mu$-almost everywhere. Moreover, the integrand is upper-bounded by $(1-\kappa)^{-1}$, independently of $\lambda$ and $x$. Dominated convergence guarantees that for every $v>0$:
    \begin{equation} \label{eq:AbstractSCPointwise}
        \Phi_\lambda(v) \xrightarrow[\lambda\to\lambda_\infty]{} \Phi(v).
    \end{equation}
    Let us first assume that $u_\infty\in(0,\infty)$. Consider $\epsilon\in(0,u_\infty)$. Since $\Phi$ is decreasing, we have:
    \begin{equation*}
        \Phi(u_\infty-\epsilon)>1>\Phi(u_\infty+\epsilon). 
    \end{equation*}
    By the previous convergence, this also holds for $\Phi_\lambda$ when $\lambda$ is sufficiently close to $\lambda_\infty$. Since $\Phi_\lambda(u_\lambda)=1$, monotonicity of $\Phi_\lambda$ guarantees that $u_\infty-\epsilon<u_\lambda<u_\infty+\epsilon$. Hence $u_\lambda$ converges to $u_\infty$, proving point $(i)$ of $\textbf{2}$. 

    The same argument applies when $u_\infty=0$ or $u_\infty=+\infty$. In these two cases, for any $\epsilon>0$ or $M>0$, respectively, one has $\Phi_\lambda(\epsilon)<1$ or $\Phi_\lambda(M)>1$ when $\lambda$ is sufficiently close to $\lambda_\infty$. The monotonicity of $\Phi_\lambda$ then proves $(ii)$ and $(iii)$. 

    Finally, consider $(f_\lambda)_{\lambda\in\Lambda}$ as in the proposition and $u_\infty\in(0,\infty)$. We have the convergence $\mu$-almost everywhere:
    \begin{equation*}
        \frac{f_\lambda(x)}{1 - \kappa + h_\lambda a_\lambda(x)}
        \xrightarrow[\lambda\to\lambda_\infty]{}
        \frac{f(x)}{1 - \kappa + u_\infty a_\infty(x)}.
    \end{equation*}
    Additionally, 
    \begin{equation*}
        \left| \frac{f_\lambda(x)}{1 - \kappa + h_\lambda a_\lambda(x)} \right|
        \leq \frac{1}{1 - \kappa}\sup_{\lambda\in\Lambda}\|f_\lambda\|_\infty.
    \end{equation*}
    This bound is finite by assumption, therefore dominated convergence guarantees the result. 
\end{proof}

\section{Finite-Dimensional Dynamics} \label{App:FiniteDim}
In this section we prove the results stated in \cref{Sec:FiniteDim}.

\subsection{Closed-Form Solution} \label{App:Subsec:ProofPropSolutionOja}

We shall now prove \cref{Prop:SolutionOja}. Starting from the dynamics \eqref{eq:GFDynamics} with $\bSigma = \bI_d$, we have:
\begin{equation*}
\begin{aligned}
    \dot \bU(t) &= - \bU(t) \bV(t)^\top \bV(t) + \bC \bV(t), \\
    \dot \bV(t) &= - \bV(t) \bU(t)^\top \bU(t) + \bC^\top \bU(t).
\end{aligned}
\end{equation*}
Now, recall the definition of $\bW(t)$. We have:
\begin{equation*}
    \bW(t)^\top \bW(t) = \bV(t)^\top \bV(t) + \bU(t)^\top \bU(t). 
\end{equation*}
Additionally, since the imbalance matrix $\bImb(t)=\bV(t)^\top \bV(t)-\bU(t)^\top \bU(t)$ remains constant along the dynamics, and equal to $2 \rho \bI_m$ as a consequence of \cref{Assumption1:Initialization}, we have:
\begin{equation*}
    \bV(t)^\top \bV(t) = \frac{\bW(t)^\top \bW(t)}{2} + \rho \bI_m, \hspace{1.5cm} \bU(t)^\top \bU(t) = \frac{\bW(t)^\top \bW(t)}{2} - \rho \bI_m.
\end{equation*}
Therefore:
\begin{equation*}
\begin{aligned}
    \dot \bU(t) &= -  \frac{1}{2} \bU(t) \bW(t)^\top \bW(t) - \rho \bU(t) + \bC \bV(t), \\
    \dot \bV(t) &= - \frac{1}{2} \bV(t) \bW(t)^\top \bW(t) + \rho \bV(t) + \bC^\top \bU(t).
\end{aligned}
\end{equation*}
Stacking the two equations gives the differential equation \eqref{eq:ODEW} on $\bW(t)$. Therefore $\bY(t) = \bW(t) \bW(t)^\top$ satisfies:
\begin{equation} \label{eq:EDOY}
    \dot \bY(t) = \bA_\rho \bY(t) + \bY(t) \bA_\rho - \bY(t)^2. 
\end{equation}
To prove equation \eqref{eq:SolutionOja}, let $\bY_0(t)$ be the solution displayed in this equation:
\begin{equation*}
    \bY_0(t) = e^{t \bA_\rho} \bW_0 \bN(t)^{-1} \bW_0^\top e^{t \bA_\rho}, \hspace{1.5cm} \bN(t) = \bI_m + \bW_0^\top \int_0^t e^{2s\bA_\rho} \d s \, \bW_0. 
\end{equation*}
Then $\bY_0(t = 0) = \bW_0 \bW_0^\top = \bY(t = 0)$ and
\begin{equation*}
    \dot \bY_0(t) = \bA_\rho \bY_0(t) + \bY_0(t) \bA_\rho - e^{t\bA_\rho} \bW_0 \bN(t)^{-1} \dot \bN(t) \bN(t)^{-1} \bW_0^\top e^{t\bA_\rho}. 
\end{equation*}
Since $\dot \bN(t) = \bW_0^\top e^{2t\bA_\rho} \bW_0$, computing the last term gives equation \eqref{eq:EDOY} for $\bY_0(t)$. The proof follows by uniqueness of the solution given the initial condition. 

\subsection{Block Expressions for the Solution} \label{App:Subsec:BlockExpressions}

Recall that the matrices $\bU(t) \bU(t)^\top, \bV(t) \bV(t)^\top, \bU(t) \bV(t)^\top$ can be recovered from equation \eqref{eq:SolutionOja}. We exploit the block structure of $\bA_\rho$ and $\bW_0$ to obtain explicit expressions for these three matrices. 

The key observation is that the matrix
\begin{equation*}
    \bA_\rho = \begin{pmatrix}
        - \rho \bI_p & \bC \\ \bC^\top & \rho \bI_d
    \end{pmatrix} \in \mathcal{S}_{p+d}(\R) 
\end{equation*}
satisfies the identity
\begin{equation} \label{eq:SquareArho}
    \bA_\rho^2 = \begin{pmatrix}  \bC\bC^\top + \rho^2 \bI_p & \bzero \\ \bzero & \bC^\top\bC + \rho^2 \bI_d \end{pmatrix}.
\end{equation}

\begin{lemma} \label{Lemma:BlockDecomposition1}
    Consider the setting and the assumptions of \cref{Prop:SolutionOja}. Define $\bS = (\bC^\top\bC)^{1/2}$, $\bGamma_\rho = (\bS^2 + \rho^2 \bI_d)^{1/2}$, and $\gamma_\rho(x)=\sqrt{x^2+\rho^2}$. For $x \geq 0$ and $\gamma \geq \rho$, let
    \begin{equation*}
    \begin{aligned}
        \chi_t^U(x) &= \frac{x}{\gamma_\rho(x)} \sinh \big( t \gamma_\rho(x) \big), \\
        \chi_t^V(x) &= \cosh\big(t\gamma_\rho(x)\big) + \frac{\rho}{\gamma_\rho(x)} \sinh\big(t\gamma_\rho(x)\big), \\
        q_t(\gamma) &= 1 + \frac{\rho}{\gamma} \sinh(2t\gamma) + \frac{\rho^2}{\gamma^2} \big( \cosh(2t\gamma) - 1 \big). 
    \end{aligned}
    \end{equation*}
    Then, we have the expressions, for all $t \geq 0$:
    \begin{equation*}
    \begin{aligned}
        \bU(t) \bU(t)^\top &= 2 \rho \chi_t^U(\bC) \bR \Big( \bR^\top q_t(\bGamma_\rho) \bR \Big)^{-1} \bR^\top \chi_t^U(\bC)^\top, \\
        \bV(t) \bV(t)^\top &= 2 \rho \chi_t^V(\bS) \bR \Big( \bR^\top q_t(\bGamma_\rho) \bR \Big)^{-1} \bR^\top \chi_t^V(\bS), \\
        \bU(t) \bV(t)^\top &= 2 \rho \chi_t^U(\bC) \bR \Big( \bR^\top q_t(\bGamma_\rho) \bR \Big)^{-1} \bR^\top \chi_t^V(\bS).
    \end{aligned}
    \end{equation*}
\end{lemma}

\begin{proof}
    We first compute $e^{t\bA_\rho}$ from $\bA_\rho^2$. We have, expanding the exponential and decomposing odd and even indices
    \begin{equation*}
        e^{t\bA_\rho} = \sum_{k=0}^\infty \frac{t^{2k}}{(2k)!} (\bA_\rho^2)^k + \sum_{k=0}^\infty \frac{t^{2k+1}}{(2k+1)!} \bA_\rho (\bA_\rho^2)^k. 
    \end{equation*}
    This leads to:
    \begin{equation*}
        e^{t\bA_\rho} = \cosh \big(t (\bA_\rho^2)^{1/2} \big) + \bA_\rho (\bA_\rho^2)^{-1/2} \sinh \big( t (\bA_\rho^2)^{1/2} \big). 
    \end{equation*}
    Then, using the expression of $\bA_\rho^2$ and $\bW_0$, we get:
    \begin{equation} \label{eq:IdentityBlock1}
        e^{t \bA_\rho} \bW_0 = \sqrt{2\rho} \begin{pmatrix} \bC \bGamma_\rho^{-1} \sinh(t \bGamma_\rho) \\
            \cosh(t\bGamma_\rho) + \rho \bGamma_\rho^{-1} \sinh(t\bGamma_\rho) \end{pmatrix} \bR = \sqrt{2\rho}\begin{pmatrix}
            \chi_t^U(\bC) \\ \chi_t^V(\bS) \end{pmatrix}\bR.
    \end{equation}
    The last equality follows directly from the SVD of $\bC$: if $\bC = \sum_i \sigma_i \bu_i \bv_i^\top$, then
    \begin{equation*}
        \bC \bGamma_\rho^{-1} \sinh(t\bGamma_\rho) = \sum_i\frac{\sigma_i}{\sqrt{\sigma_i^2 + \rho^2}} \sinh \Big( t \sqrt{\sigma_i^2 + \rho^2} \Big) \bu_i \bv_i^\top = \chi_t^U(\bC).
    \end{equation*}
    Now evaluating $e^{t\bA_\rho} \bW_0$ at $t = 2s$ and contracting with $\bW_0$, we get:
    \begin{equation*}
        \bW_0^\top e^{2s \bA_\rho} \bW_0 = 2 \rho \bR^\top \chi_{2s}^V(\bS) \bR. 
    \end{equation*}
    Then, since $\bR^\top \bR = \bI_m$:
    \begin{equation*}
        \bI_m + \bW_0^\top \int_0^t e^{2s \bA_\rho} \d s \, \bW_0 = \bR^\top \left( \bI_d + 2 \rho \int_0^t \chi_{2s}^V (\bS) \d s \right) \bR.
    \end{equation*}
    Now, we have, integrating $\chi_{2s}^V$
    \begin{equation*}
        \int_0^t \chi_{2s}^V (\bS) \d s = \frac{1}{2} \bGamma_\rho^{-1} \sinh \big( 2t \bGamma_\rho \big) + \frac{\rho}{2} \bGamma_\rho^{-2}\big( \cosh \big( 2t \bGamma_\rho \big) - \bI_d \big). 
    \end{equation*}
    Therefore, we have:
    \begin{equation*}
        \bI_m + \bW_0^\top \int_0^t e^{2s \bA_\rho} \d s \, \bW_0 = \bR^\top q_t(\bGamma_\rho) \bR. 
    \end{equation*}
    Combining this identity with the solution in equation \eqref{eq:SolutionOja} and the expression of $e^{t \bA_\rho} \bW_0$ in equation \eqref{eq:IdentityBlock1}, we get the desired results by identifying the three blocks of $\bW(t) \bW(t)^\top$.
\end{proof}

\subsection{SVD of the Prediction Matrix} \label{App:Subsec:SVDPrediction}

Let us now prove \cref{Prop:SVDPrediction}. Before going on, we state a useful lemma describing the range of a random matrix $\bR$ uniformly distributed on the Stiefel manifold $\mathrm{St}(d,m)$.

\begin{lemma} \label{Lemma:GenericPositionStiefel}
    Let $\bR$ be uniformly distributed on $\mathrm{St}(d,m)$. Then, we have:
    \begin{description}
        \item[$(i)$] Given $E \subset \R^d$ a deterministic subspace, we have $\mathrm{dim}(\mathrm{Im}(\bR) \cap E) = \max(m + \mathrm{dim}(E) - d, 0)$ almost surely.
        \item[$(ii)$] Given $(\bv_1, \dots, \bv_k)$ a deterministic linearly independent family of $\R^d$ with $k \leq m$, the family $(\bR^\top \bv_1, \dots, \bR^\top \bv_k)$ is linearly independent almost surely. 
    \end{description}
\end{lemma}

\begin{proof}
    By \citet[Theorem 2.2.1]{chikuse2003statistics}, $\bR$ has the same law as $\bG(\bG^\top\bG)^{-1/2}$, where $\bG\in\R^{d\times m}$ has i.i.d. Gaussian entries. Both statements only depend on the law of $\bR$, so we may assume $\bR=\bG(\bG^\top\bG)^{-1/2}$. For the first point, let $e = \mathrm{dim}(E)$, and let $\bV_E \in \R^{d \times (d-e)}$ be an orthonormal basis of $E^\perp$. Then:
    \begin{equation*}
        \dim \big(\mathrm{Im}(\bR) \cap E \big) = m - \mathrm{rank}(\bV_E^\top \bR) = m - \mathrm{rank}(\bV_E^\top \bG).
    \end{equation*}
    Since $\bV_E^\top\bG$ is a $(d-e)\times m$ Gaussian matrix with i.i.d. entries, it has rank $\min(d-e,m)$ almost surely. We then obtain $(i)$. 

    Finally, let $\bV = (\bv_1, \dots, \bv_k)\in\R^{d \times k}$ with full column rank and $k\leq m$. Since $(\bG^\top\bG)^{-1/2}$ is invertible,
    \begin{equation*}
        \mathrm{rank}(\bR^\top\bV)=\mathrm{rank}(\bG^\top\bV).
    \end{equation*}
    The matrix $\bG^\top\bV$ is an $m\times k$ Gaussian matrix whose $m$ rows are i.i.d. $\N(0, \bV^\top \bV)$, hence it has rank $k$ almost surely. Therefore $(\bR^\top \bv_1,\dots,\bR^\top \bv_k)$ is linearly independent almost surely.
\end{proof}

We are now ready to prove \cref{Prop:SVDPrediction}. The proof first proves the rank identities, and then the SVD for the prediction. 

\paragraph{Step 1. Rank of the weights.} We start by showing the rank identities \eqref{eq:RanksAlltimes} for $\bU(t), \bV(t)$. Recall that the imbalance matrix $\bImb(t)=\bV(t)^\top \bV(t)-\bU(t)^\top \bU(t)$ is constant along the dynamics, so that at all times, we have the identity:
\begin{equation*}
    \bV(t)^\top \bV(t) = \bU(t)^\top \bU(t) + 2 \rho \bI_m \succeq 2 \rho \bI_m,
\end{equation*}
and therefore $\mathrm{rank} \, \bV(t) = m$. For $\bU(t)$, we go back to the expression of $\bU(t) \bU(t)^\top$ in \cref{Lemma:BlockDecomposition1}, and get:
\begin{equation*}
    \mathrm{rank} \, \bU(t) = \mathrm{rank} \big( \chi_t^U(\bC) \bR \big).
\end{equation*}
Additionally:
\begin{equation*}
    \chi_t^U(x) = \frac{x}{\gamma_\rho(x)} \sinh \big(t\gamma_\rho(x)\big),
\end{equation*}
so that $\chi_t^U(0)=0$ and $\chi_t^U(x)>0$ whenever $x>0$. Hence $\chi_t^U(\bC)$ has the same kernel as $\bC$ and the same rank $m^*$. Therefore
\begin{equation*}
    \mathrm{rank} \big(\chi_t^U(\bC)\bR\big)
    = m - \mathrm{dim}\big(\mathrm{Im}(\bR)\cap \mathrm{Ker}(\bC)\big).
\end{equation*}
By \cref{Lemma:GenericPositionStiefel}, the dimension of this intersection is $\max(m-m^*,0)$ almost surely. We get that $\mathrm{rank}\,\bU(t)=\min(m,m^*)$ almost surely.

\paragraph{Step 2. SVD of the prediction.} Let us now move to the SVD of the prediction. Since $\bV(t)^\top \bV(t) - \bU(t)^\top \bU(t) = 2 \rho \bI_m$, the two matrices have a common orthonormal eigenbasis $\bo_1(t),\dots,\bo_m(t)$. Setting $r_Z=\min(m,m^*)$, we can choose compatible singular value decompositions
\begin{equation*}
    \bU(t) = \sum_{i=1}^{r_Z} \sqrt{\lambda_i^U(t)} \widehat{\bu}_i(t) \bo_i(t)^\top, \hspace{1.5cm} \bV(t) = \sum_{i=1}^m \sqrt{\lambda_i^V(t)} \widehat{\bv}_i(t) \bo_i(t)^\top.
\end{equation*}
Order the eigenvalues of $\bU(t)^\top\bU(t)$ decreasingly. The imbalance identity gives $\lambda_i^V(t) = \lambda_i^U(t) + 2 \rho$ for $i \in \{1, \dots, r_Z\}$ and $\lambda_i^V(t) = 2 \rho$ for $i > r_Z$. Hence the singular values of $\bV(t)$ are ordered in the same basis, including when some eigenvalues have multiplicities.

Computing $\bZ(t) = \bU(t) \bV(t)^\top$ now gives equation \eqref{eq:SVDPrediction}. Since the map $x \mapsto \sqrt{x(x + 2 \rho)}$ is increasing on $\R^+$, this is an ordered singular value decomposition and $\mathrm{rank}\,\bZ(t)=r_Z$.

\subsection{Convergence of the Dynamics} \label{App:Subsec:ConvergenceOja}

We now prove \cref{Prop:LimitBlockMatrices}. We first establish a general convergence result for the Oja flow in \cref{App:Subsubsec:ConvergenceOja}, and then describe the spectral structure of $\bA_\rho$ in \cref{App:Subsubsec:DiagonalizationArho}. We finally isolate an invariant subspace that appears in the overparameterized case (\cref{App:Subsubsec:ReductionOjaOverparam}) and apply the general result in the overparameterized and underparameterized regimes (\cref{App:Subsubsec:ConvergenceOjaUnderparam,App:Subsubsec:ConvergenceOjaOverparam}).

\subsubsection{Convergence of the Oja Flow} \label{App:Subsubsec:ConvergenceOja}

In the following, we consider the Oja flow dynamics:
\begin{equation} \label{eq:OjaFlow}
    \dot \bH(t) = \big( \bA - \bH(t) \bH(t)^\top \big) \bH(t),
\end{equation}
where $\bH(t) \in \R^{D \times M}$ (with $M \leq D$), $\bA \in \mathcal{S}_D(\R)$. We are interested in the convergence of $\bH(t) \bH(t)^\top$ as $t \to \infty$. The proof of \cref{Prop:SolutionOja} does not use the structure of $\bA_\rho$, so that applying it to $\bW(t) = \sqrt{2} \bH(t)$, which solves equation \eqref{eq:ODEW} with $\bA$ instead of $\bA_\rho$, gives for all $t \geq 0$ \citep[see also][Lemma 2.1]{yan1994global}:
\begin{equation} \label{eq:SolOjaFlow}
    \bH(t) \bH(t)^\top = e^{t \bA} \bH_0 \left( \bI_M + 2 \bH_0^\top \int_0^t e^{2s \bA} \d s \, \bH_0 \right)^{-1} \bH_0^\top e^{t\bA},
\end{equation}
where $\bH_0 = \bH(t = 0) \in \R^{D \times M}$. 

For a closely related flow, \citet{martin2024impact} characterize the limit through the local minimizers of the associated potential and a stable manifold argument \citep{smale1963stable}. Here, we provide a proof by a direct analysis of the dynamics, which requires weaker assumptions on the initialization than the other route (see assumption $(ii)$ in the following proposition).

\begin{proposition} \label{Prop:ConvergenceOja}
    Consider the solution of the Oja flow dynamics \eqref{eq:OjaFlow}. Let $\bA = \sum_{i=1}^D \lambda_i \bu_i \bu_i^\top$, where $\lambda_1 \geq \dots \geq \lambda_D$ and $\bu_1, \dots, \bu_D$ form an orthonormal basis of $\R^D$. Assume that:
    \begin{description}
        \item[$(i)$] $\# \big\{ i \in \{1, \dots, D \}, \lambda_i > 0 \big\} \geq M$. 
        \item[$(ii)$] The family $\big( \bH_0^\top \bu_1, \dots, \bH_0^\top \bu_M \big)$ is a basis of $\R^M$. 
        \item[$(iii)$] Whenever $M < D$, we have $\lambda_M > \lambda_{M+1}$.     
    \end{description}
    Then, we have the convergence:
    \begin{equation*}
        \bH(t) \bH(t)^\top \xrightarrow[t \to \infty]{} \sum_{i=1}^M \lambda_i \bu_i \bu_i^\top. 
    \end{equation*}
\end{proposition}

\begin{proof}
    We start by briefly showing that we can consider without loss of generality that $\bA$ is diagonal. The eigenvalue decomposition for $\bA$ can be written $\bA = \bU \bLambda \bU^\top$, where $\bLambda = \mathrm{diag}(\lambda_1, \dots, \lambda_D)$ and $\bU \in O_D(\R)$ has columns $\bu_1, \dots, \bu_D$. Let $\tilde \bH = \bU^\top \bH$. Then, $\tilde \bH$ is a solution of the differential equation:
    \begin{equation*}
        \dot{ \tilde \bH}(t) = \big( \bLambda - \tilde \bH(t) \tilde \bH(t)^\top \big) \tilde \bH(t). 
    \end{equation*}
    Then, the condition on $\bH_0^\top \bu_1, \dots, \bH_0^\top \bu_M$ can be rephrased as the first $M$ rows of $\tilde \bH_0$ forming a basis of $\R^M$. In the following, if $M=D$, the proof below applies with the lower blocks omitted. 

    Let us write:
    \begin{equation*}
        \bLambda = \begin{pmatrix} \bLambda_1 & \bzero \\ \bzero & \bLambda_2
        \end{pmatrix},
    \end{equation*}
    where $\bLambda_1 = \mathrm{diag}(\lambda_1, \dots, \lambda_M) \in \R^{M \times M}$ and $\bLambda_2 = \mathrm{diag}(\lambda_{M+1}, \dots, \lambda_D) \in \R^{(D-M) \times (D-M)}$. $\bLambda_1$ contains the $M$ largest eigenvalues (all positive) of $\bA$. Then, decompose:
    \begin{equation*}
        \tilde \bH_0 = \begin{pmatrix}
            \bS \\ \tilde \bS
        \end{pmatrix},
    \end{equation*}
    where $\bS \in \R^{M \times M}$ and $\tilde \bS \in \R^{(D-M) \times M}$. By assumption $\bS$ is invertible. We now exploit the solution \eqref{eq:SolOjaFlow}. We have:
    \begin{equation*}
        \bI_M + 2 \tilde \bH_0^\top \int_0^t e^{2s\bLambda} \d s \, \tilde \bH_0 = \bI_M + \bS^\top \varphi_t(\bLambda_1) \bS + \tilde \bS^\top \varphi_t(\bLambda_2) \tilde \bS,
    \end{equation*}
    where
    \begin{equation*}
        \varphi_t(\lambda) = \frac{e^{2t \lambda} - 1}{\lambda},
    \end{equation*}
    and $\varphi_t(0) = 2t$. Then $\varphi_t(\bLambda_1)$ is positive definite and we can factorize:
    \begin{equation*}
    \begin{aligned}
        &\bI_M + 2 \tilde \bH_0^\top \int_0^t e^{2s\bLambda} \d s \, \tilde \bH_0 \\
        &\hspace{1cm}= \bS^\top \varphi_t(\bLambda_1)^{1/2} \left( \bI_M + \varphi_t(\bLambda_1)^{-1/2} \bS^{-\top} \Big( \bI_M + \tilde \bS^\top \varphi_t(\bLambda_2) \tilde \bS \Big) \bS^{-1} \varphi_t(\bLambda_1)^{-1/2} \right) \varphi_t(\bLambda_1)^{1/2} \bS. 
    \end{aligned}
    \end{equation*}
    Moreover, since we assumed $\lambda_M > \lambda_{M+1}$ and $\lambda_1 \geq \dots \geq \lambda_M > 0$, we have:
    \begin{equation*}
        \frac{\varphi_t(\lambda_j)}{\varphi_t(\lambda_i)} \xrightarrow[t \to \infty]{} 0, \hspace{1.5cm} \varphi_t(\lambda_i) \xrightarrow[t \to \infty]{} \infty,
    \end{equation*}
    for every $i \in \{1, \dots, M\}$ and $j \in \{M+1, \dots, D\}$. Therefore, for $i,j \in \{1, \dots, M\}$:
    \begin{equation*}
    \begin{aligned}
        &\left( \varphi_t(\bLambda_1)^{-1/2} \bS^{-\top} \Big( \bI_M + \tilde \bS^\top \varphi_t(\bLambda_2) \tilde \bS \Big) \bS^{-1} \varphi_t(\bLambda_1)^{-1/2} \right)_{ij} \\
        & \hspace{1.5cm} = \frac{1}{\sqrt{\varphi_t(\lambda_i) \varphi_t(\lambda_j)}} \left( \big( \bS^{-\top} \bS^{-1} \big)_{ij} + \sum_{k=1}^{D-M} \big( \tilde \bS \bS^{-1} \big)_{ki} \big( \tilde \bS \bS^{-1} \big)_{kj} \varphi_t(\lambda_{M+k}) \right) \\
        &\hspace{1.5cm} \xrightarrow[t \to \infty]{} 0. 
    \end{aligned}
    \end{equation*}
    Therefore:
    \begin{equation*}
        \bN_1(t) := \bI_M + \varphi_t(\bLambda_1)^{-1/2} \bS^{-\top} \Big( \bI_M + \tilde \bS^\top \varphi_t(\bLambda_2) \tilde \bS \Big) \bS^{-1} \varphi_t(\bLambda_1)^{-1/2} \xrightarrow[t \to \infty]{} \bI_M. 
    \end{equation*}
    Then, we have, using equation \eqref{eq:SolOjaFlow}:
    \begin{equation*}
        \tilde \bH(t) \tilde \bH(t)^\top = \begin{pmatrix}
            e^{t\bLambda_1} \\ e^{t\bLambda_2} \tilde \bS \bS^{-1} 
        \end{pmatrix} \varphi_t(\bLambda_1)^{-1/2} \bN_1(t)^{-1} \varphi_t(\bLambda_1)^{-1/2} \begin{pmatrix}
            e^{t\bLambda_1} & \bS^{-\top} \tilde \bS^\top e^{t \bLambda_2}
        \end{pmatrix}. 
    \end{equation*}
    Now, we have:
    \begin{equation*}
        e^{t \bLambda_1} \varphi_t(\bLambda_1)^{-1/2} \xrightarrow[t \to \infty]{} \bLambda_1^{1/2}, \hspace{1.5cm} e^{t\bLambda_2} \tilde \bS \bS^{-1} \varphi_t(\bLambda_1)^{-1/2} \xrightarrow[t \to \infty]{} \bzero. 
    \end{equation*}
    The first identity is a simple consequence of the expression of $\varphi_t$, and the second can be shown by considering the coefficients of the matrix. Since $\bN_1(t) \xrightarrow[t \to \infty]{} \bI_M$, we finally get:
    \begin{equation*}
        \tilde \bH(t) \tilde \bH(t)^\top \xrightarrow[t \to \infty]{} \begin{pmatrix}
            \bLambda_1 & \bzero \\ \bzero & \bzero
        \end{pmatrix},
    \end{equation*}
    and therefore:
    \begin{equation*}
        \bH(t) \bH(t)^\top \xrightarrow[t \to \infty]{} \bU \begin{pmatrix}
            \bLambda_1 & \bzero \\ \bzero & \bzero
        \end{pmatrix} \bU^\top,
    \end{equation*}
    which is the desired limit.   
\end{proof}

\subsubsection{Spectral Structure of $\bA_\rho$} \label{App:Subsubsec:DiagonalizationArho}

To apply \cref{Prop:ConvergenceOja}, we now diagonalize the matrix $\bA_\rho$ from the singular value decomposition of $\bC$.

\begin{proposition} \label{prop:Diagonalization}
    Let $\bC = \sum_{i=1}^{m^*} \sigma_i \bu_i \bv_i^\top$ be the SVD of $\bC$. Complete the families $\{\bu_i\}_{1 \leq i \leq m^*}$ and $\{\bv_i\}_{1 \leq i \leq m^*}$ into orthonormal bases of $\R^p$ and $\R^d$. For $i \in \{1, \dots, m^*\}$, define $\gamma_i = \sqrt{\rho^2 + \sigma_i^2}$, and the vectors of $\R^{p+d}$:
    \begin{equation*}
    \begin{aligned}
        \bz_i^{\pm} &= \frac{1}{\sqrt{\sigma_i^2 + (\rho \pm \gamma_i)^2}} \begin{pmatrix} \sigma_i \bu_i \\ (\rho \pm \gamma_i) \bv_i \end{pmatrix}, &\hspace{1.5cm} &i \in \{1, \dots, m^*\}, \\
        \bq_j^- &= \begin{pmatrix} \bu_j \\ \bzero \end{pmatrix}, &\hspace{1.5cm} &j \in \{m^*+1, \dots, p\}, \\
        \bq_k^+ &= \begin{pmatrix} \bzero \\ \bv_k \end{pmatrix}, &\hspace{1.5cm} &k \in \{m^*+1, \dots, d \}. 
    \end{aligned}
    \end{equation*}
    Then, we have the eigenvalue decomposition:
    \begin{equation*}
        \bA_\rho = \sum_{i=1}^{m^*} \gamma_i \bz_i^+(\bz_i^+)^\top - \sum_{i=1}^{m^*} \gamma_i \bz_i^- (\bz_i^-)^\top + \rho \sum_{k=m^*+1}^d \bq_k^+ (\bq_k^+)^\top - \rho \sum_{j=m^*+1}^p \bq_j^- (\bq_j^-)^\top.   
    \end{equation*}
\end{proposition}

Indeed, for $i \leq m^*$, a direct computation gives $\bA_\rho \bz_i^\pm = \pm \gamma_i \bz_i^\pm$, using $\sigma_i^2=(\gamma_i-\rho)(\gamma_i+\rho)$. Moreover $\bA_\rho \bq_j^-=-\rho \bq_j^-$ and $\bA_\rho \bq_k^+=\rho \bq_k^+$. These $p+d$ vectors form an orthonormal basis, which proves the decomposition.

\subsubsection{Reduction in the Overparameterized Case} \label{App:Subsubsec:ReductionOjaOverparam}

In the following we study the overparameterized case $m > m^*$. We consider the notation of \cref{Prop:SolutionOja}. Recall that the dynamics on $\bW(t)$ is initialized with:
\begin{equation*}
    \bW_0 = \sqrt{2 \rho} \begin{pmatrix} \bzero \\ \bR
    \end{pmatrix},
\end{equation*}
where $\bR \in \R^{d \times m}$ is such that $\bR^\top \bR = \bI_m$. In the following, we consider the subspace:
\begin{equation} \label{eq:SubspaceE0}
    E_0 = \begin{pmatrix}
        \bzero \\ \mathrm{Im}(\bR) \cap \mathrm{Ker}(\bC) 
    \end{pmatrix} \subset \R^{p+d}. 
\end{equation}

\begin{lemma} \label{Lemma:TrivialSubspace}
    Consider the notation of \cref{Prop:SolutionOja}. Under \cref{Assumption1:Covariance,Assumption1:Initialization}, with probability one, we have $\mathrm{dim}(E_0) = \max(m-m^*, 0)$. Moreover, for any $\bx \in E_0$, we have $\bW(t) \bW(t)^\top \bx = 2 \rho \bx$ for all $t \geq 0$. 
\end{lemma}

\begin{proof}
    The dimension follows from $(i)$ in \cref{Lemma:GenericPositionStiefel}, since $\mathrm{rank}(\bC) = m^*$. For $\bx \in E_0$, one has $\bA_\rho \bx = \rho \bx$ and $\bW_0 \bW_0^\top \bx = 2\rho \bx$. As a consequence,
    \begin{equation*}
        \left( \bI_m + \bW_0^\top \int_0^t e^{2s\bA_\rho} \d s \, \bW_0 \right) \bW_0^\top \bx = \left( 1 + 2\rho \int_0^t e^{2\rho s} \d s \right) \bW_0^\top \bx = e^{2\rho t} \, \bW_0^\top \bx.
    \end{equation*}
    Applying the closed form \eqref{eq:SolutionOja} then gives the result. 
\end{proof}

From now on, we assume $m > m^*$ (overparameterized case), so that $\mathrm{dim}(E_0) = m-m^* > 0$. For $t > 0$, the Woodbury identity applied to the solution in equation \eqref{eq:SolutionOja} gives
\begin{equation*}
\begin{aligned}
    \bW(t) \bW(t)^\top &= e^{t\bA_\rho} \Big( \bF(t)^{-1} - \bF(t)^{-1} \big( \bF(t)^{-1} + \bW_0 \bW_0^\top \big)^{-1} \bF(t)^{-1} \Big) e^{t \bA_\rho}, \\
    \bF(t) &= \int_0^t e^{2s \bA_\rho} \d s,
\end{aligned}
\end{equation*}
where $\bF(t)$ is invertible. The space $E_0$ and its orthogonal complement are invariant under both $\bA_\rho$ and $\bW_0\bW_0^\top$. Thus, decomposing $\R^{p+d}$ along this orthogonal sum, we have:
\begin{equation} \label{eq:SolOjaOrthogonal}
    \bY(t) = 2 \rho \bPi + e^{t\tilde \bA} \Big( \tilde \bF(t)^{-1} - \tilde \bF(t)^{-1} \big( \tilde \bF(t)^{-1} + \tilde \bY_0 \big)^{-1} \tilde \bF(t)^{-1} \Big) e^{t \tilde \bA},
\end{equation}
where $\bPi$ is the orthogonal projection onto $E_0$, and
\begin{equation*}
    \tilde \bY_0 = \bPi^\perp \bW_0 \bW_0^\top \bPi^\perp, \hspace{1.2cm} \tilde \bF(t) = \int_0^t e^{2s \tilde \bA} \d s, \hspace{1.2cm} \tilde \bA = \bPi^\perp \bA_\rho \bPi^\perp,
\end{equation*}
and $\bPi^\perp = \bI_{p+d} - \bPi$ is the orthogonal projection onto $E_0^\perp$. 

\begin{lemma} \label{Lemma:ReducedInitialization}
    Consider the setting of \cref{Lemma:TrivialSubspace}. In the overparameterized case $m > m^*$, with probability one, there exists $\tilde \bR \in \R^{d \times m^*}$ such that $\tilde \bR^\top \tilde \bR = \bI_{m^*}$, and:
    \begin{equation*}
        \tilde \bY_0 = 2 \rho \begin{pmatrix}
            \bzero & \bzero \\ \bzero & \tilde \bR \tilde \bR^\top
        \end{pmatrix}.
    \end{equation*}
\end{lemma}

\begin{proof}
    Due to the form of $E_0$ in equation \eqref{eq:SubspaceE0}, $\bPi$ can be written as:
    \begin{equation} \label{eq:BlockProjection}
        \bPi = \begin{pmatrix} \bzero & \bzero \\ \bzero & \overline \bPi
        \end{pmatrix},
    \end{equation}
    where $\overline \bPi$ is the orthogonal projection onto $\mathrm{Im}(\bR) \cap \mathrm{Ker}(\bC)$. Then, the expression of $\bW_0$ guarantees that:
    \begin{equation*}
        \tilde \bY_0 = 2 \rho \begin{pmatrix}
            \bzero & \bzero \\ \bzero & (\bI_d - \overline \bPi) \bR \bR^\top (\bI_d - \overline \bPi)
        \end{pmatrix}.
    \end{equation*}
    Since $\bR \bR^\top$ is a projection and $\mathrm{Im}(\overline \bPi) \subset \mathrm{Im}(\bR)$, we have $\bR\bR^\top \overline \bPi = \overline \bPi$, and therefore:
    \begin{equation*}
        (\bI_d - \overline \bPi) \bR \bR^\top (\bI_d - \overline \bPi) = \bR \bR^\top - \overline \bPi.
    \end{equation*}
    The matrix $\bR \bR^\top - \overline \bPi$ is the orthogonal projection onto the orthogonal complement of $\mathrm{Im}(\bR)\cap\mathrm{Ker}(\bC)$ within $\mathrm{Im}(\bR)$. Its rank is
    \begin{equation*}
        m-\mathrm{dim}\big(\mathrm{Im}(\bR)\cap\mathrm{Ker}(\bC)\big)=m^*,
    \end{equation*}
    with probability one (from \cref{Lemma:TrivialSubspace}). Therefore, we can write $\bR \bR^\top - \overline \bPi = \tilde \bR \tilde \bR^\top$ where $\tilde \bR \in \R^{d \times m^*}$ is such that $\tilde \bR^\top \tilde \bR = \bI_{m^*}$. This concludes the proof.    
\end{proof}

This lemma leads to:
\begin{equation*}
    \tilde \bY_0 = \tilde \bW_0 \tilde \bW_0^\top, \hspace{1.5cm} \tilde \bW_0 = \sqrt{2 \rho} \begin{pmatrix}
        \bzero \\ \tilde \bR
    \end{pmatrix} \in \R^{(p+d) \times m^*}. 
\end{equation*}
Therefore, applying the Woodbury formula again to equation \eqref{eq:SolOjaOrthogonal}, we can write:
\begin{equation} \label{eq:SolOjaOrthogonal2}
    \bY(t) = 2 \rho \bPi + e^{t \tilde \bA} \tilde \bW_0 \left( \bI_{m^*} + \tilde \bW_0^\top \int_0^t e^{2s \tilde \bA} \d s \, \tilde \bW_0 \right)^{-1} \tilde \bW_0^\top e^{t \tilde \bA}. 
\end{equation}

\subsubsection{Convergence in the Overparameterized Case} \label{App:Subsubsec:ConvergenceOjaOverparam}

The goal is now to apply \cref{Prop:ConvergenceOja} to the matrix in equation \eqref{eq:SolOjaOrthogonal2}. Up to the constant term $2 \rho \bPi$, this matrix has the form of the solution of the Oja flow in equation \eqref{eq:SolOjaFlow} with an initialization $\tilde \bW_0 \in \R^{(p+d) \times m^*}$, which brings the case $m > m^*$ back to the case $m = m^*$. This step is needed since, with $M = m$, assumption $(iii)$ of \cref{Prop:ConvergenceOja} fails for $\bA_\rho$, whose $m$-th and $(m+1)$-th eigenvalues both equal $\rho$ (see \cref{prop:Diagonalization}). To this end, we investigate how $\bPi^\perp$ acts on $\bA_\rho$. Back to \cref{prop:Diagonalization}, we have, using the block structure of $\bPi$ in equation \eqref{eq:BlockProjection}:
\begin{equation*}
\begin{aligned}
    \bPi \bz_i^{\pm} &= \frac{1}{\sqrt{\sigma_i^2 + (\rho \pm \gamma_i)^2}} \begin{pmatrix} \bzero \\ (\rho \pm \gamma_i) \overline \bPi \bv_i \end{pmatrix}, &\hspace{1.5cm} &i \in \{1, \dots, m^*\}, \\
    \bPi \bq_j^- &= \bzero, &\hspace{1.5cm} &j \in \{m^*+1, \dots, p\}, \\
    \bPi \bq_k^+ &= \begin{pmatrix} \bzero \\ \overline \bPi \bv_k \end{pmatrix}, &\hspace{1.5cm} &k \in \{m^*+1, \dots, d \}. 
\end{aligned}
\end{equation*}
Since $\mathrm{Im}(\overline \bPi) \subset \mathrm{Ker}(\bC)$ and $\bv_i \in \mathrm{Ker}(\bC)^\perp$ for $i = 1, \dots, m^*$, we have $\bPi \bz_i^{\pm} = \bzero$. Therefore, only the vectors $\bq_k^+$ for $k \in \{m^*+1, \dots, d\}$ are changed by $\bPi^\perp$. Therefore:
\begin{equation*}
    \bPi^\perp \bA_\rho \bPi^\perp = \sum_{i=1}^{m^*} \gamma_i \bz_i^+(\bz_i^+)^\top - \sum_{i=1}^{m^*} \gamma_i \bz_i^- (\bz_i^-)^\top + \rho \sum_{k=m^*+1}^d \bPi^\perp \bq_k^+ (\bq_k^+)^\top \bPi^\perp - \rho \sum_{j=m^*+1}^p \bq_j^- (\bq_j^-)^\top.
\end{equation*}
Apart from the third term, this is still an eigendecomposition for $\bPi^\perp \bA_\rho \bPi^\perp$. Indeed, $\bz_i^\pm$ and $\bq_j^-$ are orthogonal to $\bq_k^+$ and to $E_0$, hence to $\bPi^\perp \bq_k^+$. The third term is positive semidefinite and has operator norm at most $\rho$:
\begin{equation*}
    \rho \lambda_\text{max} \left( \bPi^\perp \sum_{k=m^*+1}^d \bq_k^+ (\bq_k^+)^\top \bPi^\perp \right) \leq \rho \lambda_\text{max} \left( \sum_{k=m^*+1}^d \bq_k^+ (\bq_k^+)^\top \right) = \rho, 
\end{equation*}
where we used that $\bPi^\perp$ is an orthogonal projection and $(\bq_k^+)_{m^* + 1 \leq k \leq d}$ is an orthonormal family. Therefore, every eigenvalue outside the span of $\bz_1^+,\dots,\bz_{m^*}^+$ is at most $\rho$. Since $\gamma_i=\sqrt{\sigma_i^2+\rho^2}>\rho$, the top $m^*$ eigenvalues of $\tilde\bA$ are $\gamma_1,\dots,\gamma_{m^*}$, with eigenvectors $\bz_1^+,\dots,\bz_{m^*}^+$. This verifies assumptions $(i)$ and $(iii)$ of \cref{Prop:ConvergenceOja}.

Let us now verify assumption $(ii)$ of \cref{Prop:ConvergenceOja}. To do so, we need to prove that the family $\tilde \bW_0^\top \bz_1^+, \dots, \tilde \bW_0^\top \bz_{m^*}^+$ forms a basis of $\R^{m^*}$. Since $\tilde \bW_0^\top \tilde \bW_0 = 2 \rho \bI_{m^*}$ is invertible, $\tilde \bW_0$ is injective, and it is therefore enough to show that the vectors $\tilde \bW_0\tilde \bW_0^\top \bz_1^+, \dots, \tilde \bW_0 \tilde \bW_0^\top \bz_{m^*}^+$ are linearly independent in $\R^{p+d}$. We have:
\begin{equation*}
    \tilde \bW_0 \tilde \bW_0^\top \bz_i^+ = \frac{2 \rho (\rho + \gamma_i)}{\sqrt{\sigma_i^2 + (\rho + \gamma_i)^2}} \begin{pmatrix}
        \bzero \\ \tilde \bR \tilde \bR^\top \bv_i
    \end{pmatrix}. 
\end{equation*}
Since $\tilde \bR \tilde \bR^\top = \bR \bR^\top - \overline \bPi$, and $\overline \bPi \bv_i = \bzero$ for $i \in \{1, \dots, m^*\}$, we have $\tilde \bR \tilde \bR^\top \bv_i = \bR \bR^\top \bv_i$. Since $m > m^*$, point $(ii)$ in \cref{Lemma:GenericPositionStiefel} shows that $\bR^\top \bv_1, \dots, \bR^\top \bv_{m^*}$ are linearly independent almost surely. Since $\bR$ is injective on $\R^m$, the family $\bR\bR^\top \bv_1, \dots, \bR \bR^\top \bv_{m^*}$ is linearly independent in $\R^d$, which proves the claim. 

We are now ready to apply the result of \cref{Prop:ConvergenceOja}. Back to equation \eqref{eq:SolOjaOrthogonal2}, we let $\bH_0 = \frac{1}{\sqrt{2}} \tilde \bW_0 \in \R^{(p+d) \times m^*}$, so that:
\begin{equation*}
    \bY(t) = 2 \rho \bPi + 2 e^{t \tilde \bA} \bH_0 \left( \bI_{m^*} + 2 \bH_0^\top \int_0^t e^{2s \tilde \bA} \d s \, \bH_0 \right)^{-1} \bH_0^\top e^{t \tilde \bA}. 
\end{equation*}
Given the eigendecomposition of $\tilde \bA$, we get the limit:
\begin{equation} \label{eq:LimitOverparam}
    \bY(t) \xrightarrow[t \to \infty]{} 2 \rho \bPi + 2 \sum_{i=1}^{m^*} \gamma_i \bz_i^+ (\bz_i^+)^\top. 
\end{equation}

\subsubsection{Convergence in the Underparameterized Case} \label{App:Subsubsec:ConvergenceOjaUnderparam}

Whenever $m \leq m^*$, $E_0 = \{ \bzero \}$ with probability one, and we can directly apply \cref{Prop:ConvergenceOja}. Using \cref{prop:Diagonalization}, the largest eigenvalues of $\bA_\rho$ are $\gamma_1 \geq \dots \geq \gamma_{m^*}$. Thus there are at least $m$ positive eigenvalues. If $m < m^*$, the assumed gap $\sigma_m > \sigma_{m+1}$ gives $\gamma_m > \gamma_{m+1}$, and if $m = m^*$, then $\gamma_m > \rho$, and we have shown that the next eigenvalue is at most $\rho$. This guarantees that assumptions $(i), (iii)$ of \cref{Prop:ConvergenceOja} are verified. 

For assumption $(ii)$, we have the expression:
\begin{equation*}
    \bW_0^\top \bz_i^+ = \sqrt{\frac{2\rho(\rho + \gamma_i)^2}{\sigma_i^2 + (\rho + \gamma_i)^2}} \bR^\top \bv_i. 
\end{equation*}
By point $(ii)$ in \cref{Lemma:GenericPositionStiefel}, with probability one, $\bR^\top \bv_1, \dots, \bR^\top \bv_m$ is a basis of $\R^m$. This confirms assumption $(ii)$, and we have the limit:
\begin{equation} \label{eq:LimitUnderparam}
    \bY(t) \xrightarrow[t \to \infty]{} 2 \sum_{i=1}^m \gamma_i \bz_i^+ (\bz_i^+)^\top. 
\end{equation}

\subsubsection{Identification of the Block Matrices}

Recall that:
\begin{equation*}
    \bY(t) = \begin{pmatrix}
        \bU(t) \bU(t)^\top & \bU(t) \bV(t)^\top \\ \bV(t) \bU(t)^\top & \bV(t) \bV(t)^\top
    \end{pmatrix}. 
\end{equation*}
Due to the expression of $\bz_i^+$ in \cref{prop:Diagonalization}, we have the identity:
\begin{equation*}
    \bz_i^+(\bz_i^+)^\top = \frac{1}{2\gamma_i} \begin{pmatrix}
        (\gamma_i - \rho) \bu_i \bu_i^\top & \sigma_i \bu_i \bv_i^\top \\ \sigma_i \bv_i \bu_i^\top & (\gamma_i + \rho) \bv_i\bv_i^\top
    \end{pmatrix}.
\end{equation*}
Using the block structure of the projection $\bPi$ in equation \eqref{eq:BlockProjection} and the limit of equations \eqref{eq:LimitOverparam}, \eqref{eq:LimitUnderparam}, we end up with:
\begin{equation*}
\begin{aligned}
    \bU(t) \bU(t)^\top &\xrightarrow[t \to \infty]{} \sum_{i=1}^{\min(m,m^*)} (\gamma_i - \rho) \bu_i \bu_i^\top, \\
    \bV(t) \bV(t)^\top &\xrightarrow[t \to \infty]{} \sum_{i=1}^{\min(m,m^*)} (\gamma_i + \rho) \bv_i \bv_i^\top + 2 \rho \1_{m > m^*} \overline \bPi, \\
    \bU(t) \bV(t)^\top &\xrightarrow[t \to \infty]{} \sum_{i=1}^{\min(m,m^*)} \sigma_i \bu_i \bv_i^\top.
\end{aligned}
\end{equation*}
This is precisely \cref{Prop:LimitBlockMatrices}.

\section{High-Dimensional Limit} \label{App:HighDim}
This appendix proves the high-dimensional results of \cref{Sec:HighDim}. We start by establishing three random matrix convergences (\cref{App:Subsec:RMTTheorem}) and apply them to the Stieltjes transforms and their long-time limits (\cref{App:Subsec:Stieltjes}). We then give an additional result for two-point Stieltjes transforms (\cref{App:Subsec:DoubleStieltjes}) and prove the range-projector formula (\cref{App:Subsec:Projectors1}).

\subsection{Random Matrix Results} \label{App:Subsec:RMTTheorem}

In this section, we are interested in the high-dimensional limit of quantities of the form:
\begin{equation} \label{eq:DefTrace1}
    T_d(a,b) = \frac{1}{d} \tr \Big( \bX^\top b(\bD) \bX \big( \bX^\top a(\bD) \bX \big)^{-1} \Big),
\end{equation}
where $\bX$ is uniformly drawn on $\mathrm{St}(d, m)$, $\bD$ is deterministic and symmetric, and the functions of $\bD$ are applied spectrally. We prove three high-dimensional limits: we study the quantity \eqref{eq:DefTrace1} and derive its high-dimensional limit whenever $a$ is real positive (\cref{Thm:RMTConvergenceReal}) or when it is complex (\cref{Thm:RMTConvergence}), in which case the previous quantity has a Stieltjes-like structure. Finally, in \cref{Thm:RMTConvergence2}, we derive the limit of a second-order trace quantity. All these results follow from the deterministic equivalents for sketched pseudoinverses of \citet{lejeune2024asymptotics}.

\begin{theorem} \label{Thm:RMTConvergenceReal}
    Let $\bX \in \R^{d\times m}$ be uniformly distributed on $\mathrm{St}(d,m)$. Let $\bD\in\mathcal S_d(\R)$ be deterministic, with spectrum contained in a compact set $S \subset \R$ independent of $d$, and assume that its empirical spectral measure $\mu_d$ converges weakly to a probability measure $\mu$. Moreover, assume that $m/d \xrightarrow[d \to \infty]{} \kappa \in (0, 1)$.

    Let $a : S \to \R^+$ and $b : S \to \mathbb{C}$ be continuous. Assume that:
    \begin{equation} \label{eq:SemidefiniteKappaCriterion}
        \mu( \{x : a(x) > 0 \}) > \kappa,
    \end{equation}
    and that $\inf_{d \geq 1} \lambda_{\min}^+ (a(\bD)) > 0$, where $\lambda_{\min}^+$ denotes the smallest non-zero eigenvalue. Let $h > 0$ be the unique solution of
    \begin{equation} \label{eq:SemidefiniteSC}
        1=\int \frac{\mu(\d x)}{1-\kappa+h a(x)}.
    \end{equation}
    Then, with probability one,
    \begin{equation} \label{eq:SemidefiniteCompressionLimit}
        \frac{1}{d}\tr\left(
        \bX^\top b(\bD)\bX
        \big(\bX^\top a(\bD)\bX\big)^{-1}
        \right)
        \xrightarrow[d\to\infty]{}
        h\int \frac{b(x)}{1-\kappa+h a(x)}\mu(\d x).
    \end{equation}
\end{theorem}
\begin{proof}
    First, the solution of equation \eqref{eq:SemidefiniteSC} exists and is unique. Indeed, the right-hand side as a function of $h$ is continuous and strictly decreasing on $\R^+$, is equal to $(1-\kappa)^{-1}>1$ at $h=0$, and converges as $h\to\infty$ to $\mu(\{a=0\})/(1-\kappa)<1$ by the assumption $\mu(\{a>0\})>\kappa$. By linearity, it is enough to prove the convergence for real-valued $b$.

    Consider $T_d(a,b)$ as defined in equation \eqref{eq:DefTrace1}. We first show that $\bX$ can be replaced by a Gaussian matrix. Let $\tilde\bG\in\R^{d\times m}$ have i.i.d. $\N(0,1/m)$ entries, independent of $\bX$, and set $\bG=\bX(\tilde\bG^\top\tilde\bG)^{1/2}$. Since $\tilde\bG(\tilde\bG^\top\tilde\bG)^{-1/2}$ is uniform on $\mathrm{St}(d,m)$ \citep[Theorem 2.2.1]{chikuse2003statistics} and independent of $\tilde\bG^\top\tilde\bG$, $\bG$ has the same law as $\tilde\bG$, and $\bX=\bG(\bG^\top\bG)^{-1/2}$. Therefore:
    \begin{equation*}
        T_d(a,b) = \frac{1}{d} \tr \Big(\bG^\top b(\bD) \bG
        \big(\bG^\top a(\bD) \bG \big)^{-1} \Big).
    \end{equation*}
    Since $a$ is continuous, the set $\{a>0\}$ is open, and weak convergence gives
    \begin{equation*}
        \liminf_{d\to\infty} \frac{\operatorname{rank}(a(\bD))}{d} = \liminf_{d\to\infty} \mu_d(\{a>0\}) \geq \mu(\{a>0\}) > \kappa.
    \end{equation*}
    Therefore \citet[Theorem~4.1(ii)]{lejeune2024asymptotics} applies at regularization parameter $\lambda = 0$ to the matrix $a(\bD)$. Thus, in the sense of deterministic equivalents of \citet[Sec.~2]{lejeune2024asymptotics},
    \begin{equation*}
        \bG \big(\bG^\top a(\bD) \bG\big)^{-1} \bG^\top \simeq
        \big(a(\bD) + \nu_d \bI_d \big)^{-1},
    \end{equation*}
    where $\nu_d>0$ is determined by
    \begin{equation} \label{eq:SCnud}
        \frac{m}{d} = \frac{1}{d} \tr \left( a(\bD) \big(a(\bD)+ \nu_d \bI_d \big)^{-1} \right).
    \end{equation}
    By the definition of deterministic equivalents, the convergence is almost sure when testing along the matrix $d^{-1} b(\bD)$, since
    \begin{equation*}
        \sup_{d \geq 1} \frac{1}{d} \tr \, \big|b(\bD) \big| \leq \sup_{x \in S} | b(x) | < \infty. 
    \end{equation*}
    Therefore, almost surely,
    \begin{equation} \label{eq:DeterministicEquivalent}
        \frac{1}{d} \tr \left(\bX^\top b(\bD) \bX \big(\bX^\top a(\bD) \bX \big)^{-1} \right) - \frac{1}{d} \tr \left(b(\bD) \big(a(\bD) + \nu_d \bI_d \big)^{-1} \right) \xrightarrow[d\to\infty]{\mathrm{a.s.}} 0.
    \end{equation}

    Now, the equation for $\nu_d$ can be rewritten as
    \begin{equation*}
        \frac{m}{d} = \int \frac{a(x)}{a(x) + \nu_d} \, \mu_d(\d x).
    \end{equation*}
    Now it is clear that $\nu_d > 0$ remains bounded in $(0, \infty)$: if it had a subsequence diverging to $+\infty$, the RHS would converge to zero along this subsequence, leading to a contradiction. Additionally, $\nu_d$ cannot have a subsequence going to zero. Otherwise, for any fixed $\nu > 0$, we would have $\nu_d < \nu$ along this subsequence for $d$ large enough, hence:
    \begin{equation*}
        \frac{m}{d} \geq \int \frac{a(x)}{a(x)+\nu} \, \mu_d(\d x) \xrightarrow[d\to\infty]{} \int \frac{a(x)}{a(x)+\nu} \, \mu(\d x),
    \end{equation*}
    and letting $\nu \to 0$ would give $\kappa \geq \mu(\{a>0\})$ and contradict equation \eqref{eq:SemidefiniteKappaCriterion}. Hence taking a convergent subsequence of $(\nu_d)_{d \geq 1}$ and using that $\mu_d \Rightarrow \mu$ gives the equation:
    \begin{equation} \label{eq:SCnu}
        \kappa = \int \frac{a(x)}{a(x) + \nu} \mu(\d x).
    \end{equation}
    The right-hand side is strictly decreasing in $\nu$, so this equation has a unique positive solution. Therefore $\nu_d$ converges to $\nu$.
    Setting $h=(1 - \kappa) / \nu$ gives equation \eqref{eq:SemidefiniteSC}. Finally, as a consequence of the deterministic equivalent in equation \eqref{eq:DeterministicEquivalent}, we have the almost sure convergence:
    \begin{equation*}
        \frac{1}{d} \tr \left(\bX^\top b(\bD) \bX \big(\bX^\top a(\bD) \bX \big)^{-1} \right) \xrightarrow[d \to \infty]{} \int \frac{b(x)}{a(x) + \nu} \mu(\d x) = h \int \frac{b(x)}{1 - \kappa + h a(x)}\mu(\d x),
    \end{equation*}
    which proves \eqref{eq:SemidefiniteCompressionLimit}.
\end{proof}

We shall now move on to the complex version of the previous result. Before stating it, we prove a first lemma that derives uniform bounds for traces of complex matrices.
\begin{lemma} \label{Lemma:RMTBounds}
    Let $\bX_0 \in \R^{d\times m}$ be such that $\bX_0^\top \bX_0 = \bI_m$. Let $\bD \in \mathcal S_d(\R)$ have spectrum contained in a compact set $S \subset \R$ independent of $d$. Additionally, for $\zeta \in \mathbb{C} \setminus \R^+$, consider $a_\zeta(x) = q(x) - \zeta r(x)$, where $q, r \colon S \to \R^+$ are continuous and $q(x) \geq \eta > 0$ on $S$. Then, given $K$ a compact set of $\mathbb{C} \setminus \R^+$, we have:
    \begin{equation*}
        \sup_{\zeta \in K} \sup_{d \geq 1} \frac{1}{d} \left| \tr \Big( \big(\bX_0^\top a_\zeta(\bD) \bX_0 \big)^{-1} \Big) \right| < \infty, \hspace{1.5cm} \sup_{\zeta \in K} \sup_{d \geq 1} \frac{1}{d} \Big\| \big(\bX_0^\top a_\zeta(\bD) \bX_0 \big)^{-1} \Big\|_F^2 < \infty,
    \end{equation*}
    where $\| \, \cdot \, \|_F$ is the Frobenius norm on $\mathbb{C}^{m \times m}$. 
\end{lemma}

\begin{proof}
    We let:
    \begin{equation*}
        \bM = \bX_0^\top q(\bD) \bX_0, \hspace{1.5cm} \bN = \bM^{-1/2} \bX_0^\top r(\bD) \bX_0 \bM^{-1/2}.
    \end{equation*}
    Both of these matrices belong to $\mathcal{S}_m^+(\R)$. Since $q \geq \eta$ on $S$ and $\bX_0^\top \bX_0 = \bI_m$, we have $\bM \succeq \eta \bI_m$. Moreover $\bN \succeq 0$, because $r \geq 0$ on $S$. Additionally, for every $\zeta \in \mathbb{C} \setminus \R^+$, we have:
    \begin{equation*}
        \bX_0^\top a_\zeta(\bD)\bX_0 = \bM^{1/2}(\bI_m - \zeta \bN)\bM^{1/2}.
    \end{equation*}
    Now, define:
    \begin{equation*}
        c_K = \inf_{\zeta \in K}\inf_{\lambda \geq 0}|1-\zeta \lambda|.
    \end{equation*}
    Since $K$ is compact in $\mathbb{C}\setminus\R^+$, this constant is positive. Hence, for $\zeta \in K$, we have:
    \begin{equation*}
        \|(\bI_m - \zeta \bN)^{-1} \|_{\text{op}} \leq c_K^{-1},
    \end{equation*}
    where $\| \, \cdot \, \|_{\text{op}}$ is the operator norm. Therefore:
    \begin{equation*}
        \Big\|\big(\bX_0^\top a_\zeta(\bD) \bX_0 \big)^{-1}\Big\|_{\text{op}} \leq \big\| \bM^{-1/2} \big\|_{\text{op}}^2 \big\|(\bI_m - \zeta \bN)^{-1} \big\|_{\text{op}} \leq \frac{1}{\eta c_K}.
    \end{equation*}
    It follows that:
    \begin{equation*}
        \frac{1}{d} \left| \tr \Big( \big(\bX_0^\top a_\zeta(\bD) \bX_0 \big)^{-1}\Big) \right| \leq \frac{m}{d} \Big\| \big(\bX_0^\top a_\zeta(\bD) \bX_0 \big)^{-1}\Big\|_{\text{op}} \leq
        \frac{1}{\eta c_K},
    \end{equation*}
    because $m \leq d$. Similarly,
    \begin{equation*}
        \frac{1}{d} \Big\| \big(\bX_0^\top a_\zeta(\bD) \bX_0 \big)^{-1} \Big\|_F^2 \leq \frac{m}{d} \Big\| \big(\bX_0^\top a_\zeta(\bD)\bX_0 \big)^{-1} \Big\|_{\text{op}}^2 \leq \frac{1}{\eta^2 c_K^2}.
    \end{equation*}
    Both bounds are uniform over $\zeta \in K$, $d \geq 1$, and $\bX_0$.
\end{proof}

We now state the complex version of \cref{Thm:RMTConvergenceReal}.

\begin{theorem} \label{Thm:RMTConvergence}
    Let $\bX \in \R^{d\times m}$ be uniformly distributed on $\mathrm{St}(d,m)$. Let $\bD\in\mathcal S_d(\R)$ be deterministic, with spectrum contained in a compact set $S \subset \R$ independent of $d$, and assume that its empirical spectral measure $\mu_d$ converges weakly to a probability measure $\mu$. Moreover, assume that $m/d \xrightarrow[d \to \infty]{} \kappa \in (0, 1)$.

    Let $q, r : S \to \R^+$ be continuous functions such that $q(x) \geq \eta > 0$ on $S$. For $\zeta \in \mathbb{C}\setminus \R^+$, set:
    \begin{equation*}
        a_\zeta(x) = q(x) - \zeta r(x).
    \end{equation*}
    Then, there is a unique holomorphic function $h : \mathbb{C}\setminus \R^+ \to \mathbb{C}$ such that $h(\zeta) > 0$ for $\zeta < 0$ and
    \begin{equation} \label{eq:RMTSelfConsistentStieltjes}
        1 = \int \frac{\mu(\d x)}{1 - \kappa + h(\zeta) a_\zeta(x)}.
    \end{equation}
    Let $b : S \to \mathbb{C}$ be continuous. Then, with probability one, for every compact set $K \subset \mathbb{C} \setminus \R^+$,
    \begin{equation} \label{eq:RMTConvergenceStieltjes}
        \sup_{\zeta \in K} \left| \frac{1}{d} \tr \Big( \bX^\top b(\bD)\bX
        \big(\bX^\top a_\zeta(\bD) \bX \big)^{-1} \Big) - h(\zeta) \int \frac{b(x)}{1 - \kappa + h(\zeta) a_\zeta(x)} \mu(\d x) \right| \xrightarrow[d \to \infty]{} 0.
    \end{equation}
\end{theorem}

\begin{proof}
    Let us first prove the claim on the function $h(\zeta)$. For fixed $\zeta < 0$, the function $a_\zeta$ is positive and satisfies the assumptions of \cref{Thm:RMTConvergenceReal}. Denoting by $h_+(\zeta)$ the corresponding positive solution of equation \eqref{eq:RMTSelfConsistentStieltjes}, \cref{Thm:RMTConvergenceReal} gives the desired convergence on the negative real axis for this fixed $\zeta$ and every fixed continuous $b$. In particular, when $b=1$, $h_+(\zeta)$ is obtained as the limit of
    \begin{equation*}
        H_d(\zeta) := T_d(a_\zeta, 1) = \frac{1}{d} \tr \Big( \big(\bX^\top a_\zeta(\bD) \bX \big)^{-1} \Big),
    \end{equation*}
    where we exploited that $\bX^\top \bX = \bI_m$. Then, for every compact set $K \subset \mathbb{C} \setminus \R^+$, \cref{Lemma:RMTBounds} guarantees that:
    \begin{equation*}
        \sup_{\zeta \in K} \sup_{d \geq 1} \big| H_d(\zeta) \big| < \infty.
    \end{equation*}
    Let $D^-$ be a dense countable subset of $(-\infty,0)$. Since \cref{Thm:RMTConvergenceReal} guarantees convergence almost surely for each fixed $\zeta\in D^-$, it holds simultaneously on $D^-$ on an event of probability one. On this event, \cref{Thm:VitaliPorter} gives convergence of $H_d$ to a holomorphic function $h$ on $\mathbb{C}\setminus\R^+$, and $h=h_+$ on $D^-$. 
    
    Let us now show that $h$ satisfies equation \eqref{eq:RMTSelfConsistentStieltjes} everywhere. To do so, we show that the RHS of equation \eqref{eq:RMTSelfConsistentStieltjes} is holomorphic in $\zeta$ and use the identity theorem \cref{Thm:IdentityTheorem}. Recall that $\Im$ denotes the imaginary part, and denote by $\bar z$ the complex conjugate of $z \in \mathbb{C}$. Let us consider $\zeta \in \mathbb{C}$ such that $\Im \zeta > 0$. Then:
    \begin{equation*}
        \Im H_d(\zeta) = \frac{\Im\zeta}{d}\tr\Big(\big(\bX^\top a_\zeta(\bD)\bX\big)^{-1}\,\bX^\top r(\bD)\bX\,\big(\bX^\top a_{\bar\zeta}(\bD)\bX\big)^{-1}\Big) \geq 0,
    \end{equation*}
    since $\bX^\top r(\bD)\bX \succeq 0$, and in the limit we also have $\Im h(\zeta) \geq 0$. By the open mapping theorem \citep[][Theorem 7.1]{bak2010complex}, either $\Im h(\zeta) > 0$ on the upper half-plane, or $h$ is constant and real. This is the case if $r = 0$ $\mu$-almost everywhere. This constant then can be found by taking the limit on the negative real axis, and therefore is positive. For $\Im \zeta < 0$, since $H_d(\bar\zeta) = \overline{H_d(\zeta)}$, we get by conjugation that either $\Im h(\zeta) < 0$ or $h$ is a positive constant.

    With these conclusions, let us study the integral in equation \eqref{eq:RMTSelfConsistentStieltjes}, applied to the function $h$. We first show that, for every compact set $K \subset \mathbb{C} \setminus \R^+$,
    \begin{equation} \label{eq:BoundDenominator}
        \inf_{\zeta \in K} \inf_{x \in S} \big| 1 - \kappa + h(\zeta) a_\zeta(x) \big| > 0.
    \end{equation}
    For $\zeta<0$, both $h(\zeta)$ and $a_\zeta(x)$ are positive, so the denominator is positive. If $\Im\zeta\ne0$ and the denominator vanished, then $h(\zeta)=-(1-\kappa)/(q(x)-\zeta r(x))$. If $r(x)>0$, this implies $\Im h(\zeta)\Im\zeta<0$. If $r(x)=0$, then $h(\zeta)=-(1-\kappa)/q(x)$ is a negative real number. Both cases are excluded, since we have shown that either $\Im h(\zeta)\Im\zeta>0$, or $h$ is a positive constant. Thus the denominator never vanishes. Since $K$ and $S$ are both compact, it is easily seen that the uniform bound in equation \eqref{eq:BoundDenominator} holds. Therefore, the function:
    \begin{equation*}
        \zeta \longmapsto \int \frac{\mu(\d x)}{1 - \kappa + h(\zeta) a_\zeta(x)}
    \end{equation*}
    is holomorphic (since $h$ is and the denominator is uniformly bounded away from zero) on $\mathbb{C} \setminus \R^+$. As it is equal to $1$ on $D^-$, the identity theorem (\cref{Thm:IdentityTheorem}) shows that it is equal to $1$ on $\mathbb{C} \setminus \R^+$. Uniqueness of the holomorphic solution on $\mathbb{C} \setminus \R^+$ follows from the uniqueness on $(-\infty, 0)$ and the same theorem. 

    Let us now prove the convergence in equation \eqref{eq:RMTConvergenceStieltjes}. We start by fixing $b : S \to \mathbb{C}$ continuous. As previously, since $D^-$ is countable, there is a probability one event for which $T_d(a_\zeta, b)$ converges to its desired limit for all $\zeta \in D^-$. Moreover, as a consequence of the Cauchy--Schwarz inequality and the fact that $\bX^\top \bX = \bI_m$,
    \begin{equation*}
    \begin{aligned}
        \Big| T_d(a_\zeta, b) \Big| &= \frac{1}{d} \left| \tr \Big( b(\bD) \bX \big( \bX^\top a_\zeta(\bD) \bX \big)^{-1} \bX^\top \Big) \right| \\
        &\leq \sqrt{\frac{1}{d} \big\| b(\bD) \big\|_F^2} \sqrt{\frac{1}{d} \Big\| \big( \bX^\top a_\zeta(\bD) \bX \big)^{-1} \Big\|_F^2}. 
    \end{aligned}
    \end{equation*}
    Then, \cref{Lemma:RMTBounds} guarantees that $\zeta \mapsto T_d(a_\zeta, b)$ is holomorphic and locally bounded on $\mathbb{C} \setminus \R^+$, since $\bX^\top a_\zeta(\bD)\bX$ is affine in $\zeta$ and invertible there. Additionally, equation \eqref{eq:BoundDenominator} guarantees that the limiting function
    \begin{equation*}
        \zeta \longmapsto h(\zeta) \int \frac{b(x)}{1 - \kappa + h(\zeta) a_\zeta(x)} \mu(\d x)
    \end{equation*}
    is holomorphic on $\mathbb{C} \setminus \R^+$. On the event of probability one chosen above, $T_d(a_\zeta, b)$ converges to this function for every $\zeta \in D^-$. Hence \cref{Cor:HolomorphicExtension} gives uniform convergence on every compact subset of $\mathbb{C}\setminus\R^+$, which proves equation \eqref{eq:RMTConvergenceStieltjes}.
\end{proof}

In the following theorem, we consider the case $r = 0$, corresponding to the case where $h$ is real, and derive the high-dimensional limit for a second-order version of the previous trace.

\begin{theorem} \label{Thm:RMTConvergence2}
    Under the assumptions of \cref{Thm:RMTConvergence}, with $r = 0$, let $h$ be the unique positive solution of equation \eqref{eq:RMTSelfConsistentStieltjes} and $b_1,b_2 : S \to \R$ be continuous, and define:
    \begin{equation*}
        \bH = \bX \big(\bX^\top q(\bD)\bX \big)^{-1}\bX^\top.
    \end{equation*}
    Then:
    \begin{equation} \label{eq:RMTConvergence2}
    \begin{aligned}
        \frac{1}{d} \tr \Big(b_1(\bD) \bH b_2(\bD) \bH \Big) &\xrightarrow[d\to\infty]{\mathrm{a.s.}} \frac{(1-\kappa)h}{J} \left(\int \frac{b_1(x)}{\Delta(x)^2}\mu(\d x)\right) \left(\int \frac{b_2(x)}{\Delta(x)^2}\mu(\d x)\right) \\
        &\hspace{1cm} + h^2 \int \frac{b_1(x)b_2(x)}{\Delta(x)^2}\mu(\d x),
    \end{aligned}
    \end{equation}
    where
    \begin{equation*}
        \Delta(x) = 1 - \kappa + h q(x), \hspace{1.5cm} J = \int \frac{q(x)}{\Delta(x)^2} \mu(\d x). 
    \end{equation*}
\end{theorem}

\begin{proof}
    As in the proof of \cref{Thm:RMTConvergenceReal}, we write $\bX=\bG(\bG^\top\bG)^{-1/2}$, where $\bG \in \R^{d \times m}$ has i.i.d. $\N(0,1/m)$ entries. Hence $\bH=\bG\big(\bG^\top q(\bD)\bG\big)^{-1}\bG^\top$. Since $q(\bD) \succeq \eta \bI_d$ has full rank and $\kappa < 1$, \citet[Theorem~4.8]{lejeune2024asymptotics} applies with $\lambda = 0$ and gives the deterministic equivalent:
    \begin{equation} \label{eq:DeterministicEquivalent1}
        \bH b_2(\bD) \bH \simeq \big( q(\bD) + \nu_d \bI_d \big)^{-1} \big( b_2(\bD) + \theta_d \bI_d \big) \big( q(\bD) + \nu_d \bI_d \big)^{-1}.
    \end{equation}
    As this deterministic equivalent requires $b_2(\bD)$ to be positive semidefinite, we prove it for $b_2 \geq 0$ and extend to a general continuous $b_2$ by decomposing $b_2 = \max(0, b_2) - \max(0, -b_2)$, and using the linearity of equation \eqref{eq:RMTConvergence2}. 

    Recall that $\nu_d$ is the unique positive solution of \eqref{eq:SCnud} and converges as $d \to \infty$ to $\nu$, solving \eqref{eq:SCnu}. Moreover, $h = (1 - \kappa) / \nu$, and $\theta_d$ is given by
    \begin{equation*}
        \theta_d = \frac{\nu_d \tr \Big( \big( q(\bD) + \nu_d \bI_d \big)^{-1} b_2(\bD) \big( q(\bD) + \nu_d \bI_d \big)^{-1} \Big)}{\tr \Big( q(\bD) \big( q(\bD) + \nu_d \bI_d \big)^{-2} \Big)}.
    \end{equation*}
    Now, we have the deterministic limits
    \begin{equation*}
    \begin{aligned}
        \frac{1}{d} \tr \Big( \big( q(\bD) + \nu_d \bI_d \big)^{-1} b_2(\bD) \big( q(\bD) + \nu_d \bI_d \big)^{-1} \Big) &\xrightarrow[d \to \infty]{} \int \frac{b_2(x)}{(q(x) + \nu)^2} \mu(\d x), \\
        \frac{1}{d} \tr \Big( q(\bD) \big( q(\bD) + \nu_d \bI_d \big)^{-2} \Big) &\xrightarrow[d \to \infty]{} \int \frac{q(x)}{\big(q(x) + \nu \big)^2} \mu(\d x). 
    \end{aligned}
    \end{equation*}
    The convergence is guaranteed since $\mu_d \Rightarrow \mu$ and $\mu_d$ is integrated against $d$-dependent functions that converge uniformly on $S$. Hence, $\theta_d$ converges as $d \to \infty$. Additionally, from the deterministic equivalent in equation \eqref{eq:DeterministicEquivalent1}, and the fact that
    \begin{equation*}
        \sup_{d \geq 1} \frac{1}{d} \tr \, \big|b_1(\bD) \big| \leq \sup_{x \in S} | b_1(x) | < \infty,
    \end{equation*} 
    we have the almost sure limit
    \begin{equation*}
        \frac{1}{d} \tr \Big( b_1(\bD) \bH b_2(\bD) \bH \Big) - \frac{1}{d} \tr \Big( b_1(\bD) \big( b_2(\bD) + \theta_d \bI_d \big) \big( q(\bD) + \nu_d \bI_d \big)^{-2} \Big) \xrightarrow[d \to \infty]{\mathrm{a.s.}} 0. 
    \end{equation*}
    To conclude, we have the limit for the deterministic trace:
    \begin{equation*}
        \frac{1}{d} \tr \Big( b_1(\bD) \big( b_2(\bD) + \theta_d \bI_d \big) \big( q(\bD) + \nu_d \bI_d \big)^{-2} \Big) \xrightarrow[d \to \infty]{} \int \frac{b_1(x) b_2(x)}{(q(x) + \nu)^2} \mu(\d x) + \theta \int \frac{b_1(x)}{(q(x) + \nu)^2} \mu(\d x). 
    \end{equation*}
    Combining the relationship between $\nu$ and $h$ with the previous equations gives the result. 
\end{proof}

\subsection{Stieltjes Transforms} \label{App:Subsec:Stieltjes}

In this section we study the Stieltjes transforms $\mathfrak m_t^U, \mathfrak m_t^V, \mathfrak m_t^Z$ defined in equations \eqref{eq:DefStieltjes}, \eqref{eq:DefStieltjesZ} and prove the results stated in \cref{Subsec:StieltjesTransforms}. 

\subsubsection{High-Dimensional Limit} \label{App:Subsubsec:StieltjesProof}

We now prove \cref{Thm:Stieltjes} which derives the high-dimensional limit of the Stieltjes transforms.

\paragraph{Step 1. Stieltjes transforms of the Gram matrices.} To start, we prove the high-dimensional limit for $\mathfrak m_t^U, \mathfrak m_t^V$. Let
\begin{equation*}
    \bS = \big(\bC^\top \bC\big)^{1/2}, \hspace{1.5cm} \bGamma_\rho = \big(\bS^2 + \rho^2 \bI_d\big)^{1/2}.
\end{equation*}
We define:
\begin{equation*}
\begin{aligned}
    \bK(t) &= \bR^\top q_t \big( \bGamma_\rho \big) \bR \hspace{2cm} && \in \mathcal{S}_m^{++}(\R), \\
    \bB_U(t) &= \sqrt{2\rho} \chi_t^U(\bC) \bR && \in \R^{p \times m}, \\
    \bB_V(t) &= \sqrt{2\rho} \chi_t^V(\bS) \bR && \in \R^{d \times m}, 
\end{aligned}
\end{equation*}
so that, as a consequence of \cref{Lemma:BlockDecomposition1}
\begin{equation*}
\begin{aligned}
    \bU(t) \bU(t)^\top &= \bB_U(t) \bK(t)^{-1} \bB_U(t)^\top, \\
    \bV(t) \bV(t)^\top &= \bB_V(t) \bK(t)^{-1} \bB_V(t)^\top.
\end{aligned}
\end{equation*}
Now using the Woodbury matrix identity, we have for $w\in \mathbb{C} \setminus \R^+$,
\begin{equation*}
\begin{aligned}
    \big(\bU(t) \bU(t)^\top -w \bI_p \big)^{-1} &= - \frac{1}{w} \bI_p - \frac{1}{w^2} \bB_U(t) \left( \bK(t) - \frac{1}{w} \bB_U(t)^\top \bB_U(t) \right)^{-1} \bB_U(t)^\top, \\
    \big(\bV(t) \bV(t)^\top -w \bI_d \big)^{-1} &= - \frac{1}{w} \bI_d - \frac{1}{w^2} \bB_V(t) \left( \bK(t) - \frac{1}{w} \bB_V(t)^\top \bB_V(t) \right)^{-1} \bB_V(t)^\top.
\end{aligned}
\end{equation*}
We now state a useful lemma.
\begin{lemma} \label{Lemma:RectangulartoSquare}
    Let $\bS=(\bC^\top\bC)^{1/2}$. For any $\psi, G \colon \R \to \R$ such that $G(0) = 0$, we have:
    \begin{equation*}
        G(\bC)^\top \psi \big( \bC \bC^\top \big) G(\bC) = G(\bS)^2 \psi \big( \bS^2 \big). 
    \end{equation*}
\end{lemma}

\begin{proof}
    Write $\bC=\sum_{i=1}^{m^*}\sigma_i \bu_i \bv_i^\top$ over the positive singular values. Then, we have
    \begin{equation*}
        G(\bC) = \sum_{i=1}^{m^*}G(\sigma_i)\bu_i\bv_i^\top, \hspace{1.5cm} \psi(\bC \bC^\top) = \sum_i \psi(\sigma_i^2) \bu_i \bu_i^\top + \psi(0) \bP_\text{Ker},
    \end{equation*}
    where $\bP_\text{Ker}$ is the orthogonal projection onto $\mathrm{Ker}(\bC\bC^\top)$. This projection vanishes after multiplication by $G(\bC)$, which gives
    \begin{equation*}
        G(\bC)^\top \psi(\bC\bC^\top)G(\bC)
        =\sum_{i=1}^{m^*}G(\sigma_i)^2\psi(\sigma_i^2)\bv_i\bv_i^\top
        =G(\bS)^2\psi(\bS^2).
    \end{equation*}
\end{proof}
Therefore, since $\chi_t^U(0)=0$, this gives:
\begin{equation*}
\begin{aligned}
    \bB_U(t)^\top \bB_U(t) &= 2 \rho \bR^\top \chi_t^U(\bS)^2 \bR, \\
    \bB_V(t)^\top \bB_V(t) &= 2 \rho \bR^\top \chi_t^V(\bS)^2 \bR.
\end{aligned}
\end{equation*}
Now, we have
\begin{equation*}
\begin{aligned}
    \bK(t) - \frac{1}{w} \bB_U(t)^\top \bB_U(t) &= \bR^\top \left( q_t \big( \bGamma_\rho \big) - \frac{2 \rho}{w} \chi_t^U(\bS)^2 \right) \bR, \\
    \bK(t) - \frac{1}{w} \bB_V(t)^\top \bB_V(t) &= \bR^\top \left( q_t \big( \bGamma_\rho \big) - \frac{2 \rho}{w} \chi_t^V(\bS)^2 \right) \bR.
\end{aligned}
\end{equation*}
Everything is now expressed in terms of $\bR$ and the deterministic symmetric matrix $\bS = (\bC^\top \bC)^{1/2}$. Using the functions defined in \cref{Thm:Stieltjes}, we have
\begin{equation*}
    2 \rho \chi_t^U(\bS)^2 = r_t^\mathsf{L} \big( \bGamma_\rho \big), \hspace{1.5cm} 2 \rho \chi_t^V(\bS)^2 = r_t^\mathsf{R}(\bGamma_\rho).
\end{equation*}
Then, putting everything together, we arrive at the expressions:
\begin{equation*}
\begin{aligned}
    \mathfrak m_t^U(w) &= - \frac{1}{w} - \frac{1}{w^2 p} \tr \Bigg[ \left( \bR^\top \left( q_t \big( \bGamma_\rho \big) - \frac{1}{w} r_t^\mathsf{L} \big( \bGamma_\rho \big) \right) \bR \right)^{-1} \bR^\top r_t^\mathsf{L}(\bGamma_\rho) \bR \Bigg], \\
    \mathfrak m_t^V(w) &= - \frac{1}{w} - \frac{1}{w^2 d} \tr \Bigg[ \left(\bR^\top \left( q_t \big( \bGamma_\rho \big) - \frac{1}{w} r_t^\mathsf{R} \big( \bGamma_\rho \big) \right) \bR \right)^{-1} \bR^\top r_t^\mathsf{R} \big( \bGamma_\rho \big) \bR \Bigg]. 
\end{aligned}
\end{equation*}
The empirical spectral distribution of $(\bC^\top \bC)^{1/2}$ converges to $\tilde \mu_C = (1 - \beta) \delta_0 + \beta \mu_C$, with $\beta = \min(1, \tau)$, and $\tau = \lim_{d \to \infty} p/d$. To conclude, we apply \cref{Thm:RMTConvergence}, with $\zeta = w^{-1} \in \mathbb{C} \setminus \R^+$. We get the almost sure limits: 
\begin{equation*}
\begin{aligned}
    \mathfrak m_t^U(w) &\xrightarrow[d \to \infty]{\mathrm{a.s.}} -\frac{1}{w} - \frac{1}{\tau w^2} h_t^\mathsf{L}(w) \int \frac{r_t^\mathsf{L}(\gamma_x)}{1 - \kappa + h_t^\mathsf{L}(w)\left(q_t(\gamma_x)- \frac{1}{w} r_t^\mathsf{L}(\gamma_x) \right)} \tilde \mu_C(\d x), \\
    \mathfrak m_t^V(w) &\xrightarrow[d \to \infty]{\mathrm{a.s.}} -\frac{1}{w} - \frac{1}{w^2} h_t^\mathsf{R}(w) \int \frac{r_t^\mathsf{R}(\gamma_x)}{1 - \kappa + h_t^\mathsf{R}(w) \left(q_t(\gamma_x)- \frac{1}{w} r_t^\mathsf{R}(\gamma_x) \right)} \tilde \mu_C(\d x),
\end{aligned}
\end{equation*}
where $\gamma_x = \sqrt{x^2 + \rho^2}$ and $h_t^\mathsf{L}, h_t^\mathsf{R}$ are the unique holomorphic functions on $\mathbb{C} \setminus \R^+$, positive for $w<0$, solving
\begin{equation*}
    1 = \int \frac{\tilde \mu_C(\d x)}{1 - \kappa + h_t^\mathsf{a}(w) \left(q_t(\gamma_x)- \frac{1}{w} r_t^\mathsf{a}(\gamma_x) \right)},
\end{equation*}
for $\mathsf a=\mathsf L,\mathsf R$. Finally, define
\begin{equation*}
    g_t^\mathsf{L}(w) = -\frac{1}{\kappa w} h_t^\mathsf{L}(w), \hspace{1.5cm} g_t^\mathsf{R}(w) = -\frac{1}{\kappa w} h_t^\mathsf{R}(w),
\end{equation*}
so that both functions are positive for $w < 0$. Their self-consistent equations are
\begin{equation*}
    1 = \int \frac{\tilde \mu_C(\d x)}{1 - \kappa + \kappa g_t^\mathsf{a}(w) \big( r_t^\mathsf{a}(\gamma_x) - w q_t(\gamma_x) \big)}. 
\end{equation*}
Then, using that:
\begin{equation*}
    \frac{\kappa g_t^\mathsf{a}(w) r_t^\mathsf{a}(\gamma_x)}{1 - \kappa + \kappa g_t^\mathsf{a}(w) \big( r_t^\mathsf{a}(\gamma_x) - w q_t(\gamma_x) \big)} = 1 - \frac{1 - \kappa - \kappa w g_t^\mathsf{a}(w) q_t(\gamma_x)}{1 - \kappa + \kappa g_t^\mathsf{a}(w) \big( r_t^\mathsf{a}(\gamma_x) - w q_t(\gamma_x) \big)},
\end{equation*}
and the self-consistent equation for $g_t^\mathsf{a}$, we get
\begin{equation*}
\begin{aligned}
    \mathfrak m_t^U(w) &\xrightarrow[d \to \infty]{\mathrm{a.s.}} - \frac{1}{w} \left(1 - \frac{\kappa}{\tau} \right) + \frac{\kappa}{\tau} g_t^\mathsf{L}(w) \int \frac{q_t(\gamma_x)}{1 - \kappa + \kappa g_t^\mathsf{L}(w) \big( r_t^\mathsf{L}(\gamma_x) - w q_t(\gamma_x) \big)} \tilde \mu_C(\d x), \\
    \mathfrak m_t^V(w) &\xrightarrow[d \to \infty]{\mathrm{a.s.}} - \frac{1 - \kappa}{w} + \kappa g_t^\mathsf{R}(w) \int \frac{q_t(\gamma_x)}{1 - \kappa + \kappa g_t^\mathsf{R}(w) \big( r_t^\mathsf{R}(\gamma_x) - w q_t(\gamma_x) \big)} \tilde \mu_C(\d x). 
\end{aligned}
\end{equation*}
To conclude, we use the identity:
\begin{equation*}
    r_t^\mathsf{R}(\gamma) = r_t^\mathsf{L}(\gamma) + 2 \rho q_t(\gamma),
\end{equation*}
hence $g_t^\mathsf{R}$ solves the equation
\begin{equation*}
    1 = \int \frac{\tilde \mu_C(\d x)}{1 - \kappa + \kappa g_t^\mathsf{R}(w) \big( r_t^\mathsf{L}(\gamma_x) + (2 \rho - w) q_t(\gamma_x) \big)}. 
\end{equation*}
Both $g_t^\mathsf{R}(w)$ and $g_t^\mathsf{L}(w-2\rho)$ are positive for $w<0$. By uniqueness of the holomorphic solution with this property, we obtain
\begin{equation*}
    g_t^\mathsf{R}(w) = g_t^\mathsf{L}(w - 2 \rho). 
\end{equation*}
Writing $g_t=g_t^\mathsf{L}$ proves \eqref{eq:StieltjesU} and \eqref{eq:StieltjesV}. 

The identity \eqref{eq:LinkStieltjes} also follows directly from \cref{Prop:SVDPrediction}, where the non-zero eigenvalues of $\bV(t) \bV(t)^\top$ are directly related to those of $\bU(t) \bU(t)^\top$. Set $\kappa_d = m/d$ and $\tau_d = p/d$. For any test function $f$, we have the exact finite-dimensional relation
\begin{equation*}
    \int f(x) \mu_{t,d}^V(\d x) = (1 - \kappa_d) f(0) + (\kappa_d - \tau_d) f(2\rho) + \tau_d \int f(x + 2 \rho) \mu_{t,d}^U(\d x).
\end{equation*}
Taking $d\to\infty$ gives equation \eqref{eq:LinkStieltjes}. 

\paragraph{Step 2. Prediction matrix.} Let $\lambda_1(t), \dots, \lambda_p(t)$ be the full spectrum of $\bU(t)\bU(t)^\top$, including zero eigenvalues, and set $s_i(t) = \sqrt{\lambda_i(t)(\lambda_i(t) + 2 \rho)}$. By \cref{Prop:SVDPrediction}, these give the singular values of $\bZ(t)$ after adjusting for the rectangular dimensions.

Since we consider the $\min(p, d)$ singular values of $\bZ(t)$, there are two cases:
\begin{itemize}
    \item If $p \leq d$, then the singular values of $\bZ(t)$, denoted $\sigma_1(t), \dots, \sigma_p(t)$, are precisely $s_1(t), \dots, s_p(t)$. 
    \item If $p > d$, $\sigma_1(t), \dots, \sigma_d(t)$ are obtained by taking $s_1(t), \dots, s_p(t)$ (corresponding to the full spectrum of $\bU(t) \bU(t)^\top$) and removing $p-d$ zero eigenvalues. 
\end{itemize}
With $\beta_d=\min(1, p/d)$, this gives the exact finite-dimensional identity
\begin{equation} \label{eq:pushforwardUtoZ}
    \int f(x) \mu_{t,d}^Z (\d x) = \left(1 - \frac{\tau_d}{\beta_d} \right) f(0) + \frac{\tau_d}{\beta_d} \int f \left(\sqrt{x(x+2\rho)} \right) \mu_{t,d}^U(\d x),
\end{equation}
for every bounded continuous $f$. Taking the high-dimensional limit and using the pushforward characterization in \cref{App:Subsec:RMT} gives equation \eqref{eq:StieltjesUtoZ}. 

\subsubsection{Initialization} \label{App:Subsubsec:StieltjesInit}

As a consistency check, we recompute from \cref{Thm:Stieltjes} the expression of $\mathfrak m_t^U(w), \mathfrak m_t^V(w)$ at $t = 0$, leading to the spectral distributions of $\bU(0) \bU(0)^\top$ and $\bV(0) \bV(0)^\top$. 
\begin{lemma}
    At $t = 0$, \cref{Thm:Stieltjes} yields
    \begin{equation} \label{eq:StieltjesInit}
        \mathfrak m_0^U(w) = - \frac{1}{w}, \hspace{1.5cm} \mathfrak m_0^V(w) = - \frac{1 - \kappa}{w} + \frac{\kappa}{2 \rho - w}. 
    \end{equation}
    These are the Stieltjes transforms of the probability measures:
    \begin{equation} \label{eq:MeasureInit}
        \mu_0^U = \delta_0, \hspace{1.5cm} \mu_0^V = (1 - \kappa) \delta_0 + \kappa \delta_{2 \rho}.
    \end{equation}
\end{lemma}

\begin{proof}
    At $t = 0$, we have $q_0(\gamma) = 1$ and $r_0^\mathsf{L}(\gamma) = 0$. From equation \eqref{eq:SCgt}, we then get $g_0(w) = - w^{-1}$, and \cref{Thm:Stieltjes} gives the desired expressions. Their form identifies the corresponding Dirac masses.
\end{proof}

This result is consistent with \cref{Assumption1:Initialization}, where we chose $\bU$ to be initialized at zero, and $\bV(0) \bV(0)^\top = 2 \rho \bR \bR^\top$, where $\bR \bR^\top$ is an orthogonal projector of rank $m$, asymptotically $\kappa d$.

\subsubsection{Limiting Stieltjes Transforms} \label{App:Subsubsec:StieltjesLimit}

We now prove \cref{Cor:StieltjesInf}. 

\paragraph{Step 1. Gram matrices.} We start by proving the limit of $\mathfrak m_t^U, \mathfrak m_t^V$. We consider $\xi < 0$, and then extend the convergence by using \cref{Cor:HolomorphicExtension}. Now, $g_t(\xi)$ is the unique positive solution of:
\begin{equation} \label{eq:LongTimeSCPositive}
    1 = I_t(g) :=\int \frac{\tilde \mu_C(\d x)}{1 - \kappa + \kappa g \big( r_t^\mathsf{L}(\gamma_x) - \xi q_t(\gamma_x) \big)}.
\end{equation}
On $\R^+$, the map $g\mapsto I_t(g)$ is continuous and strictly decreasing from $(1-\kappa)^{-1}$ to $0$. Recall that $L$ is the uniform upper bound on the singular values of $\bC$ in \cref{Assumption1:Teacher}. A direct analysis gives, uniformly for $\gamma$ on $[\rho, \sqrt{\rho^2 + L^2}]$,
\begin{equation*}
    e^{-2t\gamma} q_t(\gamma) \xrightarrow[t \to \infty]{} q_\infty(\gamma), \hspace{1.5cm} e^{-2t\gamma} r_t^\mathsf{L}(\gamma) \xrightarrow[t \to \infty]{} r_\infty(\gamma),
\end{equation*}
where
\begin{equation*}
    q_\infty(\gamma)=\frac{\rho}{2\gamma}\left(1+\frac{\rho}{\gamma}\right), \hspace{1.5cm} r_\infty(\gamma)=\frac{\rho}{2}\left(1-\frac{\rho^2}{\gamma^2}\right).
\end{equation*}
Denote $T_C(a) = \beta \kappa^*\int_a^\infty\phi_C(x) \d x$. We apply \cref{Prop:SelfConsistentGeneral} with the quantities:
\begin{equation*}
    h_t = g_t(\xi), \hspace{1.5cm} a_t(x) = \kappa \big( r_t^\mathsf{L}(\gamma_x) - \xi q_t(\gamma_x) \big),
\end{equation*}
with parameter $t\to\infty$. Consider $a > 0$ and choose $c_t = e^{-2t\gamma_a}$. We have:
\begin{equation*}
    c_t a_t(x) \xrightarrow[t \to \infty]{}
    \begin{cases}
        0, & x < a, \\
        +\infty, & x > a.
    \end{cases}
\end{equation*}
Therefore, the limit function in \cref{Prop:SelfConsistentGeneral} is given by:
\begin{equation*}
    \Phi_a(u) = \frac{1 - T_C(a)}{1 - \kappa},
\end{equation*}
for any $u > 0$. Therefore, if $T_C(a) > \kappa$, then $\Phi_a(u) < 1$ for all $u > 0$, and \textbf{2.} $(ii)$ in \cref{Prop:SelfConsistentGeneral} gives that $g_t(\xi) e^{2t\gamma_a} \longrightarrow 0$. Likewise, if $T_C(a) < \kappa$, then \textbf{2.} $(iii)$ in \cref{Prop:SelfConsistentGeneral} guarantees that $g_t(\xi) e^{2t\gamma_a} \longrightarrow +\infty$. 

In the underparameterized case, the asymptotic behavior of $g_t(\xi)$ is therefore determined by the level set $\{x > 0 : T_C(x) = \kappa\}$. If this set contains an interval, $T_C$ is constant on this interval and $\phi_C$ vanishes almost everywhere on its interior. Therefore any choice of $x_*$ in this set leads to the same limiting integrals.

In the overparameterized case $\kappa > \beta \kappa^*$, we always have $T_C(a) < \kappa$. Then, we rather consider $a = 0$, so that $c_t = e^{-2 \rho t}$. In this case we have for $u > 0$,
\begin{equation*}
    \Phi_0(u) = \frac{1 - \beta \kappa^*}{1 - \kappa - \kappa \xi u}. 
\end{equation*}
Hence $\Phi_0(u) = 1$ has a unique solution on $(0, \infty)$. Applying \textbf{2.} $(i)$ in \cref{Prop:SelfConsistentGeneral} guarantees the convergence: 
\begin{equation} \label{eq:LimitgtOverparam}
    \kappa e^{2 \rho t} g_t(\xi) \xrightarrow[t \to \infty]{} -\frac{\kappa - \beta \kappa^*}{\xi}.
\end{equation}

Let us now derive the limit of the Stieltjes transforms. To do so, we consider:
\begin{equation*}
    F_t(x, \xi) = \frac{\kappa g_t(\xi) q_t(\gamma_x)}
    {1 - \kappa + \kappa g_t(\xi) \big( r_t^\mathsf{L}(\gamma_x) - \xi q_t(\gamma_x) \big)},
\end{equation*}
and compute its pointwise limit. Due to the previous asymptotic behavior of $g_t(\xi)$, we have for $x > 0$:
\begin{equation} \label{eq:AsymptoticFtxw}
    F_t(x, \xi) \xrightarrow[t \to \infty]{}
    \begin{cases}
        0, & T_C(x) > \kappa, \\
        \dfrac{q_\infty(\gamma_x)}{r_\infty(\gamma_x) - \xi q_\infty(\gamma_x)}, & T_C(x) < \kappa.
    \end{cases}
\end{equation}
Let us now treat the atom at $x = 0$. In the underparameterized case, $T_C(0) > \kappa$, so $F_t(0, \xi) \to 0$. In the overparameterized case, using equation \eqref{eq:LimitgtOverparam} leads to: 
\begin{equation*}
    F_t(0, \xi) \xrightarrow[t \to \infty]{} -\frac{\kappa - \beta \kappa^*}{1 - \beta \kappa^*} \frac{1}{\xi}.
\end{equation*}
We now use dominated convergence to characterize the $t \to \infty$ limit of $F_t(x, \xi)$ integrated against $\tilde \mu_C$. For $\xi < 0$, the denominator in the expression of $F_t$ is at least $-\xi \kappa g_t(\xi) q_t(\gamma_x)$, so $0 \leq F_t(x, \xi) \leq(-\xi)^{-1}$ uniformly in $t$ and $x$. Dominated convergence therefore gives
\begin{equation*}
\begin{aligned}
    \int F_t(x, \xi) \tilde \mu_C(\d x) \xrightarrow[t \to \infty]{} -\frac{(\kappa - \beta \kappa^*)^+}{\xi} + \beta \kappa^* \int_0^\infty \1_{T_C(x)<\kappa} \frac{q_\infty(\gamma_x)}{r_\infty(\gamma_x) - \xi q_\infty(\gamma_x)} \phi_C(x) \d x.
\end{aligned}
\end{equation*}
Let us now analyze the integral domains in both cases:
\begin{itemize}
    \item In the overparameterized case, every $x > 0$ satisfies $T_C(x) < \kappa$, so the integral covers the whole support of $\phi_C$. 
    \item In the underparameterized case, we can replace $\{T_C(x) < \kappa \}$ by $\{x > x_*\}$ for any $x_*$ such that $T_C(x_*) = \kappa$, since $\phi_C$ vanishes almost everywhere between any two such choices.
\end{itemize}
Finally, we use the following identity, for all $\gamma \geq \rho$:
\begin{equation*}
    \frac{q_\infty(\gamma)}{r_\infty(\gamma) - \xi q_\infty(\gamma)} = \frac{1}{\gamma - \rho - \xi}.
\end{equation*}
We now consider $w < 0$. Using the expression of $\mathfrak m_t^U$ in \cref{Thm:Stieltjes} and choosing $\xi = w$ directly yields equation \eqref{eq:StieltjesUinf}. For $\mathfrak m_t^V$, we choose $\xi = w - 2 \rho$ and use that $r_t^\mathsf{R} = r_t^\mathsf{L} + 2 \rho q_t$, so that the expression of $\mathfrak m_t^V$ in equation \eqref{eq:StieltjesV} writes:
\begin{equation*}
    \mathfrak m_t^V(w) = - \frac{1 - \kappa}{w} + \int F_t(x, w - 2 \rho) \tilde \mu_C(\d x),
\end{equation*}
and we obtain equation \eqref{eq:StieltjesVinf} taking $t \to \infty$. Finally, using that both sides of equations \eqref{eq:StieltjesUinf}, \eqref{eq:StieltjesVinf} are holomorphic on $\mathbb{C} \setminus \R^+$, \cref{Cor:HolomorphicExtension} and \cref{Lemma:StieltjesComplex} guarantee the result for all $w \in \mathbb{C} \setminus \R^+$. 

\paragraph{Step 2. Singular value distribution.} Let us now derive equation \eqref{eq:StieltjesZinf}. To do so, we use equation \eqref{eq:StieltjesUtoZ} that expresses $\mu_t^Z$ in terms of $\mu_t^U$. Equation \eqref{eq:StieltjesUinf} identifies the limiting measure of $\bU(t)\bU(t)^\top$ as:
\begin{equation*}
    \mu_\infty^U(\d x)
    =
    \left(1-\frac{\min(\kappa,\beta\kappa^*)}{\tau}\right)\delta_0(\d x)
    +\frac{\beta\kappa^*}{\tau} (\theta_\rho^{-1})_\# \big(\1_{x > x_*} \phi_C(x) \d x \big),
\end{equation*}
where $\theta_\rho(x) = \sqrt{x (x + 2\rho)}$ and $\theta_\rho^{-1}(x) = \sqrt{\rho^2 + x^2} - \rho$ on $\R^+$. Since $\theta_\rho$ is continuous, the pushforward relation \eqref{eq:StieltjesUtoZ} passes to the weak limit and leads to
\begin{equation*}
    \mu_\infty^Z(\d x) = \left( 1 - \frac{\min(\kappa, \beta \kappa^*)}{\beta} \right) \delta_0(\d x) + \kappa^* \1_{x > x_*} \phi_C(x) \d x.
\end{equation*}
Equation \eqref{eq:StieltjesZinf} follows by taking the Stieltjes transform of this measure. 

\paragraph{Step 3. Comparison with finite dimension.} We shall now relate \cref{Prop:LimitBlockMatrices} and \cref{Cor:StieltjesInf}. To do so, let $\sigma_1 \geq \dots \geq \sigma_{\min(p,d)}$ be the singular values of $\bC$, and set $r_d = \min(m,m^*)$. The finite-dimensional limit selects the first $r_d$ singular values. Under \cref{Assumption1:Teacher}, we have $m^* / d \to \beta \kappa^*$. Combining weak convergence and the absence of an atom at $x_*$, we can show that the selection of the $\min(m, m^*)$ top singular values of $\bC$ is achieved by the threshold $x_*$ in the high-dimensional limit. Then, if $f$ is bounded continuous
\begin{equation} \label{eq:LimitSelectionSingularvals}
    \lim_{d \to \infty} \frac{1}{d} \sum_{i=1}^{r_d} f(\sigma_i) = \beta \kappa^* \int_{x_*}^\infty f(x) \phi_C(x) \d x. 
\end{equation}
Starting from \cref{Prop:LimitBlockMatrices}, the Stieltjes transforms of the limiting block matrices are
\begin{equation*}
\begin{aligned}
    \mathfrak m^U(w) &= \frac{1}{p} \sum_{i=1}^{\min(m, m^*)} \frac{1}{\gamma_i - \rho - w} - \frac{1}{w} \frac{p - \min(m, m^*)}{p}, \\
    \mathfrak m^V(w) &= \frac{1}{d} \sum_{i=1}^{\min(m, m^*)} \frac{1}{\gamma_i + \rho - w} + \frac{\mathrm{rank} \, \overline \bPi}{d} \frac{1}{2 \rho - w} - \frac{1}{w} \frac{d - \min(m, m^*) -  \mathrm{rank} \, \overline \bPi}{d}, \\
    \mathfrak m^Z(w) &= \frac{1}{\min(p,d)} \sum_{i=1}^{\min(m, m^*)} \frac{1}{\sigma_i - w} - \frac{1}{w} \frac{\min(p,d) - \min(m, m^*)}{\min(p,d)},
\end{aligned}
\end{equation*}
where $\gamma_i = \sqrt{\sigma_i^2 + \rho^2}$. Using the asymptotic behavior of $m, m^*, p$ in the high-dimensional limit, along with $\mathrm{rank} \, \overline \bPi = (m-m^*)^+$ almost surely and equation \eqref{eq:LimitSelectionSingularvals}, we exactly recover the expressions of \cref{Cor:StieltjesInf}. 

\subsection{Two-Point Stieltjes Transforms} \label{App:Subsec:DoubleStieltjes}

As mentioned in \cref{Subsec:StieltjesTransforms}, the Stieltjes transform characterizes the spectrum of the associated matrix but does not give information on its eigenvectors. In order to characterize the eigenvectors of a random matrix in high dimension, one possibility is to introduce a reference matrix $\bA_0 \in \mathcal{S}_d(\R)$ and compare the eigenvectors (or the projections onto the eigenspaces) of $\bA$ and $\bA_0$. This can be achieved by considering:
\begin{equation*}
    H^{A, A_0}(w, z) = \frac{1}{d} \tr \Big( \big( \bA - w \bI_d \big)^{-1} \big( \bA_0 - z \bI_d \big)^{-1} \Big). 
\end{equation*}
This object was studied by \citet{ledoit2011eigenvectors} for sample covariance matrices, and by \citet{bun2018overlaps} for correlated random matrices, to analyze overlaps between eigenvectors \citep[see also][]{allez2014eigenvector}.

In this section, we study the two-point Stieltjes transforms:
\begin{equation*}
\begin{aligned}
    H_t^U(w, z) &= \frac{1}{p} \tr \Big( \big( \bU(t) \bU(t)^\top - w \bI_p \big)^{-1} \big( \bC \bC^\top - z \bI_p \big)^{-1} \Big), \\
    H_t^V(w, z) &= \frac{1}{d} \tr \Big( \big( \bV(t) \bV(t)^\top - w \bI_d \big)^{-1} \big( \bC^\top \bC - z \bI_d \big)^{-1} \Big),
\end{aligned}    
\end{equation*} 
and provide a complementary high-dimensional limit for these functions. As shown in \cref{Prop:LimitBlockMatrices}, in finite dimension, $\bU(t) \bU(t)^\top$ and $\bV(t) \bV(t)^\top$ respectively converge to spectral functions of $\bC \bC^\top$ and $\bC^\top \bC$ (up to the projection $\overline \bPi$ for $\bV$). Therefore the choice of these two matrices as reference matrices for the two-point Stieltjes transform is relevant.  

\begin{claim} \label{Thm:DoubleStieltjes}
    Fix $t \geq 0$. In the high-dimensional limit, under \cref{Assumption1}, given $w, z \in \mathbb{C} \setminus \R^+$, we have the almost sure convergence:
    \begin{equation*}
    \begin{aligned}
        H_t^U(w, z) &\xrightarrow[d\to\infty]{} \frac{1}{wz} \left(1 - \frac{1}{\tau} \right) - \frac{1}{\tau w} \int \frac{\tilde \mu_C(\d x)}{x^2 - z} \frac{1 - \kappa - \kappa w g_t(w) q_t (\gamma_x)}{1 - \kappa + \kappa g_t(w) \big( r_t^\mathsf{L}(\gamma_x) - w q_t(\gamma_x) \big)}, \\
        H_t^V(w, z) &\xrightarrow[d\to\infty]{} - \frac{1}{w} \int \frac{\tilde \mu_C(\d x)}{x^2 - z} \frac{1 - \kappa - \kappa w g_t(w - 2 \rho) q_t (\gamma_x)}{1 - \kappa + \kappa g_t(w - 2 \rho) \big( r_t^\mathsf{R}(\gamma_x) - w q_t(\gamma_x) \big)},
    \end{aligned}
    \end{equation*}
    where the functions $q_t,r_t^\mathsf{L},r_t^\mathsf{R},g_t$ and the measure $\tilde \mu_C$ are defined in \cref{Thm:Stieltjes}. 
\end{claim}

The proof is very similar to \cref{App:Subsubsec:StieltjesProof}, and mainly consists in applying \cref{Thm:RMTConvergence} with the additional factor $(x^2-z)^{-1}$. As we do not use this result in the rest of the paper, we omit the remaining algebra. 

\subsection{Range-Projector Distances} \label{App:Subsec:Projectors1}

In the following we prove \cref{Thm:Projector1}. Recall that we are interested in the distances:
\begin{equation*}
    \Delta^\mathsf{L}(t)^2 = \frac{1}{d} \big\| \bP^\mathsf{L}(t) - \bP^\mathsf{L}_\infty \big\|_F^2, \hspace{1.5cm} \Delta^\mathsf{R}(t)^2 = \frac{1}{d} \big\| \bP^\mathsf{R}(t) - \bP^\mathsf{R}_\infty \big\|_F^2,
\end{equation*}
where
\begin{equation*}
    \bP^\mathsf{L}(t) = \mathrm{Proj}_{\mathrm{Im} \, \bZ(t)} \in \mathcal{S}_p(\R), \hspace{1.5cm} \bP^\mathsf{R}(t) = \mathrm{Proj}_{\mathrm{Im} \, \bZ(t)^\top} \in \mathcal{S}_d(\R),
\end{equation*}
are the projections onto the left and right ranges of $\bZ(t)$, and $\bP_\infty^\mathsf{L} = \lim_{t \to \infty} \bP^\mathsf{L}(t)$ (and likewise for $\bP_\infty^\mathsf{R}$). 

\paragraph{Step 1. Reduction to the Gram matrices.} We start by relating the projections $\bP^\mathsf{L}, \bP^\mathsf{R}$ to the projections onto the ranges of $\bU(t), \bV(t)$. Let us denote:
\begin{equation*}
    \bP^U(t) = \mathrm{Proj}_{\mathrm{Im} \, \bU(t)} \in \mathcal{S}_p(\R), \hspace{1.5cm} \bP^V(t) = \mathrm{Proj}_{\mathrm{Im} \, \bV(t)} \in \mathcal{S}_d(\R).
\end{equation*} 

\begin{lemma} \label{Lemma:IdentificationProjs}
    Under the assumptions of \cref{Thm:Projector1}, with probability one, we have for all $t > 0$:
    \begin{equation} \label{eq:IdentificationProjs}
        \Delta^\mathsf{L}(t)^2 = \frac{1}{d} \big\| \bP^U(t) - \bP_\infty^U \big\|_F^2, \hspace{1.5cm} \Delta^\mathsf{R}(t)^2 = \frac{1}{d} \big\| \bP^V(t) - \bP_\infty^V \big\|_F^2. 
    \end{equation}
\end{lemma}

\begin{proof}
    We apply the result of \cref{Prop:SVDPrediction} and use the associated notation. For $t > 0$, with $r_Z = \min(m, m^*)$:
    \begin{equation*}
        \bZ(t) = \sum_{i=1}^{r_Z} \sqrt{\lambda_i^U(t)\big(\lambda_i^U(t) + 2 \rho \big)}\, \widehat{\bu}_i(t) \widehat{\bv}_i(t)^\top.
    \end{equation*}
    Therefore:
    \begin{equation*}
        \bP^\mathsf{L}(t) = \sum_{i=1}^{r_Z} \widehat{\bu}_i(t) \widehat{\bu}_i(t)^\top, \hspace{1.5cm} \bP^\mathsf{R}(t) = \sum_{i=1}^{r_Z} \widehat{\bv}_i(t) \widehat{\bv}_i(t)^\top.
    \end{equation*}
    Since $\widehat{\bu}_1(t), \dots, \widehat{\bu}_{r_Z}(t)$ are eigenvectors of $\bU(t) \bU(t)^\top$ associated with its non-zero eigenvalues, we have $\bP^\mathsf{L}(t) = \bP^U(t)$. Since $\bV(t)$ is of rank $m$ and $r_Z = \min(m, m^*)$, we also get $\bP^\mathsf{R}(t) = \bP^V(t)$ for $m \leq m^*$. When $m > m^*$, the extra eigenvalues of $\bV(t) \bV(t)^\top$ are $2 \rho$, and we have shown in \cref{Lemma:TrivialSubspace} that the associated eigenspace is the range of the projection $\overline \bPi$ (defined in equation \eqref{eq:BlockProjection}) onto $\mathrm{Im}(\bR) \cap \mathrm{Ker}(\bC)$. This space has dimension $m-m^*$ with probability one and is independent of time. In this case $\bP^V(t) = \bP^\mathsf{R}(t) + \overline \bPi$, and $\overline \bPi$ cancels out in the expression of $\bP^V(t) - \bP^V_\infty$. This concludes the proof.    
\end{proof}

\paragraph{Step 2. Reduction to spectral overlaps.} From now on, we work with the projectors $\bP^U(t), \bP^V(t)$, and characterize their long-time limits.

\begin{lemma} \label{Lemma:ProjectorLongtimes}
    Consider the notation and assumptions of \cref{Prop:LimitBlockMatrices}. With probability one
    \begin{equation*}
        \bP_\infty^U = \sum_{i=1}^{r_Z} \bu_i \bu_i^\top, \hspace{1.5cm} \bP_\infty^V = \sum_{i=1}^{r_Z} \bv_i \bv_i^\top + \1_{m > m^*} \overline \bPi,
    \end{equation*}
    where $\overline \bPi$ is the orthogonal projection onto $\mathrm{Im}(\bR) \cap \mathrm{Ker}(\bC)$.
\end{lemma}

\begin{proof}
    The result follows directly from equations \eqref{eq:LimitBlockMatricesU} and \eqref{eq:LimitBlockMatricesV}. In the overparameterized case, $\mathrm{Im}(\overline\bPi)\subset\mathrm{Ker}(\bC)$ is orthogonal to $\mathrm{Span}(\bv_1,\ldots,\bv_{m^*})$, so the two projectors can be added.
\end{proof}

We now consider the finite-dimensional projectors
\begin{equation} \label{eq:SignalProjs}
    \bQ^\mathsf{L} = \sum_{i=1}^{r_Z} \bu_i \bu_i^\top, \hspace{1.5cm} \bQ^\mathsf{R} = \sum_{i=1}^{r_Z} \bv_i \bv_i^\top.
\end{equation}
By \cref{Prop:SVDPrediction}, $\mathrm{rank} \, \bU(t) = r_Z$ and $\mathrm{rank} \, \bV(t) = m$ with probability one. Moreover, $\bP^V(t) \overline \bPi = \overline \bPi$ at all times and $\mathrm{rank} \, \overline \bPi = (m-m^*)^+$. Therefore, expanding the Frobenius norm in equation \eqref{eq:IdentificationProjs} and using \cref{Lemma:ProjectorLongtimes}
\begin{equation} \label{eq:RangeDistanceOverlapFinite}
    \Delta^\mathsf{L}(t)^2 = \frac{2r_Z}{d} - \frac{2}{d} \mathrm{Tr} \big(\bP^U(t) \bQ^\mathsf{L} \big), \hspace{1.2cm} 
    \Delta^\mathsf{R}(t)^2 = \frac{2r_Z}{d} - \frac{2}{d} \mathrm{Tr} \big(\bP^V(t)\bQ^\mathsf{R} \big).
\end{equation}
The rest of the proof computes the limit of the projector overlaps. 

\paragraph{Step 3. Projection of the weight matrices.} We now compute a more general form of projector overlaps.

\begin{lemma} \label{Lemma:ExpressionProjections}
    Consider the notation and assumptions of \cref{Lemma:BlockDecomposition1}. Let $\psi \colon \R \to \R$ be continuous. Then, for $\bV(t)$, and for $\bU(t)$ whenever $m \leq m^*$ and $t > 0$,
    \begin{equation*}
    \begin{aligned}
        \frac{1}{d} \tr\left(\bP^U(t)\psi(\bC\bC^\top)\right) &= \frac{1}{d} \tr\left(
        \bR^\top \chi_t^U(\bS)^2 \psi(\bS^2) \bR
        \left( \bR^\top \chi_t^U(\bS)^2 \bR \right)^{-1} \right), \\
        \frac{1}{d} \tr\left(\bP^V(t) \psi(\bC^\top \bC) \right) &= \frac{1}{d} \tr \left( \bR^\top \chi_t^V(\bS)^2 \psi(\bS^2) \bR
        \big(\bR^\top \chi_t^V(\bS)^2 \bR \big)^{-1}
        \right),
    \end{aligned}
    \end{equation*}
    where $\chi_t^U, \chi_t^V$ are applied spectrally to $\bS$.
\end{lemma}

\begin{proof}
    From \cref{Lemma:BlockDecomposition1}, the range of $\bV(t)$ is
    \begin{equation*}
        \mathrm{Im} \, \bV(t)=\mathrm{Im} \left(\chi_t^V(\bS)\bR \right).
    \end{equation*}
    Similarly, when $m\leq m^*$, we have $\mathrm{Ker}(\chi_t^U(\bC))=\mathrm{Ker}(\bC)$. By \cref{Lemma:GenericPositionStiefel}, $\mathrm{Im}(\bR)\cap\mathrm{Ker}(\bC)=\{\boldsymbol 0\}$ with probability one, so $\chi_t^U(\bC)\bR$ has full column rank and
    \begin{equation*}
        \mathrm{Im} \, \bU(t) = \mathrm{Im} \left(\chi_t^U(\bC) \bR \right).
    \end{equation*}
    Moreover, $\chi_t^V(\bS)$ is invertible, so $\chi_t^V(\bS)\bR$ also has full column rank. For any full-column-rank matrix $\bX$, the orthogonal projector onto its range is $\bX(\bX^\top\bX)^{-1}\bX^\top$. Applying this identity gives
    \begin{equation*}
    \begin{aligned}
        \bP^U(t) &= \chi_t^U(\bC) \bR \Big( \bR^\top \chi_t^U(\bC)^\top \chi_t^U(\bC) \bR \Big)^{-1} \bR^\top \chi_t^U(\bC)^\top, \\
        \bP^V(t) &= \chi_t^V(\bS) \bR \Big( \bR^\top \chi_t^V(\bS)^2 \bR \Big)^{-1} \bR^\top \chi_t^V(\bS).
    \end{aligned}
    \end{equation*}
    This directly implies the identity for $V$. For $U$, applying \cref{Lemma:RectangulartoSquare} leads to:
    \begin{equation*}
    \begin{aligned}
        \chi_t^U(\bC)^\top \chi_t^U(\bC) &= \chi_t^U(\bS)^2, \\
        \chi_t^U(\bC)^\top \psi(\bC \bC^\top) \chi_t^U(\bC) &=  \chi_t^U(\bS)^2 \psi(\bS^2),
    \end{aligned}
    \end{equation*} 
    which concludes the proof of the lemma.
\end{proof}

\paragraph{Step 4. High-dimensional limit.} We apply \cref{Thm:RMTConvergenceReal} with $\bX=\bR$ and $\bD=\bS$. Its empirical spectral measure converges weakly to $\tilde \mu_C$. Moreover, $\chi_t^V(x) \geq 1$, while $\chi_t^U(x)>0$ exactly when $x>0$. Hence, for $\bU$ in the underparameterized case,
\begin{equation*}
    \tilde \mu_C \big( \big\{ x \colon \chi_t^U(x) > 0 \big\} \big) = \tilde \mu_C \big( (0, \infty) \big) = \beta \kappa^* > \kappa. 
\end{equation*}
For $U$ and $V$, respectively, take $a(x) = \chi_t^U(x)^2 \geq 0$ and $a(x) = \chi_t^V(x)^2 \geq 1$, with $b(x) = a(x)\psi(x^2)$. For $U$, \cref{Assumption1:Teacher} guarantees that $a(\bS)$ has its non-zero eigenvalues uniformly bounded away from zero. Thus, \cref{Thm:RMTConvergenceReal} gives the almost sure convergence
\begin{equation*}
\begin{aligned}
    \frac{1}{d} \tr \Big( \bP^U(t) \psi(\bC \bC^\top) \Big) &\xrightarrow[d \to \infty]{} \lambda_t^\mathsf{L} \int \psi(x^2) \frac{\chi_t^U(x)^2}{1 - \kappa + \lambda_t^\mathsf{L} \chi_t^U(x)^2} \tilde \mu_C(\d x), \\
    \frac{1}{d} \tr \Big( \bP^V(t) \psi(\bC^\top \bC) \Big) &\xrightarrow[d \to \infty]{} \lambda_t^\mathsf{R} \int \psi(x^2) \frac{\chi_t^V(x)^2}{1 - \kappa + \lambda_t^\mathsf{R} \chi_t^V(x)^2} \tilde \mu_C(\d x),
\end{aligned}
\end{equation*}
where $\lambda_t^\mathsf{L},\lambda_t^\mathsf{R}$ solve
\begin{equation*}
    1 = \int \frac{\tilde \mu_C(\d x)}{1 - \kappa + \lambda_t^\mathsf{L} \chi_t^U(x)^2}, \hspace{1.5cm} 1 = \int \frac{\tilde \mu_C(\d x)}{1 - \kappa + \lambda_t^\mathsf{R} \chi_t^V(x)^2}. 
\end{equation*}
Using the notation of \cref{Thm:Stieltjes}, we have
\begin{equation*}
    2 \rho \chi_t^U(x)^2 = r_t^\mathsf{L}(\gamma_x), \hspace{1.5cm} 2 \rho \chi_t^V(x)^2 = r_t^\mathsf{R}(\gamma_x).
\end{equation*}
Writing $\lambda_t^\mathsf{a} = 2 \rho \kappa g_t^\mathsf{a}$, the two limits become
\begin{align}
    \frac{1}{d} \tr \Big( \bP^U(t) \psi(\bC \bC^\top) \Big) &\xrightarrow[d \to \infty]{} \kappa g_t^\mathsf{L} \int \psi(x^2) \frac{r_t^\mathsf{L}(\gamma_x)}{1 - \kappa + \kappa g_t^\mathsf{L} r_t^\mathsf{L}(\gamma_x)} \tilde \mu_C(\d x), \label{eq:LimitSpecProjU}\\
    \frac{1}{d} \tr \Big( \bP^V(t) \psi(\bC^\top \bC) \Big) &\xrightarrow[d \to \infty]{} \kappa g_t^\mathsf{R} \int \psi(x^2) \frac{r_t^\mathsf{R}(\gamma_x)}{1 - \kappa + \kappa g_t^\mathsf{R} r_t^\mathsf{R}(\gamma_x)} \tilde \mu_C(\d x). \label{eq:LimitSpecProjV}
\end{align}
This holds for $\kappa \in (0, 1)$ in the case of $V$, and for $\kappa \in (0, \beta \kappa^*)$ for $U$. 

\paragraph{Step 5. Overparameterized case for $U$.} Suppose $\kappa > \beta \kappa^*$. Under \cref{Assumption1:Teacher}, $m>m^*$ for all large $d$. We return to the range identity
\begin{equation*}
    \mathrm{Im} \, \bU(t) = \mathrm{Im} \left(\chi_t^U(\bC) \bR \right).
\end{equation*}
The range of $\chi_t^U(\bC)\bR$ is contained in $\mathrm{Im}(\bC)$. Moreover, since $\mathrm{Ker}(\chi_t^U(\bC))=\mathrm{Ker}(\bC)$ and $\bR$ is injective, we have:
\begin{equation*}
    \mathrm{dim} \, \mathrm{Ker} \big(\chi_t^U(\bC)\bR \big) = \dim \big(\mathrm{Im}(\bR) \cap \mathrm{Ker}(\bC)\big).
\end{equation*}
By \cref{Lemma:GenericPositionStiefel} this space has dimension $m-m^*$ almost surely, and therefore $\bU(t)$ has rank $m^*$. Since its range is included in the one of $\bC$, we have $\mathrm{Im} \, \bU(t)=\mathrm{Im}\, \bC$ almost surely for all $t > 0$. Therefore $\bP^U(t) = \bP^U_\infty$, and the left range-projector distance is identically zero for every $t > 0$:
\begin{equation*}
    \Delta^\mathsf{L}(t)^2 = 0.
\end{equation*}

\paragraph{Step 6. Approximation of the indicator.} It remains to treat the underparameterized case for $U$ and both regimes for $V$. Let
\begin{equation*}
    \psi_*(y)=\1\{\sqrt y>x_*\},
\end{equation*}
and define the projectors
\begin{equation*}
    \bQ_*^\mathsf{L} = \psi_* \big(\bC\bC^\top \big), \hspace{1.5cm} \bQ_*^\mathsf{R} = \psi_* \big(\bC^\top\bC \big).
\end{equation*}
Since the nonzero eigenvalues of $\bC \bC^\top$ and $\bC^\top \bC$ are the squared nonzero singular values of $\bC$, these finite-dimensional projectors select the directions associated with the singular values of $\bC$ above $x_*$, which is the threshold obtained in the high-dimensional limit. On the other hand, the projectors $\bQ^\mathsf{L}, \bQ^\mathsf{R}$ defined in equation \eqref{eq:SignalProjs} always select the top $r_d$ directions. For $\mathsf{a} = \mathsf{L}, \mathsf{R}$, the two projectors are diagonal in the same singular basis, and $\bQ^\mathsf{a} - \bQ_*^\mathsf{a}$ is supported only on the directions selected by one cutoff but not the other. By weak convergence and the absence of an atom at $x_*$, both ranks divided by $d$ converge to $\min(\kappa,\beta\kappa^*)$. Therefore:
\begin{equation*}
    \frac{1}{d} \mathrm{rank} \big(\bQ^\mathsf{a} - \bQ_*^\mathsf{a} \big)\xrightarrow[d \to \infty]{} 0.
\end{equation*}
As a consequence:
\begin{equation} \label{eq:ConvergenceRank}
    \frac{1}{d} \left| \tr \Big( \bP^U(t)  \big(\bQ^\mathsf{L} - \bQ_*^\mathsf{L} \big) \Big) \right| \xrightarrow[d \to \infty]{} 0, \hspace{1.5cm} \frac{1}{d} \left| \tr \Big( \bP^V(t)  \big(\bQ^\mathsf{R} - \bQ_*^\mathsf{R} \big) \Big) \right| \xrightarrow[d \to \infty]{} 0.
\end{equation}
We now approximate $\psi_*$ by a continuous function. Fix $0<\epsilon<x_*/2$ and set
\begin{equation*}
    W_\epsilon=\big\{ x \geq 0 : |x-x_*| \leq 2 \epsilon \big\}.
\end{equation*}
Choose a continuous $\psi_\epsilon$ such that $0 \leq \psi_* \leq \psi_\epsilon \leq 1$ and such that $\psi_\epsilon(y) = \psi_*(y)$ whenever $\sqrt y \notin W_\epsilon$. We would like to show that equations \eqref{eq:LimitSpecProjU}, \eqref{eq:LimitSpecProjV} extend to $\psi_*$. We then study the LHS and RHS separately.

For the LHS, we have
\begin{equation} \label{eq:BoundProj1}
    \frac{1}{d} \left|\mathrm{Tr} \Big(\bP^U(t)(\psi_\epsilon - \psi_*)
    (\bC\bC^\top)\Big) \right| \leq \frac{1}{d} \mathrm{Tr} \Big((\psi_\epsilon - \psi_*)(\bC\bC^\top)\Big) \leq \mu_d^S(W_\epsilon),
\end{equation}
where $\mu_d^S$ is the empirical spectral measure of $\bS=(\bC^\top \bC)^{1/2}$. The same holds for $V$. Moreover, for the RHS, with $\mathsf{a} = \mathsf{L}, \mathsf{R}$:
\begin{equation*}
    0 \leq \frac{\kappa g_t^\mathsf{a} r_t^\mathsf{a}(\gamma_x)}
    {1 - \kappa + \kappa g_t^\mathsf{a} r_t^\mathsf{a} (\gamma_x)} \leq 1,
\end{equation*}
so that
\begin{equation} \label{eq:BoundProj2}
    \left| \kappa g_t^\mathsf{a} \int \big( \psi_\epsilon(x^2) - \psi_*(x^2) \big) \frac{r_t^\mathsf{a}(\gamma_x)}{1 - \kappa + \kappa g_t^\mathsf{a} r_t^\mathsf{a}(\gamma_x)} \tilde \mu_C(\d x) \right| \leq \tilde \mu_C(W_\epsilon).
\end{equation}
Combining these estimates with the replacement of $\bQ^\mathsf{L}$ by $\bQ_*^\mathsf{L}$ gives
\begin{align*}
&\left| \frac{1}{d} \tr \big(\bP^U(t)\bQ^\mathsf{L} \big) - \kappa g_t^\mathsf{L} \int_{x_*}^\infty \frac{r_t^\mathsf{L}(\gamma_x)}{1 - \kappa+ \kappa g_t^\mathsf{L} r_t^\mathsf{L}(\gamma_x)}\,\tilde \mu_C(\d x) \right| \\
&\hspace{1.5cm} \leq \frac{1}{d}\left| \tr \big(\bP^U(t)(\bQ^\mathsf{L} - \bQ_*^\mathsf{L})\big) \right| +\frac{1}{d}\left| \tr \Big(\bP^U(t) \big(\bQ_*^\mathsf{L} - \psi_\epsilon(\bC\bC^\top)\big) \Big) \right| \\
&\hspace{1.5cm} + \left|\frac{1}{d} \tr\big(\bP^U(t)\psi_\epsilon(\bC\bC^\top) \big) - \kappa g_t^\mathsf{L} \int \psi_\epsilon(x^2)
\frac{r_t^\mathsf{L}(\gamma_x)}{1-\kappa+\kappa g_t^\mathsf{L}r_t^\mathsf{L}(\gamma_x)}\,\tilde\mu_C(\d x)\right| \\
&\hspace{1.5cm}+\left|\kappa g_t^\mathsf{L}\int\big(\psi_\epsilon(x^2)-\psi_*(x^2)\big)
\frac{r_t^\mathsf{L}(\gamma_x)}{1-\kappa+\kappa g_t^\mathsf{L}r_t^\mathsf{L}(\gamma_x)}\,\tilde\mu_C(\d x)\right|.
\end{align*}
Equation \eqref{eq:ConvergenceRank} guarantees that the first term goes to zero as $d \to \infty$. The second term is bounded by $\mu_d^S(W_\epsilon)$ thanks to equation \eqref{eq:BoundProj1}, which converges to $\tilde \mu_C(W_\epsilon)$. Additionally, since $\psi_\epsilon$ is continuous, equation \eqref{eq:LimitSpecProjU} shows that the third term converges to zero as $d \to \infty$. In the end, using equation \eqref{eq:BoundProj2} to bound the fourth term, we get that:
\begin{equation*}
    \limsup_{d \to \infty} \left| \frac{1}{d} \tr \big(\bP^U(t)\bQ^\mathsf{L} \big) - \kappa g_t^\mathsf{L} \int_{x_*}^\infty
\frac{r_t^\mathsf{L}(\gamma_x)}{1 - \kappa+ \kappa g_t^\mathsf{L} r_t^\mathsf{L}(\gamma_x)}\,\tilde \mu_C(\d x) \right| \leq 2 \tilde \mu_C(W_\epsilon). 
\end{equation*}  
Since $x_*$ is not an atom of $\tilde\mu_C$, the right-hand side vanishes as $\epsilon\downarrow0$. The same argument, using equation \eqref{eq:LimitSpecProjV}, applies to $V$. Using equation \eqref{eq:RangeDistanceOverlapFinite}, we exactly obtain the formula in \cref{Thm:Projector1}.

\section{Small-Initialization Limit} \label{App:SmallInit}
In this section we prove the results of \cref{Sec:SilentAlignment}.

\subsection{Dynamics at Fixed Times} \label{App:Subsec:FiniteDimIntuition}

\subsubsection{Expressions in Finite Dimension} \label{App:Subsubsec:FiniteDim}

We shall now prove \cref{Prop:FiniteDimensionIntuition}. We start from the explicit expression of $\bU(t) \bU(t)^\top$ and $\bV(t) \bV(t)^\top$ at finite $\rho$ obtained in \cref{Lemma:BlockDecomposition1}. These identities read:
\begin{equation*}
\begin{aligned}
    \bU(t) \bU(t)^\top &= 2 \rho \chi_t^U(\bC) \bR \Big( \bR^\top q_t(\bGamma_\rho) \bR \Big)^{-1} \bR^\top \chi_t^U(\bC)^\top, \\
    \bV(t) \bV(t)^\top &= 2 \rho \chi_t^V(\bS) \bR \Big( \bR^\top q_t(\bGamma_\rho) \bR \Big)^{-1} \bR^\top \chi_t^V(\bS),\\
    \bZ(t) &= 2 \rho \chi_t^U(\bC) \bR \Big( \bR^\top q_t(\bGamma_\rho) \bR \Big)^{-1} \bR^\top \chi_t^V(\bS),
\end{aligned}
\end{equation*}
where $\bS=(\bC^\top\bC)^{1/2}$ and $\bGamma_\rho = \big( \bS^2 + \rho^2 \bI_d \big)^{1/2}$. Taking the limit $\rho \to 0$, we have for all $\sigma \geq 0$, $x \geq 0$:
\begin{equation*}
    \chi_t^U(\sigma) \xrightarrow[\rho \to 0]{} \sinh(t\sigma), \hspace{1.2cm} \chi_t^V(\sigma) \xrightarrow[\rho \to 0]{} \cosh(t\sigma), \hspace{1.2cm} q_t(\gamma_x) \xrightarrow[\rho \to 0]{} 1.
\end{equation*}
Taking these limits in the three equations, together with $\bGamma_\rho \xrightarrow[\rho \to 0]{} (\bC^\top \bC)^{1/2}$, proves the result.

\subsubsection{High-Dimensional Limit}

We now start from the result of \cref{Prop:FiniteDimensionIntuition} and prove the following result.

\begin{proposition} \label{Prop:StieltjesSmallInit}
    Consider the expressions of $\rho^{-1} \bU(t) \bU(t)^\top$, $\rho^{-1} \bV(t) \bV(t)^\top$ in the $\rho \to 0$ limit given in \cref{Prop:FiniteDimensionIntuition}. In the high-dimensional limit, under \cref{Assumption1}, their Stieltjes transforms converge almost surely to:
    \begin{equation*}
    \begin{aligned}
        \varsigma_t^U(w) &= - \frac{1}{w} \left(1 - \frac{\kappa}{\tau} \right) + \frac{\kappa}{\tau} h_t(w), \\
        \varsigma_t^V(w) &= - \frac{1}{w} \big( 1 - \kappa \big) + \kappa h_t(w-2),
    \end{aligned}
    \end{equation*}
    for $w \in \mathbb{C}\setminus \R^+$, where $h_t$ is the unique holomorphic solution such that $h_t(w)>0$ for $w<0$ of
    \begin{equation*}
        1 = \int \frac{\tilde \mu_C(\d x)}{1 - \kappa + \kappa h_t(w) \big( 2\bar r_t^\mathsf{L}(x) - w \big)},
    \end{equation*}
    with
    \begin{equation}
        \bar r_t^\mathsf{L}(x) = \sinh(tx)^2.
    \end{equation}
\end{proposition}

\begin{proof}
    To start the proof, note that the $\rho \to 0$ limit of $\rho^{-1} \bU(t) \bU(t)^\top$ has the same non-zero eigenvalues as the matrix:
    \begin{equation*}
        2 \bR^\top \sinh(t\bC)^\top \sinh(t\bC) \bR \in \mathcal{S}_m^+(\R). 
    \end{equation*}
    Using \cref{Lemma:RectangulartoSquare}, we get:
    \begin{equation*}
        \sinh(t\bC)^\top \sinh(t\bC) = \sinh \Big( t \big( \bC^\top \bC \big)^{1/2} \Big)^2.
    \end{equation*}
    Likewise, the non-zero eigenvalues of $\rho^{-1} \bV(t) \bV(t)^\top$ (in the $\rho \to 0$ limit) are the same as those of:
    \begin{equation*}
        2 \bR^\top \cosh \Big( t \big( \bC^\top \bC \big)^{1/2} \Big)^2 \bR. 
    \end{equation*}
    This reformulation allows us to treat the cases of $\bU$ and $\bV$ jointly. As $d \to \infty$, the empirical spectral distribution of $\big( \bC^\top \bC \big)^{1/2}$ converges to $\tilde \mu_C$ (defined in \cref{Thm:Stieltjes}). Recall that $\bar r_t^\mathsf{R}(x)=\cosh(tx)^2$, and set
    \begin{equation*}
        \bPhi_t^U = 2\bar r_t^\mathsf{L} \Big( \big( \bC^\top \bC \big)^{1/2} \Big), \hspace{1.5cm} \bPhi_t^V = 2\bar r_t^\mathsf{R} \Big( \big( \bC^\top \bC \big)^{1/2} \Big).
    \end{equation*}
    For $w \in \mathbb{C}\setminus \R^+$, we apply \cref{Thm:RMTConvergence} with $\zeta = 1/w$, $q=1$, and $r=2\bar r_t^\mathsf{a}$, where $\mathsf a=\mathsf L,\mathsf R$ corresponds respectively to $\bullet=U,V$. If $\mathfrak h_t^\bullet(\zeta)$ denotes the unique holomorphic solution of equation \eqref{eq:RMTSelfConsistentStieltjes} for these choices, positive for $\zeta<0$, then the limiting Stieltjes transform of $\bR^\top \bPhi_t^\bullet \bR$ is
    \begin{equation*}
        h_t^\bullet(w) = - \frac{1}{\kappa w} \mathfrak h_t^\bullet(1/w),
    \end{equation*}
    because
    \begin{equation*}
        \frac{1}{m} \tr \Big(\big(\bR^\top \bPhi_t^\bullet \bR - w \bI_m \big)^{-1} \Big) = - \frac{1}{w} \frac{1}{m} \tr \left( \Big(\bR^\top \Big(\bI_d - \frac{1}{w}\bPhi_t^\bullet\Big)\bR \Big)^{-1} \right),
    \end{equation*}
    and $\bR^\top \bR = \bI_m$. Rewriting the self-consistent equation of \cref{Thm:RMTConvergence} in terms of $h_t^\bullet(w)$ gives
    \begin{equation*}
        1 = \int \frac{\tilde \mu_C(\d x)}{1 - \kappa + \kappa h_t^\bullet(w) \big( 2\bar r_t^\mathsf{a}(x) - w \big)}. 
    \end{equation*}
    For the case of $U$, set $h_t=h_t^U$, which has the characterization stated in the proposition by \cref{Thm:RMTConvergence}. Since $2\bar r_t^\mathsf{R} = 2\bar r_t^\mathsf{L} + 2$, the function $w \mapsto h_t(w-2)$ satisfies the equation for $V$ and is positive on $(-\infty,0)$. The same characterization therefore gives $h_t^V(w)=h_t(w-2)$. We recover the Stieltjes transforms of $\rho^{-1} \bU(t) \bU(t)^\top$ and $\rho^{-1} \bV(t) \bV(t)^\top$ by considering the different numbers of zero eigenvalues when passing from the $m \times m$ matrices $\bR^\top \bPhi_t^\bullet \bR$ to the two Gram matrices.
\end{proof}

\subsubsection{Small-Initialization Limit}

We shall now start from the expressions of the Stieltjes transform in \cref{Thm:Stieltjes} and take the small-initialization limit $\rho \to 0$. To do so, we let $w \in \mathbb{C} \setminus \R^+$ be fixed and consider the Stieltjes transforms $\lim_{\rho \to 0} \rho \mathfrak m_t^\bullet(\rho w)$, which are the Stieltjes transforms of $\rho^{-1} \bU(t) \bU(t)^\top$ and $\rho^{-1} \bV(t) \bV(t)^\top$ in the small-initialization limit. As shown in \cref{Prop:FiniteDimensionIntuition}, these are the relevant matrices to study as $\rho \to 0$ while $t = \Theta(1)$. 

Define $h_{t,\rho}^U(w)=\rho g_t(\rho w)$ and $h_{t,\rho}^V(w)=\rho g_t(\rho w-2\rho)$. Their self-consistent equations are
\begin{equation*}
\begin{aligned}
    1 &= \int \frac{\tilde \mu_C(\d x)}{1 - \kappa + \kappa h_{t, \rho}^U(w) (\rho^{-1}r_t^\mathsf L(\gamma_x) - w q_t(\gamma_x))},\\
    1 &= \int \frac{\tilde \mu_C(\d x)}{1 - \kappa + \kappa h_{t, \rho}^V(w) (\rho^{-1}r_t^\mathsf R(\gamma_x) - w q_t(\gamma_x))}.
\end{aligned}
\end{equation*}
We now study the $\rho \to 0$ limit. We have
\begin{equation*}
    \frac{1}{2\rho} r_t^\mathsf L(\gamma_x) \xrightarrow[\rho \to 0]{} \sinh(tx)^2, \hspace{1.5cm} \frac{1}{2\rho} r_t^\mathsf R(\gamma_x) \xrightarrow[\rho \to 0]{} \cosh(tx)^2, \hspace{1.5cm} q_t(\gamma_x) \xrightarrow[\rho \to 0]{} 1,
\end{equation*}
uniformly for $x \in \mathrm{Supp}(\tilde \mu_C)$. 

By \cref{Cor:HolomorphicExtension}, it is enough to prove the convergence for $w < 0$. To do so, we apply \cref{Prop:SelfConsistentGeneral} with parameter $\rho\to0$, $c_\rho=1$, and:
\begin{equation*}
    h_\rho = h^\bullet_{t,\rho}(w), \hspace{1.5cm} a_\rho(x) = \kappa \big(\rho^{-1}r_t^\mathsf a(\gamma_x) - w q_t(\gamma_x) \big),
\end{equation*}
where $\bullet = U,V$ and $\mathsf{a} = \mathsf{L}, \mathsf{R}$. Then, we have, from the previous uniform asymptotics:
\begin{equation*}
    a_\rho(x) \xrightarrow[\rho \to 0]{} \kappa \big( 2 \bar r_t^\mathsf{a}(x) - w \big),
\end{equation*}
which remains strictly positive since $w < 0$. It is clear that the limiting function $\Phi$ is such that $\Phi(u) = 1$ admits a unique solution on $(0, \infty)$. Applying \textbf{2.}, $(i)$ in \cref{Prop:SelfConsistentGeneral} therefore gives the convergence of $h_{t, \rho}^\bullet(w)$ as $\rho \to 0$ to $ h_t^\bullet(w)$, which is characterized by:
\begin{equation}
    1 = \int \frac{\tilde \mu_C(\d x)}{1 - \kappa + \kappa h_t^\bullet(w) (2\bar r_t^\mathsf a(x) - w)}.
\end{equation}

To extend the convergence to $w \in \mathbb{C} \setminus \R^+$, recall that $h_{t, \rho}^\bullet$ and $h_t^\bullet$ are rescaled Stieltjes transforms, hence \cref{Lemma:StieltjesComplex} and \cref{Cor:HolomorphicExtension} guarantee the extension. 

Finally, using the expressions of $\mathfrak m_t^U, \mathfrak m_t^V$ in \cref{Thm:Stieltjes}, we obtain:
\begin{equation*}
    \lim_{\rho \to 0} \rho \mathfrak m_t^\bullet(\rho w) = \varsigma_t^\bullet(w).
\end{equation*}
Together with \cref{Prop:StieltjesSmallInit}, this shows that we recover the same eigenvalue distributions independently of the order in which the limits $\rho \to 0$ and $d \to \infty$ (while $t \geq 0$ is fixed) are taken. 

\subsection{Range Projectors} \label{App:Subsec:ProjectorLimit}

We now study the range projectors in the small-initialization limit, proving \cref{Thm:ProjectorLimit} and then deriving the estimate of \cref{Thm:ProjectorLimit2}. 

\subsubsection{Small Initialization} \label{App:Subsubsec:ProjectorLimit1}

We prove the small-initialization limit in \cref{Thm:ProjectorLimit}. Recall that $\Delta_\rho^\mathsf a(t)^2$ denotes the almost sure high-dimensional limit of the range-projector distance at fixed initialization scale $\rho$, for $\mathsf a=\mathsf L,\mathsf R$. We start from the self-consistent equation for $g_t^\mathsf a$ in \cref{Thm:Projector1}:
\begin{equation*}
    1 = \int \frac{\tilde \mu_C(\d x)}{1 - \kappa + \kappa g_t^\mathsf a r_t^\mathsf a(\gamma_x)},
\end{equation*}
where $r_t^\mathsf a$ is defined in \cref{Thm:Stieltjes} and $\gamma_x = \sqrt{x^2 + \rho^2}$. For any $x \geq 0$ and $\mathsf{a} = \mathsf{L}, \mathsf{R}$, we have the convergence
\begin{equation*}
    \frac{1}{2\rho} r_t^\mathsf{a}(\gamma_x) \xrightarrow[\rho \to 0]{} \bar r_t^\mathsf{a}(x).
\end{equation*}
We now apply \cref{Prop:SelfConsistentGeneral} directly with parameter $\rho\to0$ and
\begin{equation*}
    h_\rho = g_{t,\rho}^\mathsf{a}, \hspace{1.5cm} a_\rho(x) = \kappa r_t^\mathsf{a}(\gamma_x), \hspace{1.5cm} c_\rho = \frac{1}{2\rho},
\end{equation*}
where we made explicit the dependence of $g_t$ with respect to $\rho$. Additionally, $r_t^\mathsf{R}$ does not vanish on the support of $\tilde \mu_C$ and $r_t^\mathsf{L}(\gamma_x)$ is only zero at $x = 0$. In this case the zero set of $r_t^\mathsf{L}(\gamma_x)$ has mass $1 - \beta \kappa^*$ for $\tilde \mu_C$. In the underparameterized regime we consider for $\mathsf{L}$, this mass is smaller than $1 - \kappa$, so that \cref{Prop:SelfConsistentGeneral} can be applied. The limit function in this proposition is:
\begin{equation*}
    \Phi(u) = \int \frac{\tilde \mu_C(\d x)}{1 - \kappa + \kappa u \bar r_t^\mathsf{a}(x)},
\end{equation*}
which admits a unique solution on $(0, \infty)$. As a consequence of \textbf{2.} $(i)$ in \cref{Prop:SelfConsistentGeneral}, we have the convergence:
\begin{equation*}
    2\rho g_{t, \rho}^\mathsf{a} \xrightarrow[\rho \to 0]{} b_t^\mathsf{a},
\end{equation*}
the unique solution of equation \eqref{eq:SCbt}. 

Additionally, \textbf{3.} in \cref{Prop:SelfConsistentGeneral} gives the convergence of $\Delta_\rho^\mathsf{a}(t)^2$, with $f(x) = \1_{x \geq x_*}$ in the $\rho \to 0$ limit, concluding the proof of \cref{Thm:ProjectorLimit}. 

\subsubsection{Monotonicity in Time} \label{App:Subsubsec:ProjectorMonotone}

We prove that $t \mapsto \Delta_0^\mathsf{a}(t)^2$ is strictly decreasing, for $\mathsf a = \mathsf R$, and for $\mathsf a = \mathsf L$ when $\kappa < \beta \kappa^*$. For $t > 0$, let
\begin{equation*}
    f_t^\mathsf{a}(x) = \frac{1}{1 - \kappa + \kappa b_t^\mathsf{a} \bar r_t^\mathsf{a}(x)}.
\end{equation*}
By equation \eqref{eq:SCbt} and \cref{Thm:ProjectorLimit}, we have
\begin{equation} \label{eq:MonotoneConstraint}
    \int f_t^\mathsf{a}(x) \, \tilde \mu_C(\d x) = 1, \hspace{1.5cm} \Delta_0^\mathsf{a}(t)^2 = 2(1-\kappa) \int_{(x_*, \infty)} f_t^\mathsf{a}(x) \, \tilde \mu_C(\d x).
\end{equation}
We fix $0 < t < t'$ and compare $f^\mathsf{a}_{t'}$ with $f^\mathsf{a}_t$. We first show that the ratio $x \mapsto \bar r_{t'}^\mathsf{a}(x) / \bar r_t^\mathsf{a}(x)$ is strictly increasing on $(0, \infty)$. Its logarithmic derivative is
\begin{equation*}
    \frac{\d}{\d x} \log \frac{\bar r_{t'}^\mathsf{a}(x)}{\bar r_t^\mathsf{a}(x)} = 2 \big( h^\mathsf{a}_x(t') - h^\mathsf{a}_x(t) \big),
\end{equation*}
where $h_x^\mathsf{R}(u) = u \tanh(ux)$ and $h_x^\mathsf{L}(u) = u \coth(ux)$. In both cases $h_x^\mathsf{a}$ is strictly increasing in $u > 0$, since
\begin{equation*}
\begin{aligned}
    \frac{\d}{\d u} \big( u \tanh(ux) \big) &= \tanh(ux) + \frac{ux}{\cosh(ux)^2} > 0, \\
    \frac{\d}{\d u} \big( u \coth(ux) \big) &= \frac{\sinh(ux)\cosh(ux) - ux}{\sinh(ux)^2} > 0.
\end{aligned}
\end{equation*}

Now, for $x > 0$, we have $f^\mathsf{a}_{t'}(x) \geq f^\mathsf{a}_t(x)$ if and only if
\begin{equation*}
    \frac{\bar r_{t'}^\mathsf{a}(x)}{\bar r_t^\mathsf{a}(x)} \leq \frac{b_t^\mathsf{a}}{b_{t'}^\mathsf{a}}.
\end{equation*}
Since we just showed that the left side is strictly increasing with respect to $x$, there exists $\bar x \in [0, \infty]$ such that $f^\mathsf{a}_{t'} > f^\mathsf{a}_t$ on $(0, \bar x)$, and $f^\mathsf{a}_{t'} < f^\mathsf{a}_t$ on $(\bar x, \infty)$.

Recall that $x_* > 0$ and that $\tilde \mu_C\big((x_*, \infty)\big) = \min(\kappa, \beta \kappa^*) > 0$. First, if $\bar x \leq x_*$, then $f^\mathsf{a}_{t'} < f^\mathsf{a}_t$ on $(x_*, \infty)$, and equation \eqref{eq:MonotoneConstraint} gives $\Delta_0^\mathsf{a}(t')^2 < \Delta_0^\mathsf{a}(t)^2$. 

If $\bar x > x_*$, the first identity in equation \eqref{eq:MonotoneConstraint}, applied at $t$ and $t'$, gives
\begin{equation*}
    \int_{(x_*, \infty)} \big(f^\mathsf{a}_{t'}(x) - f^\mathsf{a}_t(x) \big) \tilde \mu_C(\d x) = - \int_{[0, x_*]} \big(f^\mathsf{a}_{t'}(x) - f^\mathsf{a}_t(x) \big) \tilde \mu_C (\d x). 
\end{equation*}
Since $[0, x_*] \subset [0, \bar x)$, the integrand on the right is nonnegative. It is moreover strictly positive on a set of positive mass:
\begin{itemize}
    \item If $\kappa < \beta \kappa^*$, on $(0, x_*]$, which has mass $\tilde \mu_C\big((0, x_*]\big) = \beta \kappa^* - \kappa > 0$.
    \item If $\kappa \geq \beta \kappa^*$ (so that $\mathsf a = \mathsf R$), at $x = 0$, which has mass $\tilde \mu_C(\{0\}) = 1 - \beta \kappa^* \geq 1 - \kappa > 0$. Indeed, $\bar r_t^\mathsf{R}(0) = 1$, so the ratio $\bar r_{t'}^\mathsf{R} / \bar r_t^\mathsf{R}$ equals $1$ at $x = 0$ and is larger than $1$ on $(0, \infty)$. Since $\bar x > 0$, this gives $b_t^\mathsf{R} > b_{t'}^\mathsf{R}$, and therefore:
    \begin{equation*}
        f^\mathsf{R}_{t'}(0) = \frac{1}{1 - \kappa + \kappa b_{t'}^\mathsf{R}} > \frac{1}{1 - \kappa + \kappa b_t^\mathsf{R}} = f^\mathsf{R}_t(0).
    \end{equation*}
\end{itemize}
In both cases, the integral over $(x_*,\infty)$ is negative, so that $\Delta_0^\mathsf{a}(t')^2 < \Delta_0^\mathsf{a}(t)^2$.

\subsubsection{Long Times} \label{App:Subsubsec:ProjectorLimit2}

Let us prove \cref{Thm:ProjectorLimit2} and derive the asymptotics for $\Delta_0^\mathsf a(t)^2$ as $t \to \infty$. 

\paragraph{Step 1. Useful lemmas.} We start with two elementary lemmas. The first one is a standard Laplace estimate at the left edge of the spectrum.

\begin{lemma} \label{Lemma:LaplaceIntegral}
    Let $\phi \colon \R^+ \to \R^+$ be bounded, measurable and compactly supported, and let $\alpha, x_0 > 0$. Assume that:
    \begin{equation*}
        \phi(x_0 + h) \underset{h \to 0^+}{\sim} \phi_0 h^\ell,
    \end{equation*}
    for some $\phi_0 > 0$ and $\ell \geq 0$. Then:
    \begin{equation*}
        \int_{x_0}^\infty e^{-\alpha x t} \phi(x) \d x \underset{t \to \infty}{\sim} \frac{\phi_0 \Gamma(\ell + 1)}{\alpha^{\ell+1} t^{\ell+1}} e^{-\alpha x_0 t}. 
    \end{equation*}
\end{lemma}

\begin{proof}
    Let us perform the substitution $u = \alpha t(x- x_0)$, so that:
    \begin{equation*}
        I(t) := \int_{x_0}^\infty e^{-\alpha x t} \phi(x) \d x = \frac{e^{- \alpha x_0 t}}{\alpha t} \int_0^\infty e^{-u} \phi \left( x_0 + \frac{u}{\alpha t} \right) \d u. 
    \end{equation*}
    Now, consider:
    \begin{equation*}
        \psi(h) = \frac{\phi(x_0 + h)}{h^\ell},
    \end{equation*}
    so that $\psi(h) \xrightarrow[h \to 0^+]{} \phi_0$. Moreover, $\psi$ is bounded on $\R^+$: it is bounded on some interval $(0, \delta]$ by the previous limit, and by $\|\phi\|_\infty \delta^{-\ell}$ on $[\delta, \infty)$. Therefore, we have:
    \begin{equation*}
        I(t) = \frac{e^{-\alpha x_0 t}}{\alpha^{\ell+1} t^{\ell+1}} \int_0^\infty u^\ell e^{-u} \psi \left( \frac{u}{\alpha t} \right) \d u. 
    \end{equation*}
    Now, as a consequence of the dominated convergence theorem:
    \begin{equation*}
        \int_0^\infty u^\ell e^{-u} \psi \left( \frac{u}{\alpha t} \right) \d u \xrightarrow[t \to \infty]{} \phi_0 \int_0^\infty u^\ell e^{-u} \d u = \phi_0 \Gamma(\ell + 1),
    \end{equation*}
    which gives the result.
\end{proof}

\begin{lemma} \label{Lemma:BoundaryLayer}
    Let $c_t > 0$ be such that $\log c_t = o(t)$ as $t \to \infty$, and let $\phi$ be bounded, compactly supported and continuous at $x_*$. Then:
    \begin{equation*}
        \int \left(\frac{1}{1 + c_t e^{2t(x-x_*)}} - \1_{x < x_*} \right)\phi(x) \d x
        =
        -\frac{\phi(x_*)}{2t}\log c_t
        +
        o\left(\frac{1+|\log c_t|}{t}\right).
    \end{equation*}
    In particular, if $\phi(x_*)>0$ and the left-hand side is $o(t^{-1})$, then $c_t \xrightarrow[t \to \infty]{} 1$. In this case:
    \begin{equation*}
        \int_{x_*}^\infty \frac{\phi(x) \d x}{1 + c_t e^{2t(x-x_*)}}
        \underset{t \to \infty}{\sim}
        \frac{\phi(x_*)}{2t}\log(2).
    \end{equation*}
\end{lemma}

\begin{proof}
    We introduce:
    \begin{equation*}
        x_t = x_* - \frac{\log c_t}{2t},
    \end{equation*}
    so that $c_t e^{2t(x-x_*)}=e^{2t(x-x_t)}$ and $x_t \xrightarrow[t \to \infty]{}  x_*$. We decompose:
    \begin{equation*}
    \begin{aligned}
        \int \left(\frac{1}{1+c_t e^{2t(x-x_*)}}-\1_{x<x_*}\right)\phi(x)\d x
        &=
        \int \left(\frac{1}{1+e^{2t(x-x_t)}}-\1_{x<x_t}\right)\phi(x)\d x \\
        &\hspace{1.5cm}
        + \int \left(\1_{x<x_t}-\1_{x<x_*}\right)\phi(x)\d x.
    \end{aligned}
    \end{equation*}
    The first term is $o(t^{-1})$. Indeed, with the substitution $x=x_t+\xi t^{-1}$, it becomes:
    \begin{equation*}
        \frac{1}{t}\int \left(\frac{1}{1+e^{2\xi}}-\1_{\xi<0}\right)\phi\left(x_t+\frac{\xi}{t}\right)\d \xi.
    \end{equation*}
    The integral converges to zero by dominated convergence. For the second term, we have:
        \begin{equation*}
        \int \left(\1_{x<x_t}-\1_{x<x_*}\right)\phi(x)\d x
        =
        \int_{x_*}^{x_t}\phi(x)\d x
        =
        (x_t-x_*)\phi(x_*)+o(|x_t-x_*|).
    \end{equation*}
    Since $x_t-x_* = -(2t)^{-1}\log c_t$, this proves the first asymptotic. Additionally, if $\phi(x_*) > 0$ and the left-hand side is $o(t^{-1})$, then we must have $\log c_t = o(1)$, which implies $c_t \xrightarrow[t \to \infty]{} 1$.

    For the second asymptotic, we use the same technique. We get, after splitting the integral and introducing $x = x_t + \xi t^{-1}$, with $x_t$ defined as above:
    \begin{equation*}
        \int_{x_*}^\infty \frac{\phi(x) \d x}{1 + c_t e^{2t(x-x_*)}} = \frac{1}{t} \int_0^\infty \phi \left( x_t + \frac{\xi}{t} \right) \frac{\d \xi}{1 + e^{2\xi}} + \frac{1}{t} \int \phi \left( x_t + \frac{\xi}{t} \right) \frac{\d \xi}{1 + e^{2\xi}}\left( \1_{\xi > \frac{\log c_t}{2}} - \1_{\xi > 0} \right). 
    \end{equation*}
    The second term is $o(t^{-1})$ because $\log c_t \to 0$ and $\phi$ is bounded near $x_*$. For the first term, dominated convergence gives:
    \begin{equation*}
    \begin{aligned}
        \int_0^\infty \phi \left( x_t + \frac{\xi}{t} \right) \frac{\d \xi}{1 + e^{2\xi}} &\xrightarrow[t \to \infty]{} \phi(x_*) \int_0^\infty \frac{\d \xi}{1 + e^{2\xi}} = \frac{\phi(x_*)}{2} \log(2). 
    \end{aligned}
    \end{equation*}
\end{proof}

\paragraph{Step 2. Self-consistent functions.} We start by studying $b_t^\mathsf a$, for $\mathsf a=\mathsf L,\mathsf R$. It is the solution of the self-consistent equation
\begin{equation} \label{eq:SCbt2}
    1= \int \frac{\tilde \mu_C(\d x)}{1 - \kappa + \kappa b_t^\mathsf a \bar r_t^\mathsf a(x)}.
\end{equation}
We first determine its long-time behavior.
\begin{lemma} \label{Lemma:Asymptoticbt1}
    Consider the setting of \cref{Thm:ProjectorLimit2}, and $b_t^\mathsf a$ solving equation \eqref{eq:SCbt2}. Then, as $t \to \infty$:
    \begin{itemize}
        \item If $\kappa > \beta \kappa^*$:
        \begin{equation*} 
            b_t^\mathsf R \xrightarrow[t \to \infty]{} 1 - \frac{\beta \kappa^*}{\kappa}.
        \end{equation*}
        \item If $\kappa < \beta \kappa^*$, both $b_t^\mathsf L,b_t^\mathsf R$ converge to zero exponentially fast. Moreover, for all $y > 0$
        \begin{equation*}
            b_t^\mathsf{a} e^{2ty} \xrightarrow[t \to \infty]{}
            \begin{cases}
                0, & y < x_*, \\
                +\infty, & y > x_*,
            \end{cases}
        \end{equation*}
        where $x_*$ is characterized by $\kappa = \beta \kappa^* \displaystyle \int_{x_*}^\infty \phi_C(x) \d x$.
    \end{itemize}
\end{lemma}
We only need the asymptotic behavior of $b_t^\mathsf R$ when $\kappa > \beta \kappa^*$, since the left range-projector distance is then zero. The threshold $x_*$ is unique under the assumptions of \cref{Thm:ProjectorLimit2}, since $\phi_C(x_*) > 0$.

\begin{proof}
    We start with the case $\kappa > \beta \kappa^*$, and $\mathsf{a} = \mathsf{R}$. We apply \cref{Prop:SelfConsistentGeneral} directly with parameter $t\to\infty$ and
    \begin{equation*}
        h_t = b_t^\mathsf{R}, \hspace{1.5cm} a_t(x) = \kappa \bar r_t^\mathsf{R}(x),
    \end{equation*}
    and $c_t = 1$. As we have $\bar r_t^\mathsf{R}(x) \xrightarrow[t \to \infty]{} +\infty$ for $x > 0$ and $\bar r_t^\mathsf{R}(0) = 1$, the limiting function in \cref{Prop:SelfConsistentGeneral} is given by:
    \begin{equation*}
        \Phi(u) = \frac{1 - \beta \kappa^*}{1 - \kappa + \kappa u}.
    \end{equation*}
    Therefore, $b_t^\mathsf{R}$ converges to the desired limit in the overparameterized case by \textbf{2.} $(i)$ in \cref{Prop:SelfConsistentGeneral}. 

    Regarding the underparameterized case, consider $y > 0$ and again take
    \begin{equation*}
        h_t = b_t^\mathsf{a}, \hspace{1.5cm} a_t(x) = \kappa \bar r_t^\mathsf{a}(x),
    \end{equation*}
    with $c_t = e^{-2ty}$. We have:
    \begin{equation*}
        c_t a_t(x) \xrightarrow[t \to \infty]{}
        \begin{cases}
            0, & x < y, \\
            +\infty, & x > y.
        \end{cases}
    \end{equation*}
    Hence the limit function in \cref{Prop:SelfConsistentGeneral} is given by:
    \begin{equation*}
        \Phi_y(u) = \frac{1 - T_C(y)}{1 - \kappa},
    \end{equation*}
    for every $u > 0$. Since $x_*$ is the only point satisfying $T_C(x_*) = \kappa$, applying \textbf{2.} $(ii)$ and $(iii)$ in \cref{Prop:SelfConsistentGeneral} leads to the desired result for the underparameterized case. 
\end{proof}

We now refine the asymptotic behavior of $b_t^\mathsf a$ in the underparameterized case.
\begin{lemma} \label{Lemma:Asymptoticbt2}
    In the setting of \cref{Lemma:Asymptoticbt1}, with $\kappa < \beta \kappa^*$, we have:
    \begin{equation*}
        b_t^\mathsf a \underset{t \to \infty}{\sim} \frac{4(1-\kappa)}{\kappa}e^{-2tx_*}.
    \end{equation*}
\end{lemma}

\begin{proof}
    We consider:
    \begin{equation} \label{eq:defct}
        c_t^\mathsf a = \frac{\kappa}{4(1-\kappa)} b_t^\mathsf a e^{2tx_*},
    \end{equation}
    and fix $y \in (0, x_*)$. By \cref{Lemma:Asymptoticbt1}, $b_t^\mathsf a=o(e^{-2yt})$. Hence the contribution of the atom at zero to equation \eqref{eq:SCbt2} is
    \begin{equation*}
        \frac{1-\beta \kappa^*}{1-\kappa+\kappa b_t^\mathsf a \bar r_t^\mathsf a(0)} = \frac{1-\beta \kappa^*}{1-\kappa} + o(e^{-2yt}).
    \end{equation*}
    We also use the identity:
    \begin{equation*}
        1 = \frac{1 - \beta \kappa^*}{1 - \kappa} + \frac{\beta \kappa^*}{1 - \kappa} \int_0^{x_*}\phi_C(x) \d x,
    \end{equation*}
    which follows from the definition of $x_*$. Subtracting this identity from equation \eqref{eq:SCbt2} gives:
    \begin{equation*}
        \int \left( \frac{1}{1 + 4 c_t^\mathsf a e^{-2 t x_*}\bar r_t^\mathsf a(x)} - \1_{x < x_*} \right) \phi_C(x) \d x =
        o(e^{-2yt}).
    \end{equation*}
    Now, we have:
    \begin{equation*}
        \left| \frac{1}{1+4c_t^\mathsf a e^{-2tx_*}\bar r_t^\mathsf a(x)} - \frac{1}{1+c_t^\mathsf a e^{2t(x-x_*)}} \right| \leq c_t^\mathsf a e^{-2tx_*} \left| 4\bar r_t^\mathsf a(x) - e^{2tx} \right|.
    \end{equation*}
    Now, $4\bar r_t^\mathsf a(x)=\big(e^{tx}\mp e^{-tx}\big)^2$, depending on $\mathsf a=\mathsf L,\mathsf R$. Hence $4\bar r_t^\mathsf a(x)-e^{2tx}$ remains bounded, and the right-hand side is $o(e^{-2yt})$. Therefore,
    \begin{equation*}
        \int \left( \frac{1}{1 + c_t^\mathsf a e^{2t (x - x_*)}} - \1_{x < x_*} \right) \phi_C(x) \d x = o(e^{-2yt}).
    \end{equation*}
    \cref{Lemma:Asymptoticbt1} guarantees that $\log(c_t^\mathsf{a}) = o(t)$, hence by \cref{Lemma:BoundaryLayer}, $c_t^\mathsf{a} \xrightarrow[t \to \infty]{} 1$, which concludes the proof.
\end{proof}

\paragraph{Step 3. Asymptotics for the distance.} We now compute the asymptotic behavior of $\Delta_0^\mathsf a(t)^2$. 

Let us start with the overparameterized case $\kappa > \beta \kappa^*$. Since:
\begin{equation*}
    \bar r_t^\mathsf R(x) \underset{t \to \infty}{\sim} \frac{1}{4} e^{2tx}
\end{equation*}
uniformly on the support of $\phi_C$, using the limit of $b_t^\mathsf R$ in \cref{Lemma:Asymptoticbt1} gives
\begin{equation*}
    \Delta_0^\mathsf R(t)^2 \underset{t \to \infty}{\sim} \frac{8\beta \kappa^*(1 - \kappa)}{\kappa - \beta \kappa^*} \int e^{-2xt} \phi_C(x) \d x,
\end{equation*}
and the asymptotic equivalent follows from \cref{Lemma:LaplaceIntegral} with $\alpha=2$.

In the underparameterized case, \cref{Thm:ProjectorLimit} and equation \eqref{eq:defct} give
\begin{equation*}
    \Delta_0^\mathsf a(t)^2 = 2 \beta \kappa^* \int_{x_*}^\infty \frac{\phi_C(x) \d x}{1 + 4c_t^\mathsf a e^{-2tx_*} \bar r_t^\mathsf a(x)}. 
\end{equation*}
We can replace $4 e^{-2tx_*}\bar r_t^\mathsf a(x)$ by $e^{2t(x-x_*)}$ up to an exponentially vanishing correction. Since $c_t^\mathsf{a}$ converges to $1$, \cref{Lemma:BoundaryLayer} gives
\begin{equation*}
    \Delta_0^\mathsf a(t)^2 \underset{t \to \infty}{\sim} \frac{\beta \kappa^* \phi_C(x_*)}{t} \log(2),
\end{equation*}
which is the desired result. 

\subsection{Top Singular Subspaces} \label{App:Subsec:SpectralProjectors}

In this section we prove \cref{Thm:TopEigenvectors}. Unlike the range-projector distances, we do not compute an exact expression of the top singular vector distances in the high-dimensional limit. To achieve this, one could use the two-point Stieltjes transforms (see \cref{Thm:DoubleStieltjes} for an expression of their high-dimensional limit), which encode the necessary spectral overlaps. However, this route is technically involved and we do not pursue it here. Instead, we obtain direct and non-sharp convergence rates.

We start by giving some prior lemmas, then treat the finite-dimensional case, and finally the high-dimensional limit. We study the distance, for $\mathsf{a} = \mathsf{R}, \mathsf{L}$:
\begin{equation*}
    \Delta_{k, \rho}^{\mathsf{a}}(t)^2 = \frac{1}{d} \big\| \bP_{k, \rho}^\mathsf{a}(t) - \bP_k^{C, \mathsf{a}} \big\|_F^2, 
\end{equation*}
and we recall that $\bP^\mathsf{L}_{k, \rho}(t), \bP^\mathsf{R}_{k, \rho}(t)$ project onto the top-$k$ left and right singular vectors of the prediction $\bZ(t)$, and likewise for $\bP_k^{C, \mathsf{L}}, \bP_k^{C, \mathsf{R}}$ with $\bC$. 

\paragraph{Relation with subspace iteration.} In the limit $\rho\to0$, the rescaled Gram matrices from \cref{Prop:FiniteDimensionIntuition} have the same non-zero eigenvalues as $2\bM^\mathsf a(t)$, with $\bM^\mathsf a(t)=\bY_t^\mathsf a(\bY_t^\mathsf a)^\top$, where
\begin{equation*}
    \bY_t^\mathsf L=\sinh(t\bS)\bR, \hspace{1.5cm} \bY_t^\mathsf R=\cosh(t\bS)\bR,
\end{equation*}
and $\bS=(\bC^\top\bC)^{1/2}$. This is the continuous version of a filtered subspace iteration \citep{stewart1976simultaneous,halko2011finding,saibaba2019randomized}. Writing $f_t^\mathsf L(x)=\sinh(tx)$ and $f_t^\mathsf R(x)=\cosh(tx)$, the ratio between the lower and upper components at the cutoff $k$ is:
\begin{equation*}
    \frac{f_t^\mathsf a(\sigma_{k+1})}{f_t^\mathsf a(\sigma_k)}
    =O\!\left(e^{-t(\sigma_k-\sigma_{k+1})}\right).
\end{equation*}
This is the main idea behind the estimate given in equation \eqref{eq:FiniteDimTopProjectorRate}. Our new approach concerns the proportional-rank limit $k \sim \alpha d$. In this case the bounds derive from the same idea as the finite-dimensional case, and the non-exponential asymptotic in equation \eqref{eq:HighDimTopProjectorRate2} naturally arises from the fact that the gap $\sigma_k - \sigma_{k+1}$ vanishes as $d \to \infty$. 

\paragraph{Step 1. Gram matrices.} We start to express the distances $\Delta_{k, \rho}^\mathsf{a}(t)^2$ in terms of projections obtained from PSD matrices, related to those in \cref{Lemma:BlockDecomposition1}.

\begin{lemma} \label{Lemma:TopProj1}
    Consider the assumptions of \cref{Thm:TopEigenvectors}. For $t>0$ and whenever $k \leq \min(m, m^*)$,
    \begin{equation*}
        \Delta_{k, \rho}^{\mathsf{a}}(t)^2 = \frac{1}{d} \big\| \bQ_{k, \rho}^\mathsf{a}(t) - \bP_k^{C, \mathsf{R}} \big\|_F^2,
    \end{equation*}
    where $\bQ_{k, \rho}^\mathsf{a}(t)$ is the projection onto the top-$k$ eigenvectors of
    \begin{equation} \label{eq:DefMrho}
        \bM_\rho^\mathsf{a}(t) = r_t^\mathsf{a}(\bGamma_\rho)^{1/2} \bR
        \Big(\bR^\top q_t(\bGamma_\rho) \bR \Big)^{-1} \bR^\top r_t^\mathsf{a}(\bGamma_\rho)^{1/2},
    \end{equation}
    where $r_t^\mathsf{L}, r_t^\mathsf{R}, q_t$ are defined in \cref{Thm:Stieltjes}. 
\end{lemma}

\begin{proof}
    As a consequence of \cref{Prop:SVDPrediction}, we have the equalities:
    \begin{equation*}
        \bP_{k, \rho}^\mathsf{L}(t) = \bP_{k, \rho}^U(t), \hspace{1.5cm} \bP_{k, \rho}^\mathsf{R}(t) = \bP_{k, \rho}^V(t),
    \end{equation*}
    where $\bP_{k, \rho}^U(t), \bP_{k, \rho}^V(t)$ are the projectors onto the top-$k$ eigenvectors of $\bU(t) \bU(t)^\top$ and $\bV(t) \bV(t)^\top$. This equality is guaranteed whenever $k \leq \min(m, m^*)$. We now use the notation of \cref{Lemma:BlockDecomposition1} and the identities:
    \begin{equation*}
        2 \rho \chi_t^U(x)^2 = r_t^\mathsf{L}(\gamma_\rho(x)), \hspace{1.5cm} 2 \rho \chi_t^V(x)^2 = r_t^\mathsf{R}(\gamma_\rho(x)),
    \end{equation*}
    where $\gamma_\rho(x) = \sqrt{x^2 + \rho^2}$. Hence \cref{Lemma:BlockDecomposition1} guarantees that $\bV(t) \bV(t)^\top = \bM_\rho^\mathsf{R}(t)$. This proves the claim for $\mathsf{R}$.

    For the left projection, we recall the singular value decomposition $\bC = \sum_{i=1}^{m^*} \sigma_i \bu_i \bv_i^\top$. We have:
    \begin{equation*}
        \bO r_t^\mathsf{L}(\bGamma_\rho)^{1/2} = \sqrt{2 \rho} \chi_t^U(\bC), \hspace{1.5cm} \bO = \sum_{i=1}^{m^*} \bu_i \bv_i^\top,
    \end{equation*}
    and therefore $\bU(t) \bU(t)^\top = \bO \bM_\rho^\mathsf{L}(t) \bO^\top$. Since $r_t^\mathsf{L}(\bGamma_\rho)$ vanishes on $\mathrm{Ker}(\bC)$, the image of $\bM_\rho^\mathsf{L}(t)$ is included in $\mathrm{Span}(\bv_1, \dots, \bv_{m^*})$, where $\bO$ is an isometry. Hence there is an exact correspondence between the top-$k$ eigenvectors of $\bU(t) \bU(t)^\top$ and those of $\bM_\rho^\mathsf{L}(t)$ through $\bO$. Therefore,
    \begin{equation*}
        \bP_{k,\rho}^U(t) = \bO \bQ_{k,\rho}^\mathsf{L}(t) \bO^\top, \hspace{1.5cm} \bP_k^{C,\mathsf{L}} = \bO \bP_k^{C,\mathsf R} \bO^\top,
    \end{equation*}
    and
    \begin{equation*}
        \big\| \bP_{k,\rho}^U(t) - \bP_k^{C,\mathsf{L}} \big\|_F^2 = \Big\| \bO \bQ_{k,\rho}^\mathsf{L}(t) \bO^\top - \bO \bP_k^{C,\mathsf R} \bO^\top \Big\|_F^2.
    \end{equation*}
    Since $\bO^\top \bO$ is the orthogonal projection onto $\mathrm{Span}(\bv_1, \dots, \bv_{m^*})$ and $\mathrm{Im} \big( \bQ_{k,\rho}^\mathsf{L}(t) \big)$, $\mathrm{Im} \big(\bP_k^{C,\mathsf R} \big)$ are included in this space, the last quantity is equal to $\big\| \bQ_{k,\rho}^\mathsf{L}(t) - \bP_k^{C,\mathsf R} \big\|_F^2$, which concludes the proof. 
\end{proof}

The proof of this lemma shows that $\bQ_{k, \rho}^\mathsf{R}(t) = \bP_{k, \rho}^\mathsf{R}(t)$, and that $\bQ_{k, \rho}^\mathsf{L}(t)$ is obtained by transporting $\bP_{k, \rho}^\mathsf{L}(t)$ into the right singular basis of $\bC$. Hence both $\bQ_{k, \rho}^\mathsf{L}(t), \bQ_{k, \rho}^\mathsf{R}(t)$ are compared with $\bP_k^{C, \mathsf{R}}$. 

\paragraph{Step 2. A projector distance bound.} The next lemma gives the main bound used in the following. To do so, we define the quantity:
\begin{equation} \label{eq:DefOperatorN}
    \nu_{\bP}(\bA) = \inf_{\bu \in \mathrm{Im} \, \bP, \| \bu\| = 1} \| \bA \bu \|^2. 
\end{equation}
$\nu_\bP(\bA)$ is the smallest eigenvalue of $\bP \bA^\top \bA \bP$ restricted to the subspace $\mathrm{Im}(\bP)$. It satisfies $\bu^\top \bA^\top \bA \bu \geq \nu_{\bP}(\bA) \|\bu\|^2$ for all $\bu \in \mathrm{Im}(\bP)$.

\begin{lemma} \label{Lemma:TopProjMinMax}
    Let $\bD, \bH\in\mathcal S_d^+(\R)$, with $\bH \succeq \bI_d$, let
    $\bR\in\mathrm{St}(d,m)$, and define
    \begin{equation*}
        \bM = \bD^{1/2} \bR(\bR^\top \bH \bR)^{-1}\bR^\top \bD^{1/2}.
    \end{equation*}
    Let $\bP$ be an orthogonal projector of rank $\ell \leq m$ that commutes
    with $\bD$, and $\bQ$ be the projection onto the top-$\ell$ eigenvectors of $\bM$. Then:
    \begin{equation} \label{eq:TopProjMinMax}
        \big\| \bP - \bQ \big\|_F^2 \leq \frac{2 \| \bH\|_\text{op}}{\nu_\bP(\bR^\top) \nu_\bP(\bD^{1/2})} \tr \big( (\bI_d - \bP) \bD \big),
    \end{equation}
    whenever $\nu_\bP(\bR^\top)$, $\nu_\bP(\bD^{1/2})$ are non-zero. 
\end{lemma}

\begin{proof}
    Since $\bP, \bQ$ are both projectors, it is clear that $\big\| \bP - \bQ \big\|_F^2 = 2 \tr \big( (\bI_d - \bP) \bQ \big)$. Now, let us denote by $\lambda_\ell(\bM)$ the $\ell^\text{th}$ largest eigenvalue of $\bM$. On the image of $\bQ$, $\bM$ has eigenvalues at least $\lambda_\ell(\bM)$, hence $\bM \succeq \lambda_\ell(\bM) \bQ$. Additionally, $\bM \preceq \bD$ since $\bR^\top \bH \bR \succeq \bI_m$ and $\bR \bR^\top \preceq \bI_d$. Putting these facts together, and using that $\bI_d - \bP$ is PSD, we obtain the bound:
    \begin{equation} \label{eq:BoundsProj1}
        \big\| \bP - \bQ \big\|_F^2 \leq \frac{2}{\lambda_\ell(\bM)} \tr \big( (\bI_d - \bP) \bD \big). 
    \end{equation}
    It remains to lower-bound $\lambda_\ell(\bM)$. To do so, we use the Courant--Fischer min--max principle for eigenvalues of a symmetric matrix:
    \begin{equation} \label{eq:minmaxprinciple}
        \lambda_\ell(\bM) = \max_{\mathrm{dim} \, \mathcal{E} = \ell} \min_{\by \in \mathcal{E}, \|\by\|=1} \by^\top \bM \by \geq \min_{\by \in \mathrm{Im}(\bP), \|\by\|=1} \by^\top \bM \by,
    \end{equation}
    where we chose $\mathcal{E} = \mathrm{Im}(\bP)$ in the inequality. For $\by \in \mathrm{Im}(\bP)$, let $\bz = \bD^{1/2} \by$. Since
    $\bP$ commutes with $\bD$, we still have $\bz \in \mathrm{Im}(\bP)$. Moreover, since $\bR^\top \bR = \bI_m$,
    \begin{equation*}
        \bR^\top \bH \bR \preceq \|\bH\|_{\mathrm{op}} \bI_m,
    \end{equation*}
    and therefore
    \begin{equation*}
        \by^\top \bM \by \geq \frac{1}{\|\bH\|_{\mathrm{op}}} \bz^\top \bR \bR^\top \bz \geq \frac{1}{\|\bH\|_{\mathrm{op}}} \nu_\bP(\bR^\top) \|\bz\|^2 \geq \frac{1}{\|\bH\|_{\mathrm{op}}} \nu_\bP(\bR^\top) \nu_\bP(\bD^{1/2}),
    \end{equation*}
    where we used the definition of $\nu_\bP$ in equation \eqref{eq:DefOperatorN}. Combining this lower bound on $\lambda_\ell(\bM)$ with equation \eqref{eq:BoundsProj1} proves the desired bound.
\end{proof}

We now apply this lemma to $\bM_\rho^\mathsf{a}(t)$ defined in equation \eqref{eq:DefMrho}. More precisely, we choose:
\begin{equation*}
    \bH = q_t(\bGamma_\rho), \hspace{1.5cm} \bD = r_t^\mathsf{a}(\bGamma_\rho), \hspace{1.5cm} \bP=\bP_k^{C,\mathsf R}.
\end{equation*}
Recall that $\bGamma_\rho = \big(\bC^\top \bC + \rho^2 \bI_d \big)^{1/2}$. Since $q_t(\gamma) \geq 1$ for $\gamma \geq \rho$, we have indeed $\bH \succeq \bI_d$. Additionally, $\bGamma_\rho$ commutes with $\bP$. Therefore:
\begin{equation*}
    \nu_\bP(\bD^{1/2}) \geq \inf_{x \in [\sigma_k, L]} r_t^\mathsf{a}(\gamma_\rho(x)), \hspace{1.5cm} \gamma_\rho(x) = \sqrt{x^2 + \rho^2},
\end{equation*}
where we used that the singular values of $\bC$ are uniformly bounded by $L$ by \cref{Assumption1:Teacher}. In finite dimension, one can take $L = \sigma_1$. Additionally, $\bP$ and $\bD$ commute, so that:
\begin{equation*}
    \tr \big( (\bI_d - \bP) \bD \big) = \sum_{i=k+1}^d r_t^\mathsf{a}(\gamma_i), \hspace{1.5cm} \gamma_i = \sqrt{\sigma_i^2 + \rho^2}. 
\end{equation*}
Finally, due to the expression of $\bH$:
\begin{equation*}
    \| \bH \|_\text{op} \leq \sup_{x \in [0, L]} q_t(\gamma_\rho(x)). 
\end{equation*}
Therefore, combining \cref{Lemma:TopProj1} and \cref{Lemma:TopProjMinMax}, we get:
\begin{equation} \label{eq:BoundsProj2}
    \Delta_{k, \rho}^{\mathsf{a}}(t)^2 \leq \frac{2}{ \nu_k d} \left(\sup_{x \in [0, L]} q_t(\gamma_\rho(x)) \right) \frac{\sum_{i=k+1}^d r_t^\mathsf{a}(\gamma_i)}{\inf_{x \in [\sigma_k, L]} r_t^\mathsf{a}(\gamma_\rho(x))}.
\end{equation}
We denoted $\nu_k = \nu_\bP(\bR^\top)$. This bound forms the basis of the finite and high-dimensional computations. As $\rho \to 0$, we will show that the last ratio leads to the desired convergence rates: the numerator only concerns the singular values of $\bC$ at most $\sigma_{k+1}$ while the denominator only involves those greater than or equal to $\sigma_k$.

\paragraph{Step 3. Lower-bounding random projections.} Before deriving the convergence rates, we show that $\nu_k$ remains bounded away from zero. We provide a slightly more general version that we will need in the high-dimensional limit.
\begin{lemma} \label{Lemma:StiefelProjectorConditioning}
    Let $\bR$ be uniformly distributed on $\mathrm{St}(d, m)$, and let $\bP$ be an orthogonal projector of rank $\ell \leq m$, independent of $\bR$. Then, with probability one
    \begin{equation*}
        \nu_{\bP}(\bR^\top) > 0.
    \end{equation*}
    Moreover, if as $d \to \infty$, we have $m \sim \kappa d$ and $\ell \sim \alpha d$, where $0 < \alpha < \kappa < 1$, then there exists a constant $C_{\alpha, \kappa} > 0$ only depending on $\alpha, \kappa$ such that, with probability one, $\nu_{\bP}(\bR^\top) > C_{\alpha, \kappa}$ for $d$ large enough.
\end{lemma}

\begin{proof}
    As in the proof of \cref{Thm:RMTConvergenceReal}, we write $\bR=\bG(\bG^\top\bG)^{-1/2}$, where $\bG$ has i.i.d. $\N(0,1/m)$ entries and is independent of $\bP$. We can then choose a basis in which
    \begin{equation*}
        \bP = \begin{pmatrix} \bI_\ell & \bzero \\ \bzero & \bzero
        \end{pmatrix}, \hspace{1.5cm} \bG = \begin{pmatrix} \bG_1 \\ \bG_2 \end{pmatrix},
    \end{equation*}
    where $\bG_1 \in \R^{\ell \times m}$. Since $\nu_\bP(\bR^\top)$ is the smallest eigenvalue of $\bR \bR^\top$ restricted to $\mathrm{Im}(\bP)$, and $\bG^\top \bG \preceq \lambda_{\max}(\bG^\top \bG) \bI_m$, we have
    \begin{equation*}
        \nu_\bP(\bR^\top) = \lambda_{\min} \Big( \bG_1(\bG^\top \bG)^{-1} \bG_1^\top \Big) \geq \frac{\lambda_{\min} \big( \bG_1 \bG_1^\top \big)}{\lambda_{\max} \big( \bG^\top \bG \big)}.
    \end{equation*}
    Since $\ell \leq m$, the Gaussian matrix $\bG_1$ has full row rank with probability one, so that $\nu_\bP(\bR^\top) > 0$. In the high-dimensional limit, since $\alpha < \kappa$, the limits of the extreme eigenvalues of sample covariance matrices \citep{yin1988largest,bai1993limit} give the almost sure convergence
    \begin{equation*}
        \lambda_\text{min}(\bG_1 \bG_1^\top) \xrightarrow[d \to \infty]{} \left(1 - \sqrt{\frac{\alpha}{\kappa}} \right)^2, \hspace{1.5cm} \lambda_{\max}(\bG^\top\bG) \xrightarrow[d \to \infty]{} \left(1+ \frac{1}{\sqrt{\kappa}} \right)^2,
    \end{equation*}
    so that any $C_{\alpha, \kappa}$ strictly smaller than the ratio of these limits works.
\end{proof}

\paragraph{Step 4. The finite-dimensional case.} We now prove equation \eqref{eq:FiniteDimTopProjectorRate}. As $\rho \to 0$, we have the uniform convergence on compact sets
\begin{equation} \label{eq:UniformCVsmallinit}
    \frac{1}{2\rho} r_t^\mathsf{L} \big( \gamma_\rho(x) \big) \xrightarrow[\rho \to 0]{} \bar{r}_t^\mathsf{L}(x), \hspace{1.2cm} \frac{1}{2\rho} r_t^\mathsf{R} \big( \gamma_\rho(x) \big) \xrightarrow[\rho \to 0]{} \bar{r}_t^\mathsf{R}(x), \hspace{1.2cm} q_t \big( \gamma_\rho(x) \big) \xrightarrow[\rho \to 0]{} 1,
\end{equation}
where
\begin{equation*}
    \bar{r}_t^\mathsf{L}(x) = \sinh(tx)^2, \hspace{1.5cm} \bar{r}_t^\mathsf{R}(x) = \cosh(tx)^2. 
\end{equation*}
Hence, equation \eqref{eq:BoundsProj2} and the monotonicity of $\bar r_t^\mathsf{a}$ give:
\begin{equation*}
    \limsup_{\rho \to 0} \Delta_{k, \rho}^{\mathsf{a}}(t)^2  \leq \frac{2(d-k)}{\nu_k d} \frac{\bar r_t^\mathsf{a}(\sigma_{k+1})}{\bar r_t^\mathsf{a}(\sigma_k)}. 
\end{equation*}
Since $\nu_k > 0$ with probability one by \cref{Lemma:StiefelProjectorConditioning}, and $\bar r_t^\mathsf{a}(x) \underset{t \to \infty}{\sim} \frac{1}{4} e^{2tx}$ for $x > 0$, we obtain equation \eqref{eq:FiniteDimTopProjectorRate}.

\paragraph{Step 5. Intermediate projectors.} Before proving the high-dimensional case, we introduce slightly different projectors than the top-$k$. Indeed, taking the high-dimensional limit in equation \eqref{eq:BoundsProj2} would require knowing the limit of the $k^\text{th}$ and $(k+1)^\text{th}$ singular values of $\bC$, which is a finer requirement than \cref{Assumption1:Teacher}. However, the cutoff $x_\alpha$ defined in \cref{Thm:TopEigenvectors} asymptotically selects a fraction $\alpha$ of the singular directions.

Let us consider:
\begin{equation*}
    \bP_\alpha^C = \sum_{i=1}^{m^*} \1_{\sigma_i \geq x_\alpha} \bv_i \bv_i^\top. 
\end{equation*}
Although $\sigma_k$ can differ from $x_\alpha$ in the high-dimensional limit, the characterization of $x_\alpha$ guarantees that only $o(d)$ singular values can remain in the gap between $\sigma_k$ and $x_\alpha$. Let us denote by $\ell$ the rank of $\bP_\alpha^C$ and consider $\bQ_{\alpha, \rho}^\mathsf{a}(t)$ the projector onto the top-$\ell$ eigenspaces of $\bM_\rho^\mathsf{a}(t)$. Then $\ell / d$ converges toward $\tilde \mu_C \big( [x_\alpha, \infty) \big) = \alpha$. As a consequence, since $k \sim \alpha d$:
\begin{equation*}
    \frac{1}{d} \big\| \bP_\alpha^C - \bP_k^{C, \mathsf{R}} \big\|_F^2 = \frac{1}{d} \big|\mathrm{rank}(\bP_\alpha^C) - k \big|\xrightarrow[d \to \infty]{} 0. 
\end{equation*}
The same identity holds for $\bQ_{\alpha, \rho}^\mathsf{a}(t)$ and $\bQ_{k, \rho}^\mathsf{a}(t)$. Now, as a consequence of the triangle inequality, we have:
\begin{equation*}
    \frac{1}{\sqrt{d}} \Big| \big\| \bQ_{k, \rho}^\mathsf{a}(t) - \bP_k^{C, \mathsf{R}} \big\|_F - \big\| \bQ_{\alpha, \rho}^\mathsf{a}(t) - \bP_\alpha^C \big\|_F \Big| \leq \frac{1}{\sqrt{d}} \big\| \bP_\alpha^C - \bP_k^{C, \mathsf{R}} \big\|_F + \frac{1}{\sqrt{d}} \big\| \bQ_{\alpha, \rho}^\mathsf{a}(t) - \bQ_{k, \rho}^\mathsf{a}(t) \big\|_F. 
\end{equation*}
Therefore, \cref{Lemma:TopProj1} guarantees that:
\begin{equation*}
    \limsup_{d \to \infty} \Delta_{k, \rho}^\mathsf{a}(t)^2 \leq \limsup_{d \to \infty} \frac{1}{d} \big\| \bQ_{\alpha, \rho}^\mathsf{a}(t) - \bP_\alpha^C \big\|_F^2. 
\end{equation*}
Applying again \cref{Lemma:TopProjMinMax} to these new projectors and using the same bounds as at the end of Step 2, we obtain a similar bound to the one in equation \eqref{eq:BoundsProj2}:
\begin{equation} \label{eq:BoundsProj3}
    \limsup_{d \to \infty} \Delta_{k, \rho}^\mathsf{a}(t)^2 \leq \limsup_{d \to \infty} \frac{2}{ \nu_\alpha d}\left(\sup_{x \in [0, L]} q_t(\gamma_\rho(x)) \right) \frac{\sum_{i=1}^d \1_{\sigma_i < x_\alpha} r_t^\mathsf{a}(\gamma_i)}{\inf_{x \in [x_\alpha, L]} r_t^\mathsf{a}(\gamma_\rho(x))}. 
\end{equation}
In this case $\nu_\alpha = \nu_{\bP_\alpha^C}(\bR^\top)$ (see equation \eqref{eq:DefOperatorN}). This bound only involves the fixed cutoff $x_\alpha$ and does not require assumptions on the asymptotic behavior of the individual singular values of $\bC$. 

\paragraph{Step 6. High-dimensional case.} For the high-dimensional limit, we take $d \to \infty$ first. Since the asymptotic spectral distribution of $(\bC^\top \bC)^{1/2}$ is $\tilde \mu_C$, and due to the characterization of $x_\alpha$, we have the convergence:
\begin{equation*}
    \frac{1}{d} \sum_{i=1}^d \1_{\sigma_i < x_\alpha} r_t^\mathsf{a}(\gamma_i) \xrightarrow[d \to \infty]{} \int_0^{x_\alpha} r_t^\mathsf{a}(\gamma_\rho(x)) \tilde \mu_C(\d x). 
\end{equation*}
We apply the high-dimensional part of \cref{Lemma:StiefelProjectorConditioning} to $\nu_\alpha$, so that there exists a non-random constant $C_\alpha > 0$ (independent of $d, \rho, t$) such that with probability one, $\nu_\alpha > C_\alpha$ for $d$ large enough (guaranteed by Step 3). Therefore, equation \eqref{eq:BoundsProj3} guarantees that, with probability one:
\begin{equation*}
    \limsup_{d \to \infty} \Delta_{k, \rho}^{\mathsf{a}}(t)^2 \leq \frac{2}{C_\alpha} \sup_{x \in [0, L]} q_t(\gamma_\rho(x)) \frac{\int_0^{x_\alpha} r_t^\mathsf{a}(\gamma_\rho(x)) \tilde \mu_C(\d x)}{\inf_{x \in [x_\alpha, L]} r_t^\mathsf{a}(\gamma_\rho(x))}.
\end{equation*}
Now taking the $\rho \to 0$ limit, using the convergence in equation \eqref{eq:UniformCVsmallinit} and the monotonicity of $\bar r_t^\mathsf{a}$, we get:
\begin{equation*}
    \limsup_{\rho \to 0} \limsup_{d \to \infty} \Delta_{k, \rho}^{\mathsf{a}}(t)^2 \leq \frac{2}{C_\alpha} \frac{\int_0^{x_\alpha} \bar r_t^\mathsf{a}(x) \tilde \mu_C(\d x)}{\bar r_t^\mathsf{a}(x_\alpha)}. 
\end{equation*}
Now, we have the bound, for all $t > 0$ and $x \in [0, x_\alpha]$:
\begin{equation*}
    \frac{\sinh(tx)^2}{\sinh(tx_\alpha)^2} \leq e^{-2t(x_\alpha-x)}, \hspace{1.5cm} \frac{\cosh(tx)^2}{\cosh(tx_\alpha)^2} \leq 4e^{-2t(x_\alpha-x)},
\end{equation*}
so that for some $\tilde C_\alpha > 0$:
\begin{equation} \label{eq:TopHighDimFinalBound}
    \limsup_{\rho \to 0} \limsup_{d \to \infty} \Delta_{k, \rho}^\mathsf{a}(t)^2 \leq \tilde C_\alpha \int_0^{x_\alpha} e^{-2t(x_\alpha - x)} \tilde \mu_C(\d x).  
\end{equation}
Whenever $\tilde \mu_C \big( (x_\alpha - \epsilon_\alpha, x_\alpha) \big) = 0$, the integral is $O(e^{-2\epsilon_\alpha t})$, which proves the estimate in equation \eqref{eq:HighDimTopProjectorRate1}. In all cases, using the decomposition of $\tilde \mu_C$ and the boundedness of $\phi_C$ (\cref{Assumption1:Teacher}), we have:
\begin{equation*}
\begin{aligned}
    \int_0^{x_\alpha} e^{-2t(x_\alpha - x)} \tilde \mu_C(\d x) &\leq (1 - \beta \kappa^*) e^{-2tx_\alpha} + \beta \kappa^* \|\phi_C\|_\infty \int_0^{x_\alpha} e^{-2t(x_\alpha - x)} \d x \\
    & \leq (1 - \beta \kappa^*) e^{-2tx_\alpha} + \frac{\beta \kappa^* \|\phi_C\|_\infty}{2t}.
\end{aligned}
\end{equation*}
Together with equation \eqref{eq:TopHighDimFinalBound}, this gives the estimate in equation \eqref{eq:HighDimTopProjectorRate2}. This concludes the proof of \cref{Thm:TopEigenvectors}.

\subsection{Saddle-to-Saddle Dynamics} \label{App:Subsec:STS}

We now prove \cref{Thm:FiniteDimensionLogTimes}. Set $t_\rho= s \log(\rho^{-1})$ and let $\mathcal I_s$ index the singular values selected at rescaled time $s$,
\begin{equation*}
    \mathcal I_s = \{i \leq m^* : \sigma_i>(2s)^{-1}\} = \{1, \dots, k\}.
\end{equation*}
By assumption, $k\leq m$ and no singular value lies at the threshold. We extend $\big(\bv_1, \dots, \bv_{m^*} \big)$ into an orthonormal right singular basis of $\bC$ and set $\sigma_i = 0$ for $i > m^*$. We consider the solutions given in \cref{Lemma:BlockDecomposition1} and use the associated notation. For $i \in \mathcal I_s$, we let
\begin{equation*}
    d_{\rho,i} = q_{t_\rho}(\gamma_i) - 1, \hspace{1.5cm} \bD_\rho = \mathrm{diag} \big(d_{\rho,i}, i \in \mathcal I_s \big) \in \R^{k \times k},
\end{equation*}
and
\begin{equation*}
    \bR_s = \begin{pmatrix}
        \bv_1^\top \bR \\ \vdots \\ \bv_k^\top \bR
    \end{pmatrix} \in \R^{k \times m}.
\end{equation*}

\paragraph{Step 1. Scalar asymptotics.} In the proof, we will use the following lemma.
\begin{lemma} \label{Lemma:AsymptoticSmallRho1}
    Let $s>0$, $t_\rho = s \log(\rho^{-1})$ and $x_s = (2s)^{-1}$. As $\rho \to 0$, for $\mathsf{a} = \mathsf{L}, \mathsf{R}$ and $x > 0$, we have:
    \begin{equation} \label{eq:AsymptoticFuncs}
        q_{t_\rho}(\gamma_x)-1
        \underset{\rho\to0}{\sim}\frac{1}{2x}\rho^{1-2sx},
        \hspace{1.5cm}
        r_{t_\rho}^{\mathsf a}(\gamma_x)
        \underset{\rho\to0}{\sim}\frac{1}{2}\rho^{1-2sx}.
    \end{equation}
    Therefore:
    \begin{itemize}
        \item For $0 \leq x < x_s$, as $\rho \to 0$, $q_{t_\rho}(\gamma_x)$ converges to $1$ and both $r_{t_\rho}^\mathsf{L}(\gamma_x), r_{t_\rho}^\mathsf{R}(\gamma_x)$ converge to $0$. 
        \item For $x > x_s$, $q_{t_\rho}(\gamma_x), r_{t_\rho}^\mathsf{L}(\gamma_x), r_{t_\rho}^\mathsf{R}(\gamma_x)$ diverge to $+\infty$. 
    \end{itemize}
\end{lemma}

\begin{proof}
    Let $x > 0$. The formulas in \cref{Thm:Stieltjes} can be written as
    \begin{align*}
        q_{t_\rho}(\gamma_x)-1
        &=\frac{\rho}{2\gamma_x}\big(\rho^{-2s\gamma_x}-\rho^{2s\gamma_x}\big)
        +\frac{\rho^2}{2\gamma_x^2}
        \big(\rho^{-2s\gamma_x}+\rho^{2s\gamma_x}-2\big),\\
        r_{t_\rho}^{\mathsf L}(\gamma_x)
        &=\frac{\rho}{2}\left(1-\frac{\rho^2}{\gamma_x^2}\right)
        \big(\rho^{-s\gamma_x}-\rho^{s\gamma_x}\big)^2,\\
        r_{t_\rho}^{\mathsf R}(\gamma_x)
        &=\frac{\rho}{2}
        \left[
        \left(1+\frac{\rho}{\gamma_x}\right)\rho^{-s\gamma_x}
        +\left(1-\frac{\rho}{\gamma_x}\right)\rho^{s\gamma_x}
        \right]^2.
    \end{align*}
    As $\rho \to 0$ we have $\gamma_x \to x$ and also $\rho^{\gamma_x - x} \to 1$, which leads to equation \eqref{eq:AsymptoticFuncs}. The limits follow for $x > 0$. For $x = 0$, we have $q_{t_\rho}(\rho)=\rho^{-2s\rho}$, $r_{t_\rho}^{\mathsf L}(\rho)=0$, and $r_{t_\rho}^{\mathsf R}(\rho)=2\rho^{1-2s\rho}$, which gives the desired result. 
\end{proof}
Therefore, for $i \in \mathcal{I}_s$ we have:
\begin{equation} \label{eq:LogTimeScalarAsymptotics}
\begin{aligned}
    d_{\rho,i} \underset{\rho \to 0}{\sim} \frac{1}{2\sigma_i} \rho^{1 - 2s\sigma_i}, \hspace{1.2cm} \frac{\sqrt{2\rho} \chi_{t_\rho}^U(\sigma_i)}{\sqrt{d_{\rho,i}}} \xrightarrow[\rho \to 0]{} \sqrt{\sigma_i}, \hspace{1.2cm} \frac{\sqrt{2\rho} \chi_{t_\rho}^V(\sigma_i)}{\sqrt{d_{\rho,i}}}
    \xrightarrow[\rho \to 0]{} \sqrt{\sigma_i}.
\end{aligned}
\end{equation}
In particular $d_{\rho,i} \to \infty$ for $i \in \mathcal I_s$. For $i \notin \mathcal I_s$, we rather have $q_{t_\rho}(\gamma_i) \to 1$ and
\begin{equation*}
    \sqrt{2\rho}\chi_{t_\rho}^U(\sigma_i) \xrightarrow[\rho \to 0]{} 0,
    \hspace{1.5cm} \sqrt{2\rho} \chi_{t_\rho}^V(\sigma_i) \xrightarrow[\rho \to 0]{} 0.
\end{equation*}
Whenever $\sigma_i = 0$, we have $q_{t_\rho}(\rho) = e^{2 \rho t_\rho} \to 1$ and $\chi_{t_\rho}^V(0) = e^{\rho t_\rho}$, therefore the previous statements also apply. 

\paragraph{Step 2. Limit of the normalization matrix.}
Recall the expressions from \cref{Lemma:BlockDecomposition1}:
\begin{equation} \label{eq:GramMatricesSmallrho}
\begin{aligned}
    \bU(t_\rho) \bU(t_\rho)^\top &= 2 \rho \chi_{t_\rho}^U (\bC) \bR \bK_\rho^{-1} \bR^\top \chi_{t_\rho}^U(\bC)^\top, \\
    \bV(t_\rho) \bV(t_\rho)^\top &= 2 \rho \chi_{t_\rho}^V (\bS) \bR \bK_\rho^{-1} \bR^\top \chi_{t_\rho}^V(\bS), \\
    \bU(t_\rho) \bV(t_\rho)^\top &= 2 \rho \chi_{t_\rho}^U (\bC) \bR \bK_\rho^{-1} \bR^\top \chi_{t_\rho}^V(\bS),
\end{aligned}
\end{equation}
where $\bS = (\bC^\top \bC)^{1/2}$, and $\bK_\rho = \bR^\top q_{t_\rho}(\bGamma_\rho) \bR$. Then, we have, since $\bR^\top \bR = \bI_m$
\begin{equation*}
    \bK_\rho = \bI_m + \bR_s^\top \bD_\rho \bR_s + \bR^\top \sum_{i \notin \mathcal{I}_s} \big(q_{t_\rho}(\gamma_i) - 1 \big) \bv_i\bv_i^\top \bR.  
\end{equation*}
We call $\bE_\rho$ the last term, which converges to zero as $\rho \to 0$. Now, we let:
\begin{equation*}
    \bM_\rho = \bD_\rho^{1/2} \bR_s \bK_\rho^{-1} \bR_s^\top \bD_\rho^{1/2} =  \bD_\rho^{1/2} \bR_s \Big( \bI_m + \bR_s^\top \bD_\rho \bR_s + \bE_\rho \Big)^{-1} \bR_s^\top \bD_\rho^{1/2}. 
\end{equation*}
Using the Woodbury formula, we have:
\begin{equation*}
    \bM_\rho = \bI_k - \Big( \bI_k + \bD_\rho^{1/2} \bR_s (\bI_m + \bE_\rho)^{-1} \bR_s^\top \bD_\rho^{1/2} \Big)^{-1}. 
\end{equation*}
Since $\bE_\rho \to \bzero$, the matrix $\bI_m + \bE_\rho$ is invertible for $\rho$ small enough. As $\rho \to 0$, the matrix $\bR_s (\bI_m + \bE_\rho)^{-1} \bR_s^\top$ converges to $\bR_s \bR_s^\top$. By the definition of $\bR_s$ and \cref{Lemma:GenericPositionStiefel}, $\bR_s$ has full row rank with probability one, so $\bR_s \bR_s^\top$ is positive definite. Moreover, all diagonal coefficients of $\bD_\rho$ diverge by equation \eqref{eq:LogTimeScalarAsymptotics}, so the inverse in the above equation tends to zero. Hence, we have $\bM_\rho \to \bI_k$ as $\rho \to 0$. 

\paragraph{Step 3. Limit of the dynamics.}
To conclude the proof, decompose:
\begin{equation*}
\begin{aligned}
    \sqrt{2 \rho} \chi_{t_\rho}^U(\bC) &= \sum_{i \in \mathcal{I}_s} \sqrt{2 \rho} \chi_{t_\rho}^U(\sigma_i) \bu_i \bv_i^\top + \sum_{i \notin \mathcal{I}_s} \sqrt{2 \rho} \chi_{t_\rho}^U(\sigma_i) \bu_i \bv_i^\top, \\
    \sqrt{2 \rho} \chi_{t_\rho}^V(\bS) &= \sum_{i \in \mathcal{I}_s} \sqrt{2 \rho} \chi_{t_\rho}^V(\sigma_i) \bv_i \bv_i^\top + \sum_{i \notin \mathcal{I}_s} \sqrt{2 \rho} \chi_{t_\rho}^V(\sigma_i) \bv_i \bv_i^\top. 
\end{aligned}
\end{equation*}
For $i \notin \mathcal{I}_s$, we have shown that $\sqrt{2 \rho} \chi_{t_\rho}^U(\sigma_i)$ and $\sqrt{2 \rho} \chi_{t_\rho}^V(\sigma_i)$ converge to zero as $\rho \to 0$. Since $\bK_\rho \succeq \bI_m$, we get:
\begin{equation*}
\begin{aligned}
    \sum_{i \notin \mathcal{I}_s} \sqrt{2 \rho} \chi_{t_\rho}^U(\sigma_i) \bu_i \bv_i^\top \bR \bK_\rho^{-1/2} &\xrightarrow[\rho \to 0]{} \bzero, \\
    \sum_{i \notin \mathcal{I}_s} \sqrt{2 \rho} \chi_{t_\rho}^V(\sigma_i) \bv_i \bv_i^\top \bR \bK_\rho^{-1/2} &\xrightarrow[\rho \to 0]{} \bzero.
\end{aligned}
\end{equation*}
Therefore, when computing the limit of the matrices in equation \eqref{eq:GramMatricesSmallrho}, only the terms in $\mathcal{I}_s$ remain, and we obtain:
\begin{equation*}
\begin{aligned}
    \lim_{\rho \to 0} \bU(t_\rho) \bU(t_\rho)^\top &= \lim_{\rho \to 0} \sum_{i,j \in \mathcal{I}_s} \frac{2 \rho \chi_{t_\rho}^U (\sigma_i) \chi_{t_\rho}^U (\sigma_j)}{\sqrt{d_{\rho, i} d_{\rho, j}}} \bM_\rho^{ij} \bu_i \bu_j^\top, \\
    \lim_{\rho \to 0} \bV(t_\rho) \bV(t_\rho)^\top &= \lim_{\rho \to 0} \sum_{i,j \in \mathcal{I}_s} \frac{2 \rho \chi_{t_\rho}^V (\sigma_i) \chi_{t_\rho}^V (\sigma_j)}{\sqrt{d_{\rho, i} d_{\rho, j}}} \bM_\rho^{ij} \bv_i \bv_j^\top, \\
    \lim_{\rho \to 0} \bU(t_\rho) \bV(t_\rho)^\top &= \lim_{\rho \to 0} \sum_{i,j \in \mathcal{I}_s} \frac{2 \rho \chi_{t_\rho}^U (\sigma_i) \chi_{t_\rho}^V (\sigma_j)}{\sqrt{d_{\rho, i} d_{\rho, j}}} \bM_\rho^{ij} \bu_i \bv_j^\top.
\end{aligned}
\end{equation*}
Using the asymptotics in equation \eqref{eq:LogTimeScalarAsymptotics} and the fact that $\bM_\rho$ converges to $\bI_k$, we get:
\begin{equation*}
\begin{aligned}
    \bU(t_\rho) \bU(t_\rho)^\top &\xrightarrow[\rho \to 0]{} \sum_{i \in \mathcal{I}_s} \sigma_i \bu_i \bu_i^\top, \\
    \bV(t_\rho) \bV(t_\rho)^\top &\xrightarrow[\rho \to 0]{} \sum_{i \in \mathcal{I}_s} \sigma_i \bv_i \bv_i^\top, \\
    \bU(t_\rho) \bV(t_\rho)^\top &\xrightarrow[\rho \to 0]{} \sum_{i \in \mathcal{I}_s} \sigma_i \bu_i \bv_i^\top. 
\end{aligned}
\end{equation*}
This is precisely the result of \cref{Thm:FiniteDimensionLogTimes}. 

\paragraph{Step 4. Saddle-to-saddle interpretation.}
We finally make the saddle-to-saddle interpretation precise. Write
\begin{equation*}
    \bP_I = (\bu_i)_{i\in \mathcal{I}_s}, \hspace{1.2cm} \bQ_I = (\bv_i)_{i\in \mathcal{I}_s},
    \hspace{1.2cm} \bSigma_I = \mathrm{diag}(\sigma_i, i\in \mathcal{I}_s).
\end{equation*}
Since $\bU(t_\rho)\bU(t_\rho)^\top$ and $\bV(t_\rho)\bV(t_\rho)^\top$ converge, the pair $(\bU(t_\rho), \bV(t_\rho))$ is bounded as $\rho \to 0$. Let $(\bU_s, \bV_s)$ be any of its limit points, and $\bZ_s = \bU_s \bV_s^\top$. By \cref{Thm:FiniteDimensionLogTimes}:
\begin{equation*}
    \bU_s \bU_s^\top = \bP_I \bSigma_I \bP_I^\top, \hspace{1.2cm} \bV_s \bV_s^\top = \bQ_I \bSigma_I \bQ_I^\top, \hspace{1.2cm} \bU_s \bV_s^\top = \bP_I \bSigma_I \bQ_I^\top.
\end{equation*}
As a consequence of the relationships for $\bU_s, \bV_s$ there exist $\bO_U, \bO_V \in \R^{k \times m}$ such that $\bO_U \bO_U^\top = \bI_k$, $\bO_V \bO_V^\top = \bI_k$ and we have $\bU_s = \bP_I \bSigma_I^{1/2} \bO_U$ and $\bV_s = \bQ_I \bSigma_I^{1/2} \bO_V$. The above equation for $\bU_s \bV_s^\top$ implies that $\bO_U \bO_V^\top = \bI_k$ and therefore $\bO_U = \bO_V$, which we denote by $\bO$ from now on.

From these identities it easily follows that:
\begin{equation*}
    (\bZ_s - \bC) \bV_s = \bzero, \hspace{1.5cm} (\bZ_s - \bC)^\top \bU_s = \bzero,
\end{equation*}
since $\bZ_s - \bC$ only contains singular directions of $\bC$ orthogonal to $\bP_I$ and $\bQ_I$. Since the two previous matrices are precisely the gradient of the loss (with respect to $\bU$ and $\bV$), the point $(\bU_s, \bV_s)$ is a critical point of the loss. 

Let us now show this is a strict saddle. For $k < \min(m, m^*)$, choose a direction that is not learned yet $j \notin \mathcal{I}_s$ with $j \leq m^*$, so that $\sigma_j > 0$, and a vector $\be \in \R^m$ of unit norm in $\mathrm{Ker}(\bO)$, and consider for some $\epsilon > 0$ and $\lambda \in \R$:
\begin{equation*}
    \bU_\epsilon = \bU_s + \epsilon \bu_j \be^\top, \hspace{1.5cm} \bV_\epsilon = \bV_s + \lambda \epsilon \bv_j \be^\top.
\end{equation*}
Then, we have 
\begin{equation*}
    \bU_\epsilon \bV_\epsilon^\top = \bZ_s + \lambda \epsilon^2 \bu_j\bv_j^\top.
\end{equation*}
Since the loss is given by $\frac{1}{2} \|\bU \bV^\top - \bC\|_F^2$, up to an additive constant, this leads to
\begin{equation*}
    \L(\bU_\epsilon, \bV_\epsilon) - \L(\bU_s, \bV_s) = \frac{1}{2} \Big( (\sigma_j - \lambda \epsilon^2)^2 - \sigma_j^2 \Big) = - \sigma_j \lambda \epsilon^2 + \frac{1}{2} \lambda^2 \epsilon^4. 
\end{equation*}
If $\epsilon$ is small enough, the above term is negative for $\lambda > 0$ and positive for $\lambda < 0$. Since $\epsilon \mapsto (\bU_\epsilon, \bV_\epsilon)$ is a straight line through the critical point, the $\epsilon^2$ term shows that the Hessian of the loss has a negative eigenvalue. Hence every intermediate limiting point given in \cref{Thm:FiniteDimensionLogTimes} is a strict saddle. Between two consecutive thresholds $(2\sigma_i)^{-1}$, the selected index set $\mathcal{I}_s$ is constant, and crossing a threshold moves the limit from one critical point to the next. This is the expected saddle-to-saddle dynamics.

\subsection{Spectrum in the Small-Initialization Limit} \label{App:Subsec:StieltjesLimit}

We shall now prove \cref{Thm:StieltjesLimit}. 

\subsubsection{Eigenvalues} \label{App:Subsubsec:StieltjesLimit1}

Let us start with the limit for $\mathfrak m_t^U,\mathfrak m_t^V$ when $t=s\log(\rho^{-1})$. For this proof only, write $g_t^\mathsf L(w)=g_t(w)$ and $g_t^\mathsf R(w)=g_t(w-2\rho)$. Then, for $\mathsf a=\mathsf L,\mathsf R$,
\begin{equation*}
    1 = \int \frac{\tilde \mu_C(\d x)}{1 - \kappa + \kappa g_t^\mathsf a(w) (r_t^\mathsf a(\gamma_x) - w q_t(\gamma_x))}.
\end{equation*}
We now study the asymptotics as $\rho \to 0$ for $g_t^\mathsf L,g_t^\mathsf R$, with $t_\rho = s\log(\rho^{-1})$ and $x_s=(2s)^{-1}$. For clarity, we write $g_{t_\rho}^\mathsf a=g_\rho^\mathsf a$. We only consider $w < 0$ and will apply \cref{Cor:HolomorphicExtension} to the Stieltjes transforms. 

We apply \cref{Prop:SelfConsistentGeneral} directly with parameter $\rho\to0$ and:
\begin{equation*}
    h_\rho = g_\rho^\mathsf{a}(w), \hspace{1.5cm} a_\rho(x) = \kappa\big(r_{t_\rho}^\mathsf a(\gamma_x)-wq_{t_\rho}(\gamma_x)\big),
\end{equation*}
and $c_\rho = 1$. Then, \cref{Lemma:AsymptoticSmallRho1} gives:
\begin{equation*}
    a_\rho(x)\xrightarrow[\rho \to 0]{} a_\infty(x) = \begin{cases}
        -\kappa w, & x < x_s,\\
        +\infty,& x > x_s.
    \end{cases}
\end{equation*}
Hence the limit function in \cref{Prop:SelfConsistentGeneral} is:
\begin{equation*}
    \Phi(u) = \frac{1 - T_C(x_s)}{1 - \kappa - \kappa w u}. 
\end{equation*}
Since $x_s > x_*$, we have $T_C(x_s) < \kappa$, and $\Phi(u) = 1$ has the unique solution on $(0, \infty)$
\begin{equation*}
    u_\infty = - \frac{\kappa - T_C(x_s)}{\kappa w} > 0. 
\end{equation*}
Therefore, using \textbf{2.} $(i)$ in \cref{Prop:SelfConsistentGeneral}, $g_\rho^\mathsf{a}(w)$ converges to this value, which we denote by $g_0^\mathsf{a}(w)$.  

Let us now show the limit of the Stieltjes transforms when $w < 0$. Back to the expression of the Stieltjes transforms $\mathfrak m_t^U, \mathfrak m_t^V$ in equations \eqref{eq:StieltjesU}, \eqref{eq:StieltjesV}, we focus on the integral term:
\begin{equation*}
    \kappa g_\rho^\mathsf a(w) \int \frac{q_{t_\rho}(\gamma_x)}{1 - \kappa + \kappa g_\rho^\mathsf a(w)(r_{t_\rho}^\mathsf a(\gamma_x) - w q_{t_\rho}(\gamma_x))} \tilde \mu_C(\d x). 
\end{equation*}
The integrand satisfies the uniform bound
\begin{equation*}
    0 \leq \frac{\kappa g_\rho^\mathsf{a}(w) q_{t_\rho}(\gamma_x)}{1 - \kappa + \kappa g_\rho^\mathsf{a}(w)\big(r_{t_\rho}^\mathsf a(\gamma_x) - w q_{t_\rho}(\gamma_x)\big)} \leq -\frac{1}{w}.
\end{equation*}
Moreover, for every $x \neq x_s$, \cref{Lemma:AsymptoticSmallRho1} shows that it converges pointwise to
\begin{equation*}
    \frac{\kappa g_\rho^\mathsf a(w) q_{t_\rho}(\gamma_x)}{1 - \kappa + \kappa g_\rho^\mathsf a(w)(r_{t_\rho}^\mathsf a(\gamma_x) - w q_{t_\rho}(\gamma_x))}
    \xrightarrow[\rho \to 0]{}
    \begin{cases}
        \dfrac{\kappa g_0^\mathsf a(w)}{1 - \kappa - \kappa w g_0^\mathsf a(w)}, & x < x_s, \\
        \dfrac{1}{x-w}, & x > x_s.
    \end{cases}
\end{equation*}
Since $\tilde \mu_C$ has no atom at $x_s$, dominated convergence, the expression of $g_0^\mathsf{a}(w)$, and \cref{Thm:Stieltjes} then give equations \eqref{eq:StieltjesLogtimesU}, \eqref{eq:StieltjesLogtimesV} for $w < 0$.

The extension to $w \in \mathbb{C} \setminus \R^+$ is guaranteed by \cref{Cor:HolomorphicExtension}. Indeed, the RHS of equations \eqref{eq:StieltjesLogtimesU}, \eqref{eq:StieltjesLogtimesV} are holomorphic on $\mathbb{C} \setminus \R^+$ and the LHS are Stieltjes transforms, hence \cref{Lemma:StieltjesComplex} guarantees uniform boundedness. 

\subsubsection{Singular Values} \label{App:Subsubsec:SingularValuesSmallrho}

Let us now prove equation \eqref{eq:StieltjesLogtimesZ}. To do so, we recall equation \eqref{eq:StieltjesUtoZ} that links the empirical measure of the singular values of the prediction $\bZ(t)$ to the spectral measure of $\bU(t) \bU(t)^\top$:
\begin{equation*}
    \int f(x) \mu_t^Z(\d x) = \left(1 - \frac{\tau}{\beta} \right) f(0) + \frac{\tau}{\beta} \int f\left( \sqrt{x(x+2\rho)} \right) \mu_t^U(\d x),
\end{equation*}
for any bounded continuous test function $f$. Now, in the $\rho \to 0$ limit, we have $\sqrt{x(x+2\rho)} \to x$ for $x \geq 0$. One has to be careful since $\mu_t^U$ depends on $\rho$. For simplicity let us write $\mu_\rho^U := \mu_{t_\rho}^U$ and $\overline \mu^U$ its weak limit as $\rho \to 0$. Let $\theta_\rho(x) = \sqrt{x(x+2\rho)}$. Then:
\begin{equation} \label{eq:Bounddoublelimit}
    \left| \int f \circ \theta_\rho \, \d \mu_\rho^U - \int f \, \d \overline \mu^U \right| \leq \left| \int f \circ \theta_\rho \, \d \mu_\rho^U - \int f \, \d \mu_\rho^U \right| + \left| \int f \, \d \mu_\rho^U - \int f \, \d \overline \mu^U \right|. 
\end{equation}
We first prove the convergence for bounded Lipschitz functions. If $f$ is Lipschitz with constant $K$, then:
\begin{equation*}
    \sup_{x \in \R^+} \big| f(\theta_\rho(x)) - f(x) \big| \leq K \sup_{x \in \R^+} \left| \sqrt{x(x+2 \rho)} - x \right|.
\end{equation*}
Now, since we have, for all $x > 0$ and trivially at $x=0$:
\begin{equation*}
    0 \leq \sqrt{x(x+2 \rho)} - x = \frac{2 \rho x}{\sqrt{x(x+2\rho)} + x} \leq \rho,
\end{equation*}
the previous supremum goes to zero, and the first term of equation \eqref{eq:Bounddoublelimit} satisfies, since $\mu_\rho^U$ is a probability measure supported on $\R^+$:
\begin{equation*}
    \left| \int f \circ \theta_\rho \, \d \mu_\rho^U - \int f \, \d \mu_\rho^U \right| \leq \sup_{x \in \R^+} \big| f(\theta_\rho(x)) - f(x) \big| \xrightarrow[\rho \to 0]{} 0.
\end{equation*}
The convergence of the second term in equation \eqref{eq:Bounddoublelimit} is a simple consequence of the weak convergence of $\mu_\rho^U$ toward $\overline \mu^U$. As a consequence, for all bounded Lipschitz functions:
\begin{equation*}
    \int f(x) \mu_{t_\rho}^Z(\d x) \xrightarrow[\rho \to 0]{} \left(1 - \frac{\tau}{\beta} \right) f(0) + \frac{\tau}{\beta} \int f(x) \overline \mu^U(\d x). 
\end{equation*}
This implies the weak convergence of $\mu_{t_\rho}^Z$. Since $x \mapsto (x-w)^{-1}$ is bounded and Lipschitz on $\R^+$ for every $w \in \mathbb{C}\setminus \R^+$, the Stieltjes transforms converge to:
\begin{equation*}
    \mathfrak m_{t_\rho}^Z(w) \xrightarrow[\rho \to 0]{} - \frac{1}{w} \left(1 - \frac{\tau}{\beta} \right) + \frac{\tau}{\beta} \lim_{\rho \to 0} \mathfrak m_{t_\rho}^U(w). 
\end{equation*}
This leads to equation \eqref{eq:StieltjesLogtimesZ}.

\subsubsection{Wasserstein Distance}

Let us now prove the expression of the Wasserstein distance in equation \eqref{eq:WassersteinDistance}. Denote:
\begin{equation*}
    \overline{\mathfrak m}_s(w) = - \frac{1}{w} \left(1 - \frac{T_C(x_s)}{\beta} \right) + \kappa^* \int_{x_s}^\infty \frac{\phi_C(x) \d x}{x - w}.
\end{equation*}
Therefore, the associated measure is
\begin{equation*}
    \mu_s(\d x) =  \left(1 - \frac{T_C(x_s)}{\beta} \right) \delta_0(\d x) + \kappa^* \1_{x > x_s} \phi_C(x) \d x.
\end{equation*}
Recall the decomposition of $\mu_C$ in \cref{Assumption1:Teacher}. Then, the absolutely continuous part of $\mu_s$ is exactly the restriction of the positive part of $\mu_C$ to $(x_s,\infty)$, and the mass removed from $[0,x_s]$ is added to the atom at zero. Hence, since the monotone map is optimal for the quadratic cost on the real line \citep[Section~2.2]{santambrogio2015optimal}, the optimal transport from $\mu_C$ to $\mu_s$ is given by $\Theta_s(x)=0$ for $x\leq x_s$ and $\Theta_s(x)=x$ for $x>x_s$. Therefore:
\begin{equation*}
    W_2^2(\mu_s, \mu_C) = \kappa^* \int_0^{x_s} x^2 \phi_C(x) \d x. 
\end{equation*}

\subsection{Loss Function} \label{App:Subsec:Loss}

We now prove \cref{Thm:LossTimescales}, which computes the limit of the normalized squared distance:
\begin{equation*}
    \frac{1}{\min(p, d)} \big\| \bZ(t) - \bC \big\|_F^2. 
\end{equation*}
By equation \eqref{eq:LossDistance}, this quantity is proportional to the excess empirical loss. With our choices of scalings, the loss diverges as $d \to \infty$, and to obtain a quantity of order one we need to rescale it by a factor proportional to $d$. In the proof it is convenient to first use the $d$-normalized distance
\begin{equation*}
    \mathcal{D}_d(t)= \frac{1}{d} \big\| \bZ(t) - \bC \big\|_F^2.
\end{equation*}
Since $\min(p,d) / d \to \beta$, the normalization in \cref{Thm:LossTimescales} is recovered by multiplying the limit of $\mathcal{D}_d(t)$ by $\beta^{-1}$.

\subsubsection{High-Dimensional Limit for the Distance} 

We start by proving the high-dimensional limit for the normalized distance $\mathcal{D}_d(t)$.
\begin{proposition} \label{Prop:LossHighdim}
    Using the notation of \cref{Thm:Stieltjes}, in the high-dimensional limit, under \cref{Assumption1}, $\mathcal{D}_d(t)$ has the almost sure limit:
    \begin{equation} \label{eq:LossHighdim}
    \begin{aligned}
        \lim_{d \to \infty} \mathcal{D}_d(t) &= \int \left( \frac{\vartheta_t \sqrt{r_t^\mathsf L(\gamma_x) r_t^\mathsf R(\gamma_x)}}{1 - \kappa + \vartheta_t q_t(\gamma_x)} - x \right)^2 \tilde \mu_C(\d x) \\
        & \hspace{1.5cm} + \frac{\vartheta_t(1 - \kappa)}{J_t} \left( \int \frac{r_t^\mathsf L(\gamma_x) \tilde \mu_C(\d x)}{(1 - \kappa + \vartheta_t q_t(\gamma_x))^2} \right) \left( \int \frac{r_t^\mathsf R(\gamma_x) \tilde \mu_C(\d x)}{(1 - \kappa + \vartheta_t q_t(\gamma_x))^2} \right), 
    \end{aligned}
    \end{equation}
    where $q_t, r_t^\mathsf{L}, r_t^\mathsf{R}$ are defined in \cref{Thm:Stieltjes}, $\vartheta_t$ is the unique positive solution of the self-consistent equation
    \begin{equation} \label{eq:SCht}
        1 = \int \frac{\tilde \mu_C(\d x)}{1 - \kappa + \vartheta_t q_t(\gamma_x)},
    \end{equation}
    and
    \begin{equation} \label{eq:DefJt}
        J_t = \int  \frac{q_t(\gamma_x)}{(1 - \kappa + \vartheta_t q_t(\gamma_x))^2} \tilde \mu_C(\d x). 
    \end{equation}
\end{proposition}

\begin{proof}
Let $\bS=(\bC^\top\bC)^{1/2}$ and $\bGamma_\rho = (\bS^2 + \rho^2 \bI_d)^{1/2}$. From \cref{Lemma:BlockDecomposition1}, we have:
\begin{equation*}
    \bZ(t) = 2 \rho \chi_t^U(\bC) \bR \Big( \bR^\top q_t(\bGamma_\rho) \bR \Big)^{-1} \bR^\top \chi_t^V(\bS).
\end{equation*}
Now, expanding the Frobenius squared distance in the expression of $\mathcal{D}_d(t)$, we have the identity:
\begin{equation} \label{eq:LossFiniteDim}
    \mathcal{D}_d(t) = \frac{4 \rho^2}{d} \tr \Big(\chi_t^U(\bC)^\top \chi_t^U(\bC) \bH_t \chi_t^V(\bS)^2 \bH_t \Big) - \frac{4 \rho}{d} \tr \Big( \bC^\top \chi_t^U(\bC) \bH_t \chi_t^V(\bS) \Big) + \frac{1}{d} \tr \big( \bC^\top \bC \big),
\end{equation}
where
\begin{equation*}
    \bH_t = \bR \Big( \bR^\top q_t(\bGamma_\rho) \bR \Big)^{-1} \bR^\top. 
\end{equation*}
Using \cref{Lemma:RectangulartoSquare} and the notation of \cref{Thm:Stieltjes}, we have
\begin{equation*}
    \chi_t^U(\bC)^\top \chi_t^U(\bC) = \chi_t^U(\bS)^2 = \frac{1}{2\rho} r_t^\mathsf L(\bGamma_\rho). 
\end{equation*}
Similarly, we have:
\begin{equation*}
    \bC^\top \chi_t^U(\bC) = \bS \chi_t^U(\bS) = \frac{1}{\sqrt{2\rho}} \bS r_t^\mathsf L(\bGamma_\rho)^{1/2}. 
\end{equation*}
With these identities, we have expressed $\mathcal{D}_d(t)$ using traces of the matrix $\bH_t$ and spectral functions of $\bC^\top \bC$. We now derive the limit of each term in equation \eqref{eq:LossFiniteDim}. First of all, it is clear that:
\begin{equation*}
    \frac{1}{d} \tr \big( \bC^\top \bC \big) \xrightarrow[d \to \infty]{} \int x^2 \tilde \mu_C(\d x). 
\end{equation*}
    Then, we apply \cref{Thm:RMTConvergenceReal} to $\bD=\bS$ with $a(x) = q_t(\gamma_x)$ and $b(x)= x \sqrt{r_t^\mathsf{L}(\gamma_x) r_t^\mathsf{R}(\gamma_x)}$. Its assumptions hold since $q_t \geq 1$. We obtain:
\begin{equation*}
\begin{aligned}
    \frac{4\rho}{d}  \tr \Big( \bC^\top \chi_t^U(\bC) \bH_t \chi_t^V(\bS) \Big) &= \frac{2}{d} \tr \Big( \bS r_t^\mathsf L(\bGamma_\rho)^{1/2} r_t^\mathsf R(\bGamma_\rho)^{1/2} \bH_t \Big) \\
    &\xrightarrow[d \to \infty]{\mathrm{a.s.}} 2\vartheta_t \int \frac{x \sqrt{r_t^\mathsf L(\gamma_x) r_t^\mathsf R(\gamma_x)}}{1 - \kappa + \vartheta_t q_t(\gamma_x)} \tilde \mu_C(\d x),
\end{aligned}
\end{equation*}
where $\vartheta_t$ solves the self-consistent equation \eqref{eq:SCht}. Finally, for the last term, we use \cref{Thm:RMTConvergence2}:
\begin{equation*}
\begin{aligned}
    &\frac{4 \rho^2}{d} \tr \Big(\chi_t^U(\bC)^\top \chi_t^U(\bC) \bH_t \chi_t^V(\bS)^2 \bH_t \Big) = \frac{1}{d} \tr \Big(r_t^\mathsf{L}(\bGamma_\rho) \bH_t r_t^\mathsf{R}(\bGamma_\rho) \bH_t \Big) \\
    &\xrightarrow[d \to \infty]{\mathrm{a.s.}} \frac{(1 - \kappa) \vartheta_t}{J_t} \left( \int \frac{r_t^\mathsf{L}(\gamma_x)}{\Delta_t(\gamma_x)^2} \tilde \mu_C(\d x) \right) \left( \int \frac{r_t^\mathsf{R}(\gamma_x)}{\Delta_t(\gamma_x)^2} \tilde \mu_C(\d x) \right) + \vartheta_t^2 \int \frac{r_t^\mathsf{L}(\gamma_x) r_t^\mathsf{R}(\gamma_x)}{\Delta_t(\gamma_x)^2} \tilde \mu_C(\d x),
\end{aligned}
\end{equation*}
where $\Delta_t(\gamma) = 1 - \kappa + \vartheta_t q_t(\gamma)$ and, consistently with \cref{Thm:RMTConvergence2}, $J_t$ is given by equation \eqref{eq:DefJt}. Putting everything together, we obtain the expression in equation \eqref{eq:LossHighdim}. 
\end{proof}

\subsubsection{Small-Initialization Limit at Fixed Times}

We now take the $\rho \to 0$ limit at fixed times. We denote $\vartheta_{t,\rho}=\vartheta_t$ to recall the dependence on $\rho$. For $x \geq 0$, the expressions in \cref{Thm:Stieltjes} guarantee that:
\begin{equation*}
\begin{aligned}
    q_t(\gamma_x) &\xrightarrow[\rho \to 0]{} 1, \hspace{1.5cm} r_t^\mathsf L(\gamma_x) \xrightarrow[\rho \to 0]{} 0, \hspace{1.5cm} r_t^\mathsf R(\gamma_x) \xrightarrow[\rho \to 0]{} 0. 
\end{aligned}
\end{equation*}
For the convergence of $\vartheta_{t, \rho}$, we apply \cref{Prop:SelfConsistentGeneral} directly with parameter $\rho\to0$ and set:
\begin{equation*}
    h_\rho = \vartheta_{t,\rho}, \hspace{1.5cm} a_\rho(x) = q_t(\gamma_x),
\end{equation*}
with $c_\rho=1$. Due to the previous convergence of $q_t$, the limit function in \cref{Prop:SelfConsistentGeneral} is given by:
\begin{equation*}
    \Phi(u) = \frac{1}{1 - \kappa + u}.
\end{equation*}
The only solution to $\Phi(u) = 1$ is given by $u = \kappa$, hence \cref{Prop:SelfConsistentGeneral} guarantees that $\vartheta_{t, \rho}$ converges to $\kappa$ as $\rho \to 0$. 

A straightforward application of dominated convergence then guarantees that $J_t \longrightarrow 1$ as $\rho \to 0$, and therefore, taking the $\rho \to 0$ limit in equation \eqref{eq:LossHighdim}, we obtain:
\begin{equation*}
    \lim_{\rho \to 0} \lim_{d \to \infty} \mathcal{D}_d(t) = \int x^2 \tilde \mu_C(\d x). 
\end{equation*}
After multiplying by $\beta^{-1}$, this gives equation \eqref{eq:LossFixedtime}. This is exactly the limit one would obtain when replacing $\bZ(t)$ by $\bzero$ in the definition of $\mathcal{D}_d(t)$. 

\subsubsection{Small-Initialization Limit at Log-Timescale}

We now move on to the timescale $t_\rho = s \log(\rho^{-1})$. As previously, we let $x_s = (2s)^{-1}$ and assume $x_s > x_*$. Before deriving the limit of the loss, we start with an estimate in the small-initialization limit.

\begin{lemma} \label{Lemma:AsymptoticSmallRho2}
    Let $q_t,r_t^\mathsf L,r_t^\mathsf R$ be defined in \cref{Thm:Stieltjes}. For every $\rho > 0$, $t \geq 0$, and $x \geq 0$, with $\gamma_x = \sqrt{x^2 + \rho^2}$, one has
    \begin{equation} \label{eq:BoundsLogtime1}
        0\leq \frac{r_t^\mathsf L(\gamma_x)}{q_t(\gamma_x)}
        \leq\gamma_x-\rho,
        \hspace{1.5cm}
        0\leq \frac{r_t^\mathsf R(\gamma_x)}{q_t(\gamma_x)}
        \leq\gamma_x+\rho.
    \end{equation}
\end{lemma}
\begin{proof}
    For $\gamma > \rho$, the formulas in \cref{Thm:Stieltjes} give
    \begin{equation*}
        q_t(\gamma)-\frac{r_t^\mathsf L(\gamma)}{\gamma-\rho}
        =1+\frac{\rho}{\gamma}\big(1-e^{-2t\gamma}\big)\geq0.
    \end{equation*}
    The two bounds follow together with $r_t^\mathsf R=r_t^\mathsf L+2\rho q_t$. For $\gamma = \rho$, that is $x = 0$, we have $q_t(\rho) = e^{2t \rho}$, $r_t^\mathsf L(\rho) = 0$ and $r_t^\mathsf R(\rho) = 2 \rho e^{2t \rho}$, thus the bounds also hold. 
\end{proof}
We also recall \cref{Lemma:AsymptoticSmallRho1}, which derives the limit of $q_t, r_t^\mathsf{L}, r_t^\mathsf{R}$ at $t_\rho$ when $\rho \to 0$. Additionally, throughout the proof, we denote $\vartheta_\rho = \vartheta_{t_\rho}$ to simplify notation. 

\paragraph{Step 1. Asymptotic of the self-consistent function.} Again, we apply \cref{Prop:SelfConsistentGeneral} directly with parameter $\rho\to0$ and:
\begin{equation*}
    h_\rho = \vartheta_\rho, \hspace{1.5cm} a_\rho(x) = q_{t_\rho}(\gamma_x),
\end{equation*}
as well as $c_\rho = 1$. Then, \cref{Lemma:AsymptoticSmallRho1} shows that:
\begin{equation*}
    a_\rho(x)\xrightarrow[\rho\to0]{}
    \begin{cases}
        1,&x<x_s,\\
        +\infty,&x>x_s.
    \end{cases}
\end{equation*}
Hence the limit function in \cref{Prop:SelfConsistentGeneral} is given by:
\begin{equation*}
    \Phi(u) = \frac{1 - T_C(x_s)}{1 - \kappa + u}. 
\end{equation*}
Since $x_s > x_*$, we have $T_C(x_s) < \kappa$, so that $\Phi(u) = 1$ has a unique solution on $(0, \infty)$. By \textbf{2.} $(i)$ in \cref{Prop:SelfConsistentGeneral} and the definition of $T_C$, we then have the convergence:
\begin{equation*}
    \vartheta_\rho \xrightarrow[\rho \to 0]{} \kappa - \beta \kappa^* \int_{x_s}^\infty \phi_C(x) \d x > 0. 
\end{equation*}

\paragraph{Step 2. Asymptotics of the integrals.} Let us start with the asymptotic behavior of $J_{t_\rho}$. Since $\vartheta_\rho$ is bounded away from zero, the function
\begin{equation*}
    x \mapsto \frac{q_{t_\rho}(\gamma_x)}{(1-\kappa+\vartheta_\rho q_{t_\rho}(\gamma_x))^2}
\end{equation*}
is uniformly bounded. By \cref{Lemma:AsymptoticSmallRho1} and the preceding limit $\vartheta_\rho\to\kappa-T_C(x_s)$, it converges pointwise to $(1-T_C(x_s))^{-2}$ for $x<x_s$ and to zero for $x>x_s$. Since $\tilde \mu_C$ has no atom at $x_s$, dominated convergence in equation \eqref{eq:DefJt} gives
\begin{equation*}
    J_{t_\rho} \xrightarrow[\rho \to 0]{} \frac{1}{1 - T_C(x_s)}. 
\end{equation*}
Let us now consider the distance in equation \eqref{eq:LossHighdim}. Starting with the second term, we write:
\begin{equation*}
    \frac{r_{t_\rho}^\mathsf a(\gamma_x)}{(1-\kappa+\vartheta_\rho q_{t_\rho}(\gamma_x))^2}
    =\frac{r_{t_\rho}^\mathsf a(\gamma_x)}{q_{t_\rho}(\gamma_x)}
    \frac{q_{t_\rho}(\gamma_x)}{(1-\kappa+\vartheta_\rho q_{t_\rho}(\gamma_x))^2}.
\end{equation*}
The bounds in equation \eqref{eq:BoundsLogtime1} and the pointwise convergence obtained in \cref{Lemma:AsymptoticSmallRho1} allow us to apply dominated convergence, so that for $\mathsf{a} = \mathsf{L},\mathsf{R}$,
\begin{equation*}
    \int \frac{r_{t_\rho}^\mathsf{a}(\gamma_x)}{(1 - \kappa + \vartheta_\rho q_{t_\rho}(\gamma_x))^2}\tilde\mu_C(\d x) \xrightarrow[\rho \to 0] {} 0.
\end{equation*}
Therefore, the second term in equation \eqref{eq:LossHighdim} converges to zero. 

Let us now consider the first term. We have the pointwise convergence as a consequence of \cref{Lemma:AsymptoticSmallRho1}
\begin{equation*}
     \frac{\vartheta_\rho \sqrt{r_{t_\rho}^\mathsf L(\gamma_x) r_{t_\rho}^\mathsf R(\gamma_x)}}{1 - \kappa + \vartheta_\rho q_{t_\rho}(\gamma_x)}
     \xrightarrow[\rho \to 0]{}
     \begin{cases}
         0, & x < x_s, \\
         x, & x > x_s.
     \end{cases}
\end{equation*}
Additionally, \cref{Lemma:AsymptoticSmallRho2} guarantees that:
\begin{equation*}
    \frac{\vartheta_\rho \sqrt{r_{t_\rho}^\mathsf L(\gamma_x) r_{t_\rho}^\mathsf R(\gamma_x)}}{1 - \kappa + \vartheta_\rho q_{t_\rho}(\gamma_x)} \leq \frac{\sqrt{r_{t_\rho}^\mathsf L(\gamma_x) r_{t_\rho}^\mathsf R(\gamma_x)}}{q_{t_\rho}(\gamma_x)} \leq x. 
\end{equation*}
Then, dominated convergence gives:
\begin{equation*}
    \lim_{\rho \to 0} \lim_{d \to \infty }\mathcal{D}_d(t_\rho) = \int_0^{x_s} x^2 \tilde \mu_C(\d x) = \beta \kappa^* \int_{x_0}^{x_s} x^2 \phi_C(x) \d x. 
\end{equation*}
After multiplying by $\beta^{-1}$, this is precisely equation \eqref{eq:LossLogtime}.

\subsubsection{Comparison with the Wasserstein Distance} \label{App:Subsubsec:SpectralDynamics}

We now explain why the following two quantities have the same limit on the timescale $t = s\log(\rho^{-1})$, after taking the high-dimensional and small-initialization limits:
\begin{equation} \label{eq:FrobeniusWasserstein}
    \frac{1}{\min(p,d)} \big\| \bZ(t) - \bC \big\|_F^2, \hspace{1.5cm} \frac{1}{\min(p,d)} \sum_{i=1}^{\min(p,d)} \big( \sigma_i(t) - \sigma_i \big)^2,
\end{equation}
where $\sigma_1(t), \dots, \sigma_{\min(p,d)}(t)$ and $\sigma_1, \dots, \sigma_{\min(p,d)}$ are the ordered singular values of $\bZ(t)$ and $\bC$.

\begin{proposition}[von Neumann's trace inequality] \label{Prop:VonNeumannInequality}
    Let $\bA, \bB \in \R^{p \times d}$, and denote by $\bsigma_A, \bsigma_B \in \R^{\min(p, d)}$ the ordered singular values of $\bA, \bB$. Then:
    \begin{equation*}   
        \tr \big( \bA^\top \bB \big) \leq \bsigma_A^\top \bsigma_B,
    \end{equation*}
    and equality holds if and only if there exist $\bU \in O_p(\R), \bV \in O_d(\R)$ such that:
    \begin{equation*}
        \bA = \bU \mathrm{diag}(\bsigma_A) \bV^\top, \hspace{1.5cm} \bB = \bU \mathrm{diag}(\bsigma_B) \bV^\top. 
    \end{equation*} 
\end{proposition}

A proof of this fact can be found in \citet[Theorem~2.1]{lewis1995convex}. The notation $\mathrm{diag}$ is understood as mapping $\bsigma \in \R^{\min(p,d)}$ into a rectangular diagonal matrix in $\R^{p \times d}$ with diagonal entries $\sigma_1, \dots, \sigma_{\min(p,d)}$. Consequently, the equation:
\begin{equation*}
    \big\| \bA - \bB \big\|_F^2 = \big\| \bsigma_A - \bsigma_B \big\|_2^2
\end{equation*}
is exactly the equality case in \cref{Prop:VonNeumannInequality}. The second quantity in equation \eqref{eq:FrobeniusWasserstein} is the squared Wasserstein distance between the empirical singular-value distributions: the optimal transport matches the singular values in order. Thus, when the corresponding singular vectors are aligned, the Frobenius error equals this transport cost. In our setting, the equality of \eqref{eq:WassersteinDistance} and \eqref{eq:LossLogtime} shows that the contribution of misalignment vanishes asymptotically.

\section{Numerical Simulations} \label{App:Simulations}
This appendix describes the numerical methods used for the figures in the main text. The parameters specific to each experiment are given in the corresponding figure caption.

\paragraph{Finite-dimensional simulations.} We simulate the dynamics \eqref{eq:GFDynamics} on $\bU(t), \bV(t)$ directly. For a given $d$, we fix the values of the parameters $\tau, \kappa, \kappa^*$ and set $p=\lfloor \tau d\rceil$, $m=\lfloor\kappa d\rceil$, and $m^*=\lfloor\kappa^*\min(p,d)\rceil$. The non-zero singular values of $\bC$ are sampled from the density $\phi_C$, and we take the corresponding singular vectors to be the canonical basis. This choice is without loss of generality for the observables considered here, by rotational invariance of the initialization.

The matrix $\bR\in\mathrm{St}(d,m)$ is obtained by applying a QR decomposition to a standard Gaussian matrix, and we initialize
\begin{equation*}
    \bU_0=\bzero,
    \hspace{1.5cm}
    \bV_0=\sqrt{2\rho}\,\bR.
\end{equation*}
We use explicit Euler steps with stepsize $\eta$:
\begin{equation*}
\begin{aligned}
    \bU_{n+1}
    &=\bU_n+\eta\big(\bC-\bU_n\bV_n^\top\big)\bV_n,\\
    \bV_{n+1}
    &=\bV_n+\eta\big(\bC^\top-\bV_n\bU_n^\top\big)\bU_n,
\end{aligned}
\end{equation*}
and $t_n = n \eta$. These steps approximate the gradient flow \eqref{eq:GFDynamics} with an error of order $\eta$. In the linearized regime, a mode associated with a singular value $\sigma$ grows like $(1+\eta\sigma)^{t/\eta}$ instead of $e^{\sigma t}$, which changes its growth rate by a relative amount of order $\eta\sigma$. For $\eta=10^{-3}$, this shifts the logarithmic timescales of \cref{fig:Fig6,fig:Fig7} by at most about $0.1\%$. As a check, rerunning two repetitions of \cref{fig:Fig7} with stepsize $\eta/2$ changes the loss and the spectral mass below $x_s$ by at most $2 \cdot 10^{-3}$.

When several values of $\rho$ are compared, we use the same realization of $\bC$ and $\bR$. The curves are averaged over independent realizations, with the number of repetitions stated in each caption. Losses, singular values, and projector distances are computed from singular value decompositions at the saved times. 

\paragraph{Self-consistent equations.} We next describe how the high-dimensional curves are evaluated. Integrals with respect to the continuous part of $\tilde\mu_C$ are approximated by Gauss--Legendre quadrature on the support of $\phi_C$ and the atom at zero is added separately. For a spectral parameter $w=E+\mathrm{i}\varepsilon$, the scalar self-consistent equation has the form
\begin{equation} \label{eq:NumericalSelfConsistent}
    1=\int
    \frac{\tilde\mu_C(\d x)}
    {1-\kappa+\kappa g_t^\mathsf a(w)
    \big(r_t^\mathsf a(\gamma_x)-wq_t(\gamma_x)\big)},
\end{equation}
with $\mathsf a=\mathsf L,\mathsf R$. After quadrature, equation \eqref{eq:NumericalSelfConsistent} is a scalar complex equation. We solve it by damped Newton iterations, using the explicit derivative of its right-hand side. To select the correct branch, we first continue the positive solution from the negative real axis to the first value $E+\mathrm{i}\varepsilon$. We then move along a grid of values of $E$ and use the previous solution as the next initial value. We keep the solution whose imaginary part has the same sign as that of $w$, and monitor the residual of the self-consistent equation. 

Once $g_t^\mathsf a(w)$ is known, we evaluate the Stieltjes transforms in equations \eqref{eq:StieltjesU}, \eqref{eq:StieltjesV}. The regular part of the density is approximated through the Stieltjes--Perron formula (\cref{Thm:StieltjesPerron}):
\begin{equation*}
    \phi_t^\bullet(E)
    \simeq \frac{1}{\pi}\Im \,  \mathfrak m_t^\bullet(E+\mathrm{i}\varepsilon),
\end{equation*}
for $\bullet=U,V$. The parameter $\varepsilon$ is kept small and fixed on each plot. Before comparing with histograms of non-zero eigenvalues, we subtract the known atoms at zero and, for $\bV\bV^\top$, at $2\rho$, and retain the continuous mass. The singular-value density of $\bZ$ is then obtained from \eqref{eq:StieltjesUtoZ} by the pushforward $\theta_\rho(x)=\sqrt{x(x+2\rho)}$.

The scalar positive equations entering the range-projector formulas are solved by a bracketed root method, using the logarithm of the unknown to remain stable at long times. Finally, the small-initialization and logarithmic-time curves are evaluated directly from the formulas of \cref{Thm:ProjectorLimit,Thm:StieltjesLimit,Thm:LossTimescales}, using the same quadrature rule. No parameters are fitted to the finite-dimensional simulations.

\paragraph{Checks.} We checked the numerical methods in several ways. At $d = 40$, the Euler iterates stay within $1.2 \cdot 10^{-2}$ (in relative Frobenius norm) of a fourth-order Runge--Kutta solution, which itself matches the closed form of \cref{Prop:SolutionOja} up to $10^{-8}$. Halving the stepsize $\eta$ shifts the activation times of \cref{fig:Fig6} by less than $0.1\%$. At fixed $\rho = 0.1$, with the singular values of $\bC$ set to the quantiles of $\phi_C$, the distance between the simulations and the formulas of \cref{Thm:Stieltjes,Thm:Projector1} decreases with $d$, and is at most about $10^{-3}$ at $d = 800$, for $\tau \in \{0.7, 1, 1.4\}$. Doubling the number of quadrature nodes changes the theoretical densities by less than $0.2\%$ of their maximum. The code of the simulations, of the figures and of these checks is available at \url{https://github.com/simonmartin15/silent-alignment-linear-networks}.

\end{document}